\documentclass[10pt,twocolumn,letterpaper]{article}

\usepackage[pagenumbers]{cvpr} 

\def\cvprPaperID{9545} 
\def\confName{CVPR}
\def\confYear{2022}

\PassOptionsToPackage{sort&compress,numbers}{natbib}
\usepackage{url}            
\usepackage{booktabs}       
\usepackage{amsfonts}       
\usepackage{nicefrac}       

\newcommand{\figref}[1]{Figure~\ref{#1}}
\newcommand{\tabref}[1]{Table~\ref{#1}}
\newcommand{\secref}[1]{Section~\ref{#1}}

\usepackage{graphicx}
\usepackage{times}
\usepackage{epsfig}
\usepackage{microtype}
\usepackage{amsmath}
\usepackage{amssymb , bbm}
\usepackage[inline]{enumitem}
\usepackage{epstopdf}
\usepackage{subcaption}
\usepackage{caption}

\usepackage{tabularx}
\usepackage{multirow}
\usepackage{pifont}
\usepackage[table]{xcolor} 
\usepackage[toc,page]{appendix}
\usepackage[super]{nth}

\usepackage{float}
\usepackage{tabulary,multirow,overpic,xcolor}
\usepackage[english]{babel}

\DeclareCaptionFont{xipt}{\fontsize{8}{10}\mdseries}
\usepackage[font=xipt]{caption}

\usepackage{enumitem}
\usepackage[pagebackref=true,breaklinks=true,colorlinks=true,bookmarks=false,citecolor=blue]{hyperref}
\usepackage[english]{babel}

\newenvironment{tight_itemize}{
\begin{itemize}[leftmargin=10pt]
  \setlength{\topsep}{0pt}
  \setlength{\itemsep}{2pt}
  \setlength{\parskip}{0pt}
  \setlength{\parsep}{0pt}
}{\end{itemize}}

\renewcommand{\eqref}[1]{Eq.~(\ref{#1})}

\newcommand{\red}[1]{{#1}}

\newcommand{\Ours}{FLAVR}
\newcommand{\twox}{$2\times$}
\newcommand{\fourx}{$4\times$}
\newcommand{\eightx}{$8\times$}
\newcommand{\I}{\hat{I}}
\newcommand{\LL}{\mathcal{L}}

\newcommand{\stimes}{{\times}}

\newcommand{\new}[1]{\textcolor{black}{#1}}

\begin{document}

\title{FLAVR: Flow-Agnostic Video Representations for Fast Frame Interpolation}

%

\author{
Tarun Kalluri~\thanks{Work done during TK's internship at Facebook AI.}\\
UCSD
\and
Deepak Pathak\\
CMU
\and
Manmohan Chandraker\\
UCSD
\and
Du Tran\\
Facebook AI
\and
\texttt{\url{https://tarun005.github.io/FLAVR/}}
}

\maketitle

\begin{abstract}

Most modern frame interpolation approaches rely on explicit bidirectional optical flows between adjacent frames, thus are sensitive to the accuracy of underlying flow estimation in handling occlusions while additionally introducing computational bottlenecks unsuitable for efficient deployment. In this work, we propose a flow-free approach that is completely end-to-end trainable for multi-frame video interpolation. Our method, FLAVR, is designed to reason about non-linear motion trajectories and complex occlusions implicitly from unlabeled videos and greatly simplifies the process of training, testing and deploying frame interpolation models. Furthermore, FLAVR delivers up to $6\times$ speed up compared to the current state-of-the-art methods for multi-frame interpolation while consistently demonstrating superior qualitative and quantitative results compared with prior methods on popular benchmarks including Vimeo-90K, Adobe-240FPS, and GoPro. Finally, we show that frame interpolation is a competitive self-supervised pre-training task for videos via demonstrating various novel applications of FLAVR including action recognition, optical flow estimation, motion magnification, and video object tracking. Code and trained models are provided in the supplementary material.

\end{abstract}

\section{Introduction}

\begin{figure}
    \begin{center}
    \includegraphics[width=0.5\textwidth]{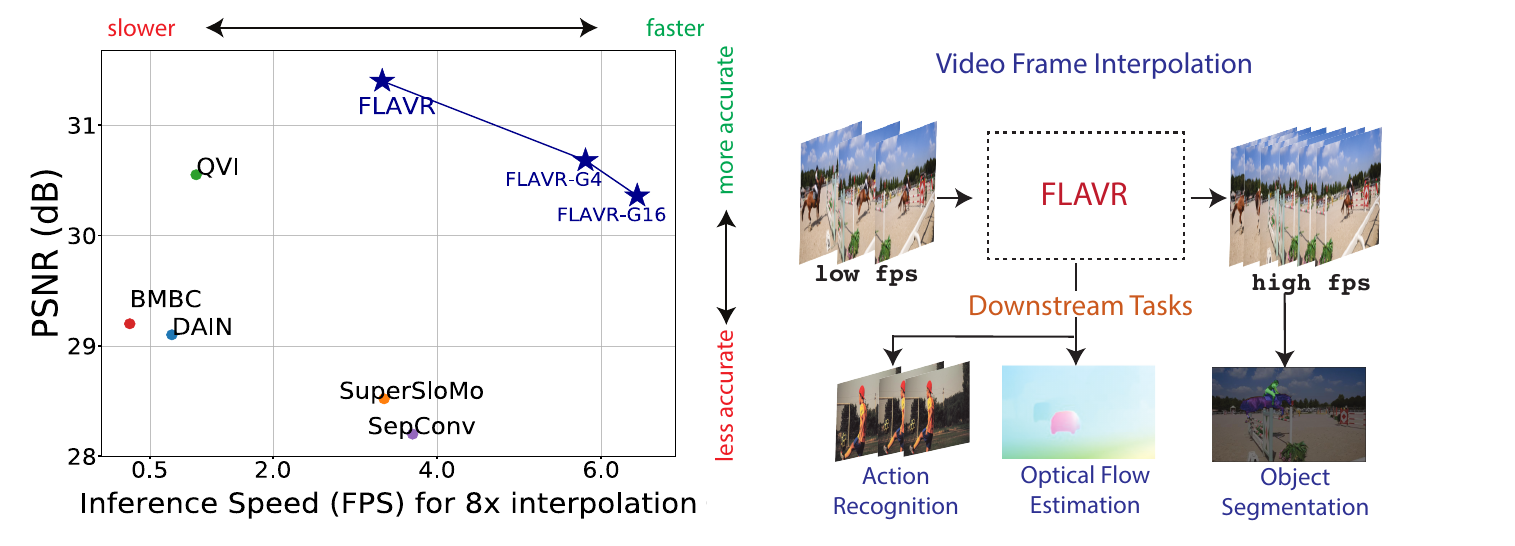}
    \end{center}
    \vspace{-16pt}
    \caption{{\bf Our contributions} We propose FLAVR, a simple and efficient architecture for single shot multi-frame interpolation. The plot of accuracy (PSNR) vs. inference speed (fps) of FLAVR compared with current methods on GoPro 8x interpolation with 512$\times$512 input images. FLAVR is {\bf 6}x faster than the current most accurate method (QVI) and {\bf 2}x faster than the current fastest method (SuperSloMo) while maintains the same quality. FLAVR is also a useful self-supervised pretext task for various downstream applications.
    }
    \vspace{-12pt}
    \label{fig:intro_pic}
\end{figure}

Video frame interpolation~\cite{meyer2018phasenet, bao2019depth, xue2019video, lee2020adacof, jiang2018super, niklaus2017video, niklaus2018context, liu2017video, choi2020channel, niklaus2020softmax} aims to generate non-existent intermediate frames in a video between existing ones that are spatially and temporally coherent with the rest of the video, finding applications in overcoming the limited acquisition frame rate and exposure time of commercial video cameras.
Traditionally, frame interpolation has been treated as a predominantly graphics problem where the approaches are complicated and hard coded. A large body of prior works use \textit{flow warping} for frame interpolation~\cite{jiang2018super, xue2019video, niklaus2020softmax}, where the input frames are used to estimate (often bidirectional) optical flow maps from a pretrained flow prediction network, possibly along with additional information like monocular depth maps~\cite{bao2019depth} and occlusion masks~\cite{bao2019memc}. The frames at intermediate time steps are then interpolated either by using backward~\cite{jiang2018super, bao2019depth} or forward warping~\cite{niklaus2018context, niklaus2020softmax}. However, these optical flow-based approaches, as well as proposed alternatives~\cite{niklaus2017video, niklaus2017video_2, peleg2019net, cheng2020video, lee2020adacof}, have to confront one or more of the following limitations:
\begin{enumerate*}
    \item{\bf Computational Costs}: As they rely on optical flow and pixel level warping procedures, they are inefficient at both training and inference in terms of speed and efficiency making them less suitable for end applications. \red{For example, QVI~\cite{xu2019quadratic}, DAIN~\cite{bao2019depth} and BMBC~\cite{park2020bmbc} take order of seconds to generate frames for \eightx interpolation (\figref{fig:intro_pic}) while requiring users to deploy custom CUDA kernels that prohibit seamless deployment across edge devices.}
    
    \item{\bf Modeling Complex Trajectories}: The modeling capacity is limited to account for only linear~\cite{jiang2018super, bao2019depth} or quadratic~\cite{xu2019quadratic, chi2020all} motion trajectories, and extending these to account for more complex motions is non-trivial using existing approaches. 
    +
    \item{\bf Representation Inflexibility}: By accepting pre-computed optical flows as inputs, current methods focus on learning only spatial warping and interpolation, thus the representations learned in the process are not transferable to tasks beyond frame interpolation. 

\end{enumerate*}

In this work, we aim to achieve a good trade-off between visual quality and inference speed for video interpolation. We do so by proposing \Ours{} (Flow-Agnostic Video Representation network), which jointly addresses the aforementioned limitations. \Ours{} is a simple, scalable approach for frame interpolation that utilizes spatio-temporal convolutions for predicting intermediate frames of a video. Without demanding access to external flow or depth maps, \Ours{} can make end-to-end multiple-frame predictions in a single forward pass. It implicitly handles complex motions and occlusions through learning from large scale video data, significantly improving ease of deployment and inference speed compared to prior approaches (\figref{fig:intro_pic}, \figref{fig:TimeVsK}), while achieving state-of-the art interpolation accuracy (\tabref{tab:twox_interpolation}, \tabref{tab:eightx_interpolation}).  

We also posit that models learned from raw videos should be able to simultaneously reason about intricate synergy between objects, motions and actions for accurate frame interpolation. This is because different actions and objects have different motion signatures, and it is essential to precisely capture these properties through the representations learned for accurate frame interpolation. We ground this argument in the context of self-supervised representation learning from videos~\cite{doersch2015unsupervised, pathak2017learning, wang2015unsupervised, he2020momentum}. While popular pretext tasks like frame ordering~\cite{xu2019self, lee2017unsupervised, fernando2017self, misra2016shuffle, wei2018learning}, pixel/color tracking~\cite{vondrick2018tracking, wang2019learning} or contrastive learning~\cite{han2019video, han2020memory, gordon2020watching} are tailored to suit specific downstream applications, we show that frame interpolation offers a more generic representation learning objective owing to its combined motion and semantic understanding. To this end, we show the utility of FLAVR pretraining to improve performance on a variety of downstream tasks like action recognition, optical flow estimation and video object segmentation. 
%
In summary:
\begin{tight_itemize}
    \item We propose \Ours{}, a scalable, flow-free, efficient 3D CNN architecture for video frame interpolation. To the best of our knowledge, \Ours{} is the first video frame interpolation approach that is {\bf both} \emph{optical flow-free} and able to make \emph{single-shot multiple-frame predictions} (\secref{sec:arch_design}).
    \item \Ours{} is quantitatively and qualitatively superior or comparable to current approaches on multiple standard benchmarks including Vimeo-90K, UCF101, DAVIS, Adobe, and GoPro \red{ while offering the best trade-off in terms of accuracy and inference speed for video interpolation} (\secref{sec:state-of-the-art}, \figref{fig:intro_pic} and \ref{fig:SpeedVsAccuracy}).
    \item We demonstrate that video representations self-supervisedly learned by \Ours{} can be useful for various downstream tasks such as action recognition, optical flow estimation and video object segmentation (\secref{sec:SSL}).
\end{tight_itemize}

\section{Related Work}

\noindent {\bf Video Frame Interpolation} Video frame interpolation is a classical computer vision problem~\cite{mahajan2009moving}
and recent methods take one of phase based~\cite{meyer2015phase, meyer2018phasenet}, kernel based~\cite{niklaus2017video, niklaus2017video_2, peleg2019net,  cheng2020video, liu2017video, shi2020video}, or flow based approaches, of which flow-based methods\cite{jiang2018super, bao2019depth, xu2019quadratic, xue2019video, chi2020all, liu2019deep, bao2019memc, yuan2019zoom, niklaus2018context, yu2013multi, zhang2019flexible, niklaus2020softmax, huang2020rife, siyao2021deep} are most successful. The key idea in flow-based methods is to 
%
use a flow prediction network, \emph{e.g.} PWC-Net~\cite{sun2018pwc}, to compute bidirectional optical flow between the input frames~\cite{jiang2018super} that guides frame synthesis along with predicting occlusion masks~\cite{jiang2018super, bao2019memc, xue2019video} or monocular depth maps~\cite{bao2019depth} to reason about occlusions.
%
%
While being largely successful in generating realistic intermediate frames, 
their performance is limited by the accuracy of the underlying flow estimator, which can be noisy in presence of complex occlusions resulting noticeable artifacts. They also assume uniform linear motion between the frames which is far from ideal for real world videos. Most importantly, the flow prediction and subsequent warping make frame prediction slow prohibiting fast interpolation. Recent works relax the linear motion assumption using quadratic warping~\cite{xu2019quadratic, liu2020enhanced} at the cost of increased model complexity and inference time.
%
CAIN~\cite{choi2020channel} uses channel attention as suitable ingredient for frame interpolation but fails to capture complex spatio-temporal dependencies explicitly between input frames. Moreover, many recent methods are only aimed towards single frame interpolation~\cite{huang2020rife, tulyakov2021time, siyao2021deep}. We address all these issues in this work by designing an end to end architecture that directly predicts any number of intermediate frames from a given video by learning to reason motion trajectories and properties through 3D space-time convolutions while jointly optimizing for output quality and inference time.
%


\noindent {\bf Spatio-temporal Filtering} Due to their proven success in capturing complex spatial and temporal dependencies, 3D space-time convolutions are very commonly used in video understanding tasks like action recognition~\cite{tran2015learning, tran2018closer, xie2018rethinking, carreira2017quo, feichtenhofer2019slowfast}, action detection~\cite{cdc,Xu2017iccv}, and captioning~\cite{xu2016msr-vtt}. We explore the use of 3D convolutions for the problem of temporal frame interpolation which requires modeling complex temporal abstractions between inputs for generating accurate and sharp predictions.

\noindent {\bf Video Self-Supervised Representation Learning}
Self-supervised learning deals with training unlabeled videos on artificial pretext tasks~\cite{doersch2015unsupervised} to extract semantic representations that serve as useful priors for sparsely labeled downstream tasks. Videos contain rich source of information in the form of temporal consistency and frame ordering, and prior works make use of such cues to build pretext tasks like predicting ordering of frames~\cite{xu2019self, lee2017unsupervised, fernando2017self, misra2016shuffle, wei2018learning}, 
correspondence across time~\cite{jayaraman2016slow, wang2015unsupervised} or contrastive predictive coding~\cite{han2019video, han2020memory}. In contrast to these approaches, we explore using video frame interpolation as a unified pretext task for both low-level and high-level downstream tasks like action recognition and optical flow.


\begin{figure*}
     \begin{center}
     \begin{subfigure}[b]{0.49\textwidth}
        \centering
        \includegraphics[width=\textwidth]{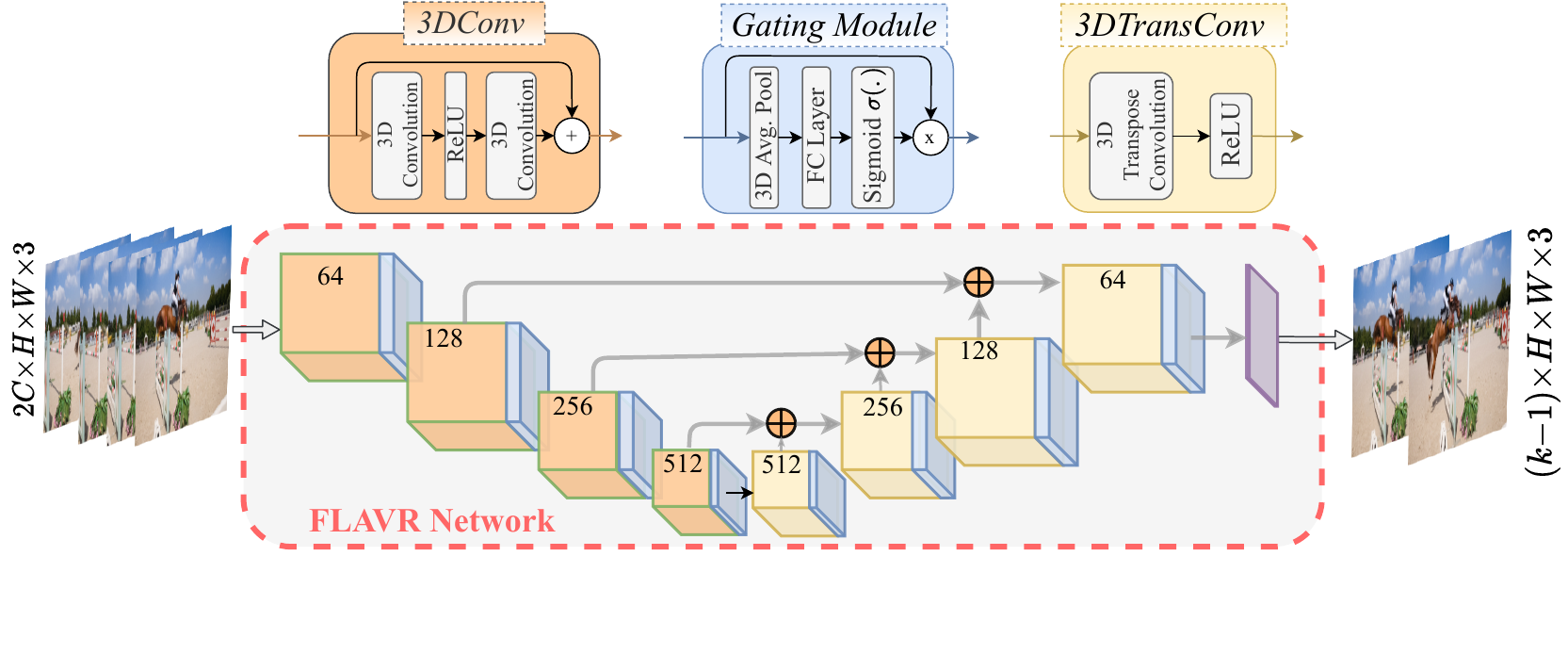}
        \captionsetup{width=\textwidth}
        \caption{Overview of the proposed architecture}
        \label{fig:arch}
     \end{subfigure}
     \hfill
     \begin{subfigure}[b]{0.49\textwidth}
        \centering
        \includegraphics[width=\textwidth]{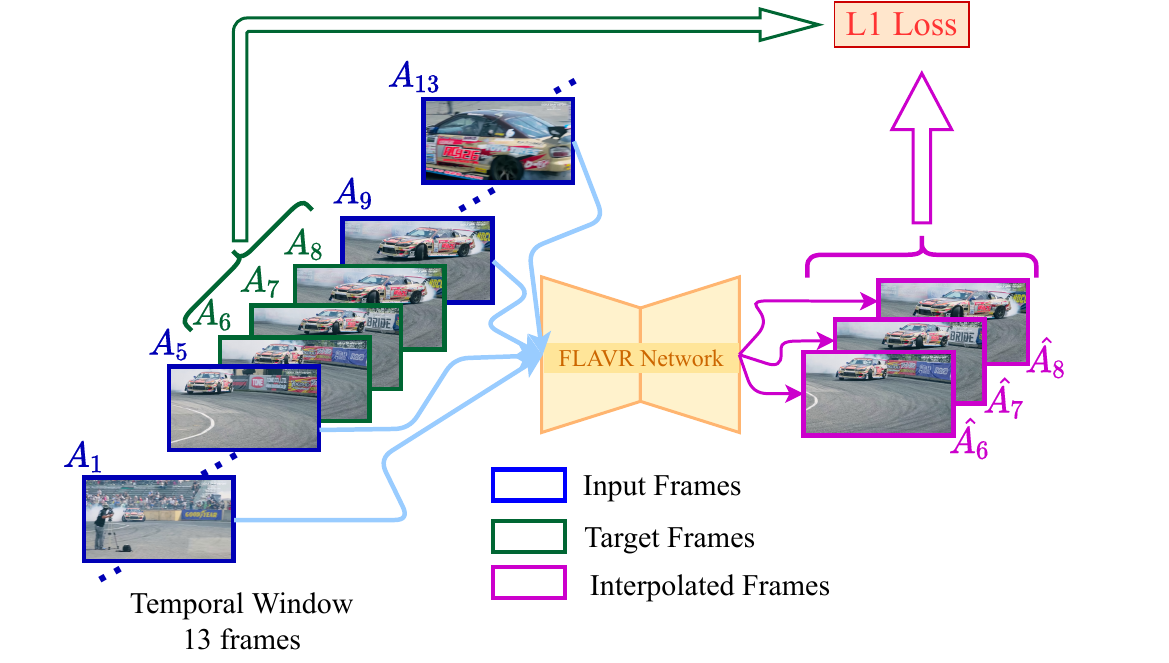}
        \caption{Sampling procedure}
        \label{fig:sampling}
     \end{subfigure}
     \end{center}
\vspace{-12pt}
\caption{{\bf FLAVR Architecture}. (a) Our FLAVR is U-Net style architecture with 3D space-time convolutions (orange blocks) and deconvolutions (yellow blocks). We use channel gating after all (de-)convolution layers (blue blocks). The final prediction layer (the purple block) is implemented as a convolution layer to project the 3D feature maps into $(k{-}1)$ frame predictions. This design allows FLAVR to predict multiple frames in one inference forward pass. (b) A concrete example of our sampling procedure for 4$\times$ interpolation ($k{=}4$) with 4-frame input ($C{=}2$). Best viewed in color.} 
\vspace{-8pt}
\label{fig:architecture}
\end{figure*}


\section{Frame Interpolation using \Ours{}}
\label{sec:arch_design}

In video frame interpolation, the task is to generate a high frame-rate video from a lower frame-rate input video. We define $k$ as the \textit{interpolation factor}, where $k\times$-video frame interpolation corresponds to generating $(k{-}1)$ additional intermediate frames between every pair of original frames in the input video, that are both spatially and temporally consistent with the rest of the video. Prior approaches are either specifically designed for \twox{} interpolation~\cite{choi2020channel, lee2020adacof, huang2020rife, tulyakov2021time, siyao2021deep} or require multiple inferences for predicting all the $k$ frames~\cite{xu2019quadratic, bao2019depth, bao2019memc, park2020bmbc}. In contrast, our aim is to design a framework which is simple yet enables single-shot $k\times$-prediction for any value of $k$. Since training on, and generating, long videos are beyond the capacity of current hardware, we propose a simple sampling procedure for efficient training on raw videos, followed by the construction of the network architecture.

{\bf Sampling Training Data from Unlabeled Videos} We can directly generate inputs and ground truths required for training from raw videos as follows. Let $k$ be the interpolation factor, $V$ is the original video with a frame rate $f$ FPS. In order to generate training data for the $k\times$-video frame interpolation problem, we sub-sample frames of $V$ with a sampling stride of $k$ to form a low frame rate video $\bar{V}$ with $\frac{f}{k}$ fps. Then, to perform interpolation between any two frames at position $(i,i{+}1)$, given by $ A_i , A_{i+1} $, of $\bar{V}$, we use a moving temporal window of size $2C$ in $\bar{V}$ centered around $A_i$ and $A_{i+1}$ as the input, and all frames between $A_i$ and $A_{i+1}$ in original video $V$ as the ground truth. This produces an input clip of size $2C$ frames (including $ A_i$ and $A_{i+1} $) and output clip of size $k{-}1$. 
\Ours{} is flexible to handle any temporal context $C$ instead of just the immediate neighbors $A_i$, $A_{i+1}$, which helps us to model complex trajectories and improve interpolation accuracy. 
The sampled input frames are concatenated in the temporal dimension resulting in input dimension $2C \stimes H \stimes W \stimes 3$, where $H,W$ are the spatial dimensions of the input video.

An illustration of this sampling procedure is demonstrated in \figref{fig:sampling} for the case of \fourx{} interpolation ($k{=}4$) with two context inputs from the past and future ($C=2$). In this case, the frames $\{A_{1} , A_{5} , A_{9} , A_{13}\}$ are used as inputs to predict the $3$ intermediate frames of $\{A_{6}, A_{7}, A_{8}\}$. Intuitively, the frames in the immediate neighborhood would be more relevant for frame interpolation than frames farther out.
In our experiments, we find that for most common settings, using four context frames ($C=2$) is sufficient for accurate prediction on the datasets considered. 
We present a detailed study on the effect of the input context $C$ in \textit{supplementary material}.

{\bf Architecture Overview} We present the proposed architecture of \Ours{} in~\figref{fig:arch}. \Ours{} is a 3D U-Net obtained by extending the popular 2D Unet~\cite{ronneberger2015u} used in pixel generation tasks, by replacing all the 2D convolutions in the encoder and decoder with 3D convolutions (\textit{3DConv}) to accurately model the temporal dynamics between the input frames, invariably resulting in better interpolation quality. Each 3D filter is a 5-dimensional filter of size $ c_i \stimes c_o \stimes t \stimes h \stimes w $, where $t$ is the temporal size and $(h,w)$ is the spatial size of the kernel. $c_i$ and $c_o$ are the number of input and output channels in the layer. The additional temporal dimension is useful in modeling the temporal abstractions like motion trajectories, actions or correspondences between frames in the video. We observed that our network indeed learns non-trivial representations along the temporal dimensions that can be reused in downstream tasks like action recognition with limited labeled data (\secref{sec:SSL}).

Practically any 3D CNN architecture can be used as the encoder backbone, and we use ResNet-3D (R3D) with 18 layers~\cite{tran2018closer} as our base backbone. We evaluate different variants of 3D CNNs with group convolutions~\cite{Tran19} as backbones to achieve the best accuracy/speed trade-off and present the complete analysis and results in \figref{fig:SpeedVsAccuracy}. We remove the last classification layer from R3D-18, resulting in 5 conv blocks \textit{conv1} to \textit{conv5}, each made up of two 3D convolutional layers and a skip connection.
We also remove all temporal striding, as downsampling operations like striding and pooling are known to remove details that are crucial for generating sharper images. However, we do use spatial stride of $2$ in \textit{conv1}, \textit{conv3} and \textit{conv4} blocks of the network to keep the computation manageable.

The decoder essentially constructs the output frames from a deep latent representation captured by the encoder by using progressive, multi-scale feature upsampling and feature fusion. For upsampling, we use 3D transpose convolution layers (\textit{3DTransConv}) with a stride of $2$. To handle the commonly observed checkerboard artefacts~\cite{odena2016deconvolution}, we add a \textit{3DConv} layer after the last \textit{3DTransConv} layer. We also include skip connections that directly combine encoder features with the corresponding decoder along the channels to fuse the low level and high level information necessary for accurate and sharp interpolation.

The output of the decoder, which is a 3D feature map, is then passed through a temporal fusion layer, implemented by a 2D conv, in which the features from the temporal dimension are concatenated along the channels and fused into a 2D spatial feature map. This helps to aggregate and merge information present in multiple frames for prediction. Finally, this output is passed through a $7 \stimes 7$ 2D convolution kernel that predicts output of size $H \stimes W \stimes 3(k{-}1)$, which is then split along the channel dimension to get the $(k{-}1)$ output frames. Our network is designed to efficiently handle interpolation for any value of $k$ with minimum changes to the architecture.

{\bf Spatio-Temporal Feature Gating} Feature gating technique is used as a form of self-attention mechanism in deep neural networks for action recognition~\cite{miech2017learnable,xie2018rethinking}, image classification~\cite{hu2018squeeze} and video interpolation~\cite{choi2020channel}. We apply the gating module after every layer in our architecture. Given an intermediate feature dimension of size $f_{i} = C \stimes T \stimes H \stimes W$, the output $f_{o}$ of the gating layer is given by
$
    f_{o} = \sigma(W . pool(f_{i}) + b) \odot f_i
$
where $W \in \mathbb{R}^{C \stimes C}$ and $b \in  \mathbb{R}^{C}$ are learnable weight and bias parameters, $pool$ is a spatio-temporal pooling layer and $\odot$ is element-wise product along the channel dimension. Such a feature gating mechanism would suitably learn to upweight and focus on certain relevant dimensions of the feature maps that learn useful cues for frame interpolation, like motion boundaries.

{\bf Loss Function} We can now train the whole network end to end using a pixel level loss like L1 loss between the predicted and ground truth frames, 
$
    \LL(\{\I\} , \{I\}) {=} \frac{1}{N} \sum_{i=1}^N \sum_{j=1}^{k-1} || \I_j^{(i)} - I_j^{(i)} ||_1
$
\noindent where $\{\I_j^{(i)}\}$ and $\{I_j^{(i)}\}$ are the j-th predicted and the j-th ground truth frame of the $i^{\text{th}}$ training clip, $k$ is the interpolation factor, and $N$ is the size of the mini-batch used in training.

{\bf Representation Learning using \Ours{}} In order to successfully predict intermediate frames, it is essential for \Ours{} to accurately reason about motion trajectories, estimate and capture motion patterns specific to objects, and reconstruct both high level semantic detail and low level texture details. It is interesting to understand what types of motion information the networks learned and which tasks this representation is useful for. Therefore, we examine the possibility of using video frame interpolation in the context of unsupervised representation learning by pre-training \Ours{} on the task of frame interpolation, and reusing the learned feature representations for the tasks of action recognition, optical flow estimation, and motion magnification. This objective serves the dual purpose of providing insights into the nature of representations learnt during training frame interpolation models, while also improving the performance of downstream tasks compared to random initialization.

\section{Experimental Setup}

{\bf Datasets}. 
We use \textit{septuplets} from the Vimeo-90K dataset~\cite{xue2019video} extracted from 30FPS videos for training single frame interpolation networks ($k{=}2$). We train our model on the train split and evaluate it on the test split of the dataset. Following \cite{xu2019quadratic}, we additionally verify the \textit{generalization} capability of our proposed approach. For single frame interpolation, we report the performance of a model trained on \textit{Vimeo-90K} on the 100 quintuples generated from UCF101\cite{UCF101} and 2,847 quintuples generated from DAVIS dataset~\cite{perazzi2016benchmark}. For multi frame interpolation, we use GoPro~\cite{nah2017deep} as the training set, and report results on the Adobe dataset~\cite{su2017deep} and GoPro dataset\cite{nah2017deep} for \eightx{} interpolation. 

{\bf Training Details}. 
We use a R3D-18 backbone as the standard encoder in \Ours{}. We also evaluate different variants of 3D CNNs with group conv~\cite{Tran19} as backbones to achieve the best accuracy/speed trade-off. 
For data augmentation, we exploit the symmetry of the problem by randomly selecting input sequences during training and inverting the temporal order of the frames. Also, we also horizontally flip all frames of randomly selected inputs. Our hyper-parameter choices and more training details are provided in \emph{supplementary}.



\begin{table*}[!t]
  \centering
  \resizebox{0.95\textwidth}{!}{
  \begin{tabular}{@{} l *{10}{c} @{}} 
    \toprule  \\[-1em]
     \multirow{2}{*}{Method} &
     \multirow{2}{*}{Inputs} &
     \multicolumn{2}{c}{Vimeo-90K}  && \multicolumn{2}{c}{UCF101 } && \multicolumn{2}{c}{DAVIS}\\ 
     \cmidrule{3-4} \cmidrule{6-7} \cmidrule{9-10}
                                            &                   &   PSNR ($\uparrow$) & SSIM($\uparrow$) &&          PSNR($\uparrow$) & SSIM($\uparrow$) &&          PSNR($\uparrow$) & SSIM($\uparrow$) \\
    \midrule
    DAIN~\cite{bao2019depth}            &   RGB+Depth+Flow  &   33.35 & 0.945    &&       31.64 & 0.957     &&   26.12& 0.870    \\
    QVI~\cite{xu2019quadratic}          &   RGB+Flow       &   35.15 & \underline{0.971}    &&      \underline{32.89}& \underline{0.970}     &&   \underline{27.17} & {\bf 0.874}     \\
    \midrule
    \midrule
    DVF~\cite{liu2017video}             &   RGB             &   27.27&0.893     &&      28.72&0.937     &&    22.13&0.800      \\
    SepConv~\cite{niklaus2017video}     &   RGB             &   33.60&0.944     &&      31.97&0.943     &&    26.21&0.857      \\
    CAIN~\cite{choi2020channel}         &   RGB             &   33.93& 0.964     &&      32.28&0.965     &&    26.46&0.856      \\
    SuperSloMo~\cite{jiang2018super}    &   RGB             &   32.90&0.957      &&      32.33&0.960     &&    25.65&0.857      \\
    \red{BMBC}~\cite{park2020bmbc} & RGB & 34.76& 0.965 && {32.61}&0.955 && 26.42 & \underline{0.868} \\
    \red{AdaCoF}~\cite{lee2020adacof} & RGB & \underline{35.40}&\underline{0.971} && {32.71}&0.969 && 26.49&0.866 \\
    \Ours        &   RGB             &  \textbf{ 36.25}$^{\pm 0.06}$&\textbf{0.975}      &&      \textbf{33.31}$^{\pm 0.02}$ & {\bf 0.971}     &&    {\bf 27.43}$^{\pm 0.02}$& \textbf{0.874  }   \\
    \bottomrule \\
  \end{tabular}
  }
\vspace{-14pt}
    \caption{{\bf Comparison with state-of-the-art methods for 2x interpolation} on Vimeo-90K, UCF101, and DAVIS datasets. The upper table includes the methods that use additional networks trained to predict optical flows and/or depth maps. The lower table represents the methods the use only RGB as input. The first and second best methods are marked in \textbf{bold} and \underline{underlined} text. Our method consistently outperforms prior works which take only RGB as input, as well as works which additionally require optical flows and/or depth inputs.}
    \label{tab:twox_interpolation}
\vspace{-8pt}
\end{table*}


{\bf Evaluation Metrics}. Following previous works, we use PSNR and SSIM metrics to report the quantitative results of our method. For multi-frame interpolation we report the average value of the metric over all the predicted frames, and also additionally report the TCC (Temporal Change Consistency)~\cite{chi2020all}. Since these quantitative measures do not strongly correlate with the human visual system~\cite{nilsson2020understanding}, we also conduct a user study to analyze and compare our generated videos with other competing approaches. 

{\bf Baselines}.
We perform comparisons with the following baselines that perform single and multi frame video interpolation.
\begin{enumerate*}[label=(\roman*)]
  \item \textbf{DAIN~\cite{bao2019depth}} performs depth aware frame interpolation. 
  \item \textbf{QVI~\cite{xu2019quadratic}} computes quadratic flow prediction and adaptive filtering.
  \item \textbf{DVF~\cite{liu2017video}} uses volumetric sampling to generate the output frames.
  \item \textbf{SepConv~\cite{niklaus2017video}} predicts optimum pairs of spatially varying kernels for generating frames using input resampling. 
  \item \textbf{SuperSloMo~\cite{jiang2018super}} performs warping based on flow and visibility maps.
  \item \textbf{CAIN~\cite{choi2020channel}} performs frame interpolation by channel attention and sub-pixel convolutions, 
  \item \textbf{AdaCoF~\cite{lee2020adacof}} uses adaptive collaboration of flows, and
  \item \textbf{\Ours{}} is our proposed approach. We could not compare against recent works like SoftSplat~\cite{niklaus2020softmax}, AAO~\cite{chi2020all} and RRPN~\cite{zhang2020flexible} as their training code is not available online for retraining on our setting.
\end{enumerate*}

\begin{table}
\centering
\begin{minipage}[t]{\linewidth}{
\centering
\captionsetup{width=\textwidth}
\resizebox{\textwidth}{!}{
  \begin{tabular}{@{} l *{6}{c} @{}} 
    \toprule  \\[-1em]
     \multirow{3}{*}{Method} &
     \multirow{3}{*}{Inputs} &
     \multicolumn{2}{c}{Adobe}  && \multicolumn{2}{c}{GoPro} \\
     \cmidrule{3-4} \cmidrule{6-7}
                                        &                   &   PSNR & SSIM &&  PSNR & SSIM \\
    \midrule
    DAIN~\cite{bao2019depth}            &   RGB+Depth+Flow  &  29.50&0.910  && 29&0.91    \\
    QVI~\cite{xu2019quadratic}          &   RGB+Flow        &  \textbf{33.68}&\textbf{0.97 } && \underline{30.55} & \underline{0.933}     \\
    \midrule
    \midrule
    DVF~\cite{liu2017video}             &   RGB             &  28.23 & 0.896 && 21.94&0.776      \\
    SuperSloMo~\cite{jiang2018super}    &   RGB             &  30.66 & 0.391 && 28.52&0.891     \\
    \Ours{}                           &   RGB             &   \underline{32.20} & \underline{0.957}     && {\bf 31.31}& {\bf 0.94}    \\
    \bottomrule 
  \end{tabular}}
  \vspace{-8pt}
    \caption{{\bf Comparison with state-of-the-art methods for 8x interpolation} on Adobe and GoPro datasets. FLAVR outperforms all previous work that use only RGB as input.}
  \label{tab:eightx_interpolation}
  \vspace{-16pt}
}
\end{minipage}
\hfill
\end{table}

{\bf Comparison across baseline models.} \red{We note that each of these prior works report their numbers using a different training and testing setup in their respective papers, so the numbers differ among various works. For example, DAIN~\cite{bao2019depth} and AdaCoF~\cite{lee2020adacof} train and test on {\em triplet-split} of Vimeo-90K while SuperSloMo~\cite{jiang2018super} and QVI~\cite{xu2019quadratic} train their models on private custom datasets. To ensure fairness and a unified evaluation testbed, we accounted for all these variations by \textit{retraining baseline models} for \cite{bao2019depth, liu2017video, xu2019quadratic, jiang2018super, choi2020channel, lee2020adacof} till convergence on {\em septuplet-split} of Vimeo for comparison in \tabref{tab:twox_interpolation}. Likewise, in \tabref{tab:eightx_interpolation}, we retrained the presented baselines on GoPro data for fair comparison.   }
\vspace{-4pt}


\begin{figure*}
     \centering
     \begin{subfigure}[t]{0.32\textwidth}
        \centering
        \includegraphics[width=\textwidth]{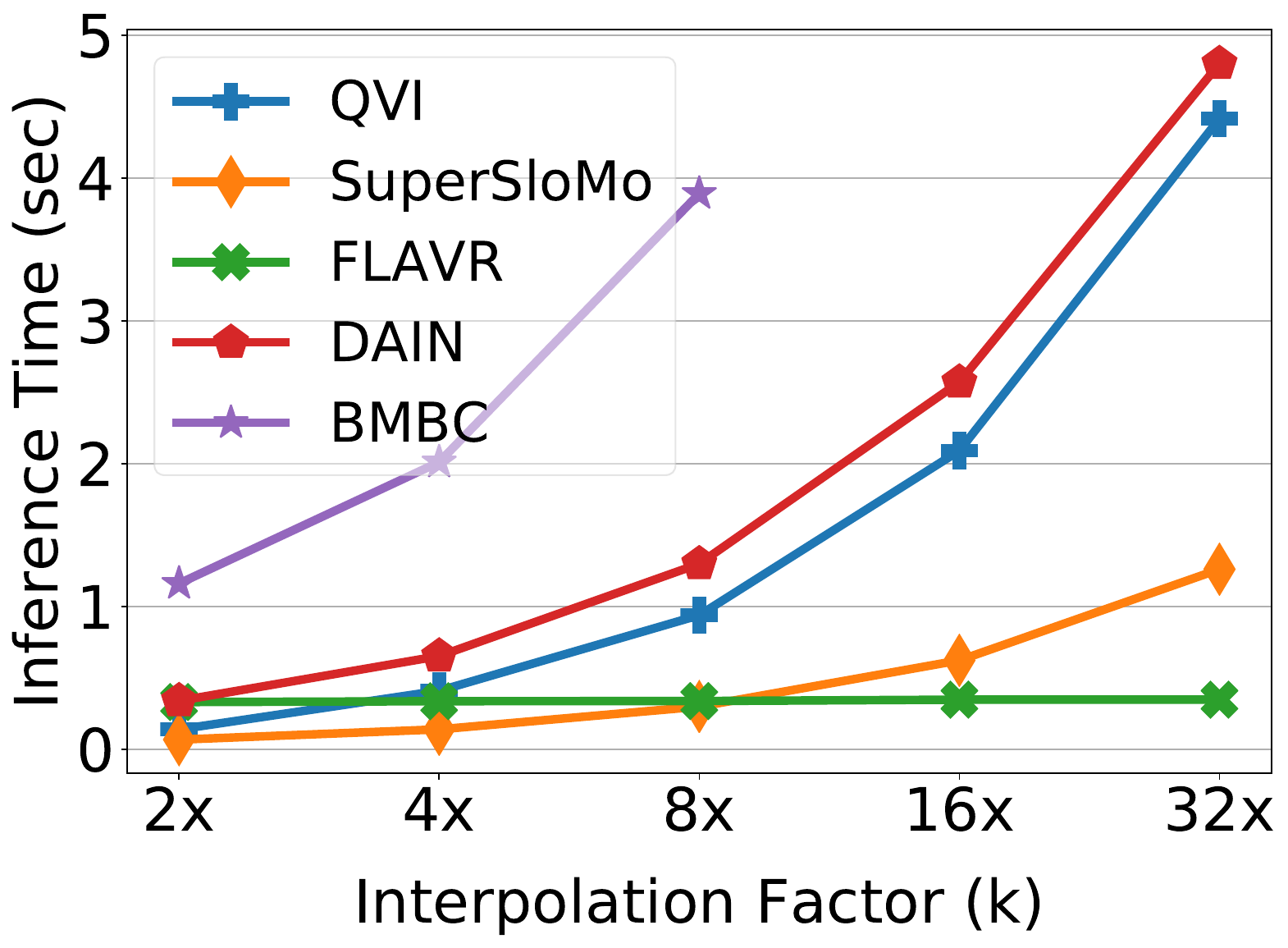}
        \caption{Inference Time vs. Interpolation Factor}
        \label{fig:TimeVsK}
     \end{subfigure}
     \hfill
     \begin{subfigure}[t]{0.32\textwidth}
        \centering
        \includegraphics[width=\textwidth]{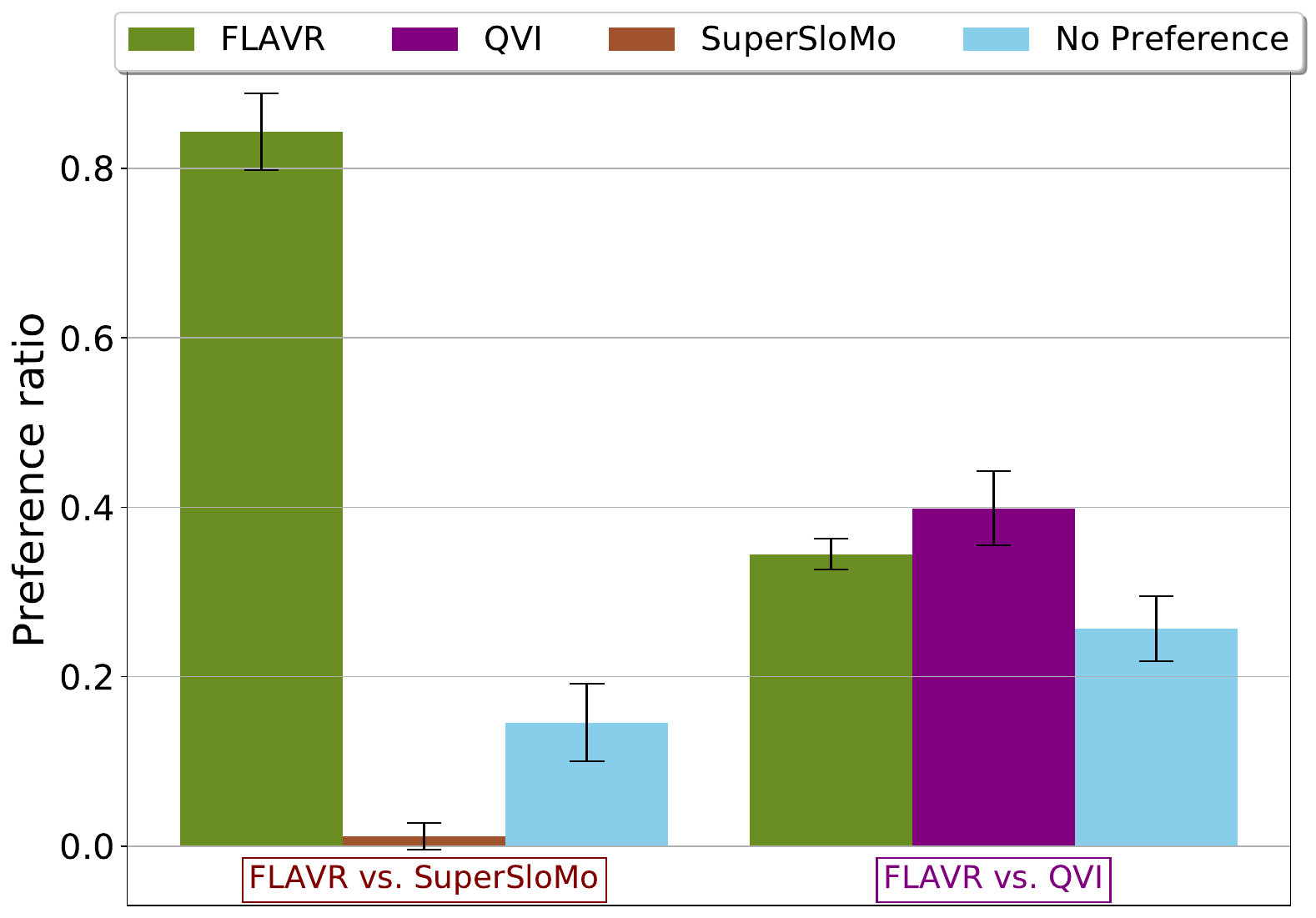}
        \caption{User Study Results}
        \label{fig:userStudy}
     \end{subfigure}
     \hfill
     \begin{subfigure}[t]{0.32\textwidth}
        \centering
        \includegraphics[width=\textwidth]{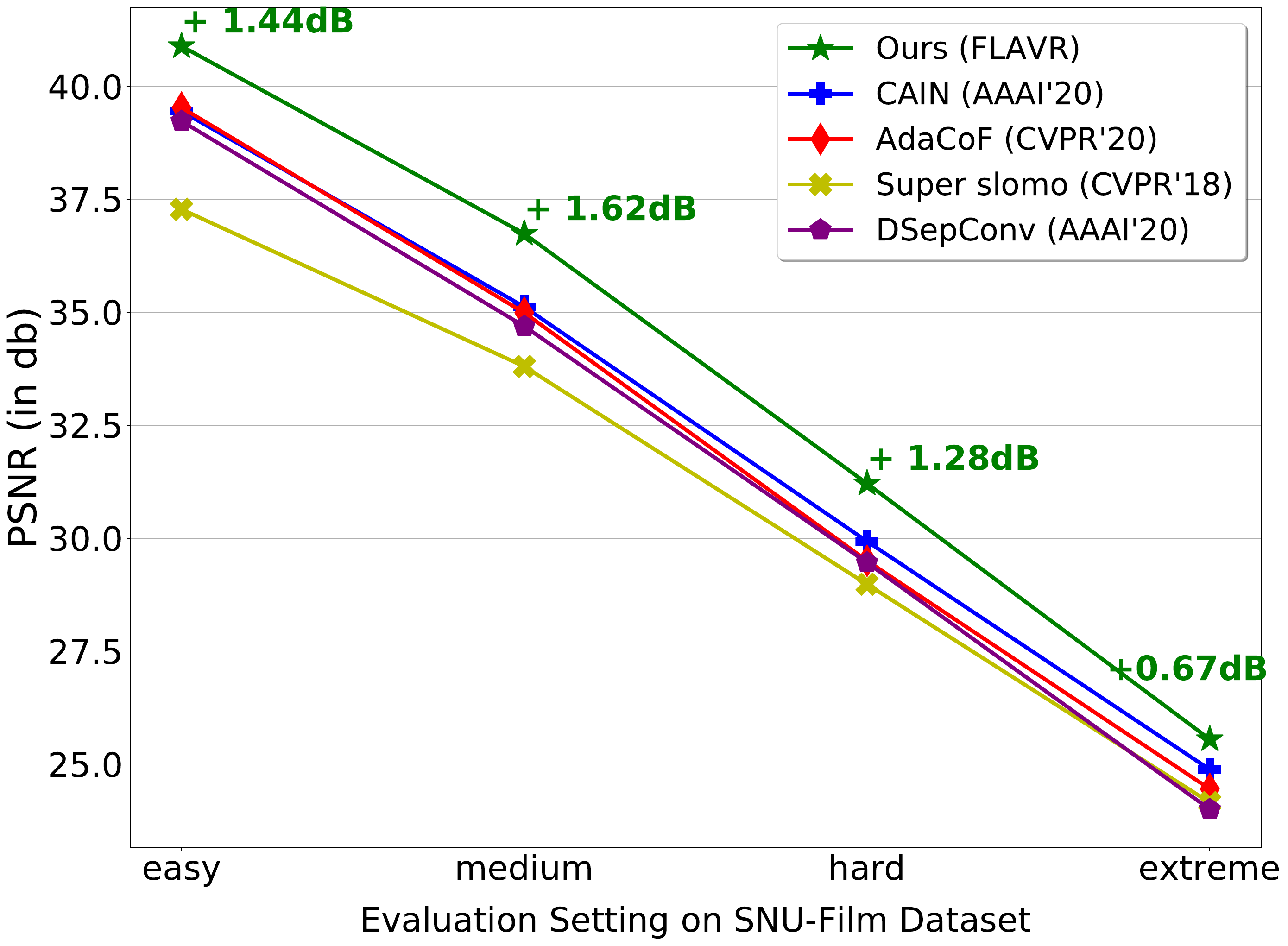}
        \captionsetup{width=\textwidth}
        \caption{Performance vs. Task Difficulty}
        \label{fig:SNUTest}
     \end{subfigure}
\vspace{-2pt}
\caption{{\bf Analysis}. (a) Inference time (forward pass w/o IO) of different methods on different interpolation factor. \Ours{} has almost no change in inference time due to its design to predict multiple frames per inference. (b) Comparison between \Ours{} with Super-SloMo and QVI in a user study on DAVIS. \Ours{} significantly outperforms Super-SloMo, and performs comparable to QVI. (c) Comparison between \Ours{} with other methods on SNU-Film dataset. \Ours{} consistently outperforms all comparing methods across all levels of task difficulty.}
\vspace{-12pt}
\label{fig:analysis}
\end{figure*}

\section{How does \Ours{} compare with the state-of-the-art?}
\label{sec:state-of-the-art}

{\bf Single-Frame Interpolation.} We report the results for single frame interpolation in \tabref{tab:twox_interpolation}, corresponding to \twox{}($k{=}2$) interpolation from 15 FPS to 30 FPS. We observe that \Ours{} outperforms prior methods by a significant margin on Vimeo-90K dataset and sets the \textit{new state-of-the-art} on this dataset with a PSNR value of 
$36.25$ and SSIM value of $0.975$. \Ours{} is a more generally applicable method and outperforms~\cite{jiang2018super, liu2017video, niklaus2017video} which assume uniform linear motion between the frames. 
\Ours{} also performs better than~\cite{choi2020channel} which uses a similar end to end architecture to predict output frames, underlining the benefits achieved using an encoder-decoder architecture with spatio-temporal kernels. More importantly, \Ours{} also outperforms DAIN~\cite{bao2019depth} and QVI~\cite{xu2019quadratic} without demanding additional knowledge in the form of bidirectional flow or depth maps.

We test the generalization capability of our method by evaluating the same trained model on UCF101 and DAVIS datasets. These are relatively more challenging for video frame interpolation, containing complex object and human motions from a range of dynamic scenes. Nevertheless, with a PSNR of $33.33$ on the UCF101 dataset and $27.44$ on the DAVIS dataset, \Ours{} clearly delivers better performance compared to all the baselines methods which take RGB images as inputs, and performs on par or better than methods that additionally demand depth or flow maps as inputs. These datasets together constitute a wide spectrum of difficulty in terms of complex motions and occlusions, and \Ours{} outperforms other methods on all the settings. 

{\bf Multi-Frame Interpolation.} 
For multi-frame setting, we report results on \eightx{} ($k{=}8$) interpolation in \tabref{tab:eightx_interpolation}, which corresponds to going from $30$ to $240$ FPS by generating $7$ intermediate frames. Our method yields a PSNR of $31.31$ and an SSIM score of $0.94$ on the GoPro dataset, which is better than all the prior approaches proposed for frame interpolation. On the Adobe dataset, our method outperforms all methods significantly except QVI, but QVI additionally uses an optical flow estimator which helps on the more challenging Adobe dataset. Additionally, we evaluate TCC~\cite{chi2020all} on GoPro to obtain {0.78, 0.76, 0.73} for {FLAVR, QVI, DAIN} respectively. It is evident that FLAVR outperforms those prior works. AOO~\cite{chi2020all} reports 0.83, but it is trained on custom data and uses GAN loss, which is biased in favor of this metric (and GAN loss is complementary to FLAVR and other VFI methods).
Similar improvements in performance can also be observed in the case of \fourx{} ($k{=}4$) interpolation, as shown in the \red{supplementary material}. Additionally, we show qualitative results by using \Ours{} on few sequences from DAVIS dataset in \figref{fig:qualt_figs}. 
These results indicate the effectiveness of the proposed \Ours{} architecture for the task of multi-frame interpolation.

\red{ {\bf Results on Middleburry} We evaluate FLAVR on the publicly available test images from Middleburry \cite{baker2011database,scharstein2014high} dataset on the task of single frame interpolation. \Ours{} is ranked {\bf \nth{2}, \nth{5}, \nth{8}} on {\em backyard, evergreen, basketball} sequences respectively, at the time of this submission. The complete results are available on the public leaderboard (\href{https://vision.middlebury.edu/flow/eval/results/results-i1.php}{link}). 
Qualitative comparisons with other approaches on Middleburry images are provided with the \emph{supplementary} material.}

{\bf Speed vs. Accuracy Trade-off.}
One major challenge for realizing the applications of video frame interpolation for real time applications on low resource hardware is to optimize the trade off between faster inference speed and better interpolation quality. 
Perhaps the most important contribution of our work is to propose an approach that strikes an optimum balance between both these factors by achieving best performance with smallest runtime. \red{As shown in \figref{fig:intro_pic}, FLAVR offers an improved run time for multi-frame interpolation models.
This improvement is possible mainly because we require no overhead in terms of computing optical flow or depth, and predict all the frames in a single forward pass. We also show in \figref{fig:TimeVsK} that the inference speed using our method scales gracefully with an increase in the interpolation factor $k$, while most prior methods incur linear growth with $k$. We achieve runtime improvements %
of $2.7\stimes, 6.2\stimes$ and $12.7\stimes$ for \eightx{}, $16\stimes$ and $32\stimes$ interpolation respectively with respect to QVI, which is the current most accurate method, while providing much higher interpolation accuracy compared to SuperSlomo, which is the current fastest.} 

\begin{figure}[t]
     \centering
     \begin{subfigure}[t]{0.23\textwidth}
         \centering
         \includegraphics[width=\textwidth]{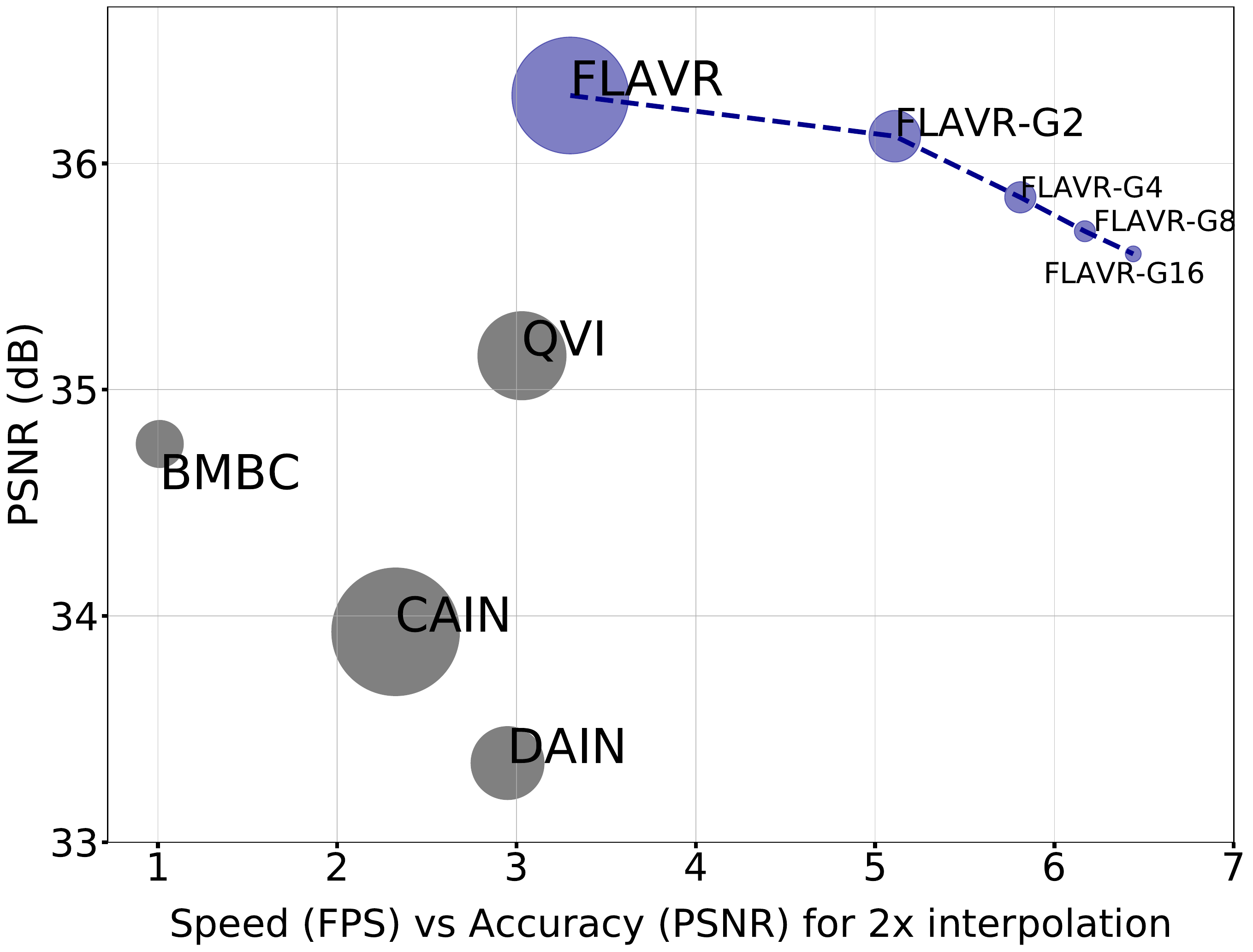}
         \subcaption{\twox{} interpolation}
         \label{fig:tradeoff2x}
     \end{subfigure}
     \begin{subfigure}[t]{0.23\textwidth}
         \centering
         \includegraphics[width=\textwidth]{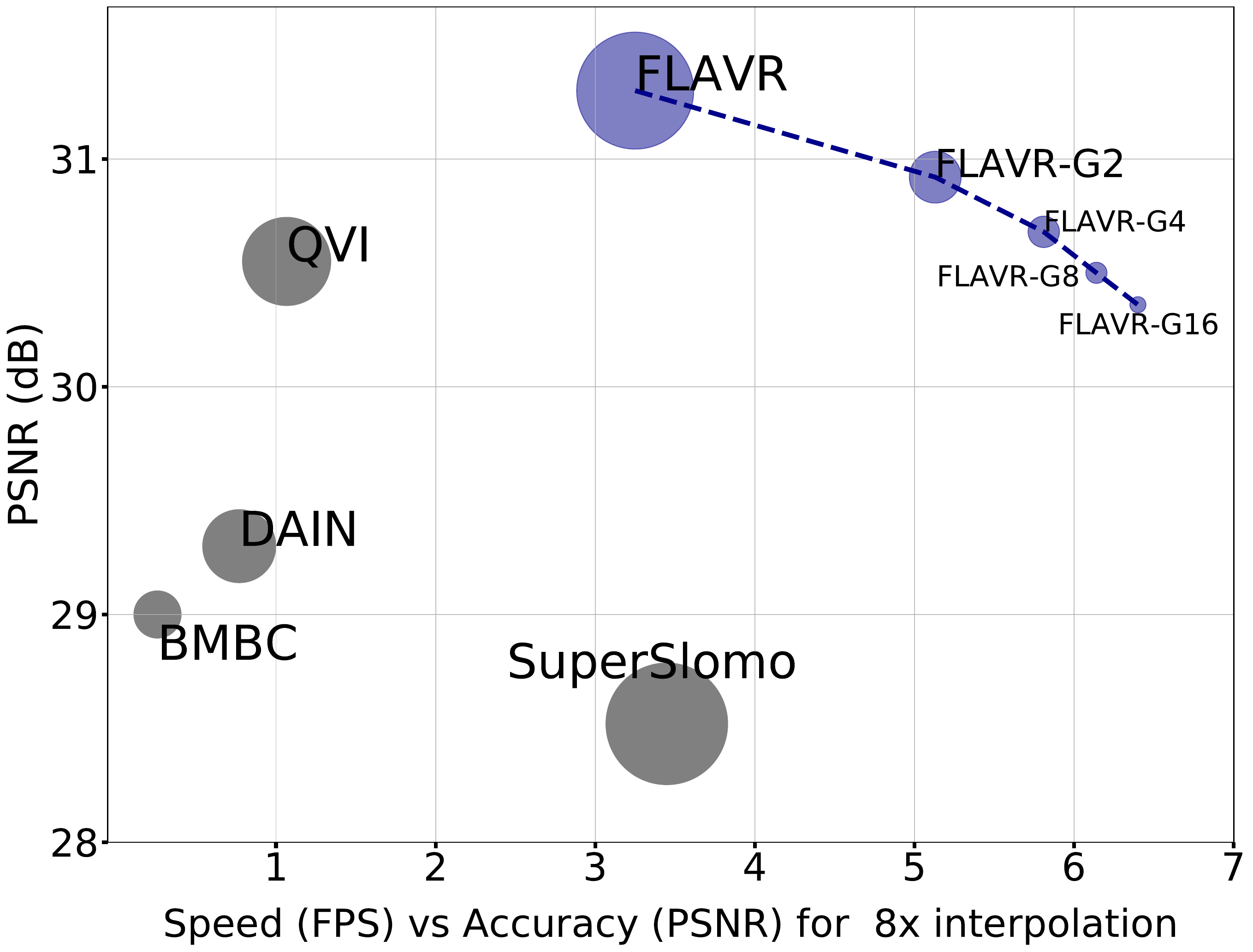}
         \subcaption{\eightx{} interpolation}
         \label{fig:tradeoff8x}
     \end{subfigure}
     \vspace{-8pt}
    \caption{{\bf Speed, accuracy and parameter tradeoff comparison.} Speed (in FPS, on x-axis) vs. accuracy (in PSNR, on y-axis) for various baselines as well as various architecture choices for \Ours{}. Number of parameters in each model is proportional to the size of the marker. \subref{fig:tradeoff2x} is for \twox{} and \subref{fig:tradeoff8x} is for \eightx{} interpolation. \texttt{FLAVR-Gx} corresponds to \Ours{} with \texttt{x} number of group convolutions. In summary, \Ours{} achieves best speed-accuracy tradeoffs compared to many recent methods.
    }
    \label{fig:SpeedVsAccuracy}
    \vspace{-16pt}
\end{figure}

\new{ We also perform an in-depth ablation on the effect of using group convolutions \cite{Tran19} on the speed-accuracy trade-offs on \Ours{}, and showcase results in \figref{fig:SpeedVsAccuracy}. Specifically, for every 3D conv block, we replace the residual block by a channel separated convolution block \cite{Tran19} with groups $g=1,2,4,8 \text{ and } 16$, indicated by \texttt{FLAVR}, \texttt{FLAVR-2x}, \texttt{FLAVR-4x} and so on in \figref{fig:SpeedVsAccuracy}. Note that $g=1$ refers to our default setting in all other experiments. We show the results on Vimeo-90K for \twox{} interpolation as well as GoPro dataset on \eightx{} interpolation. We find that compared to baselines that deliver similar performance (eg. QVI), \emph{\Ours{} is at least 6$\stimes$ faster on \eightx interpolation} (refer \figref{fig:tradeoff8x}, \texttt{FLAVR-G8} vs. \texttt{QVI}). Furthermore, compared to baselines that give similar inference time speeds, \emph{\Ours{} delivers at least 3dB accuracy gain} (refer \figref{fig:tradeoff8x}, \texttt{FLAVR} vs. \texttt{SuperSloMo}). These results indicate that \Ours{} is a flexible architecture achieving best speed accuracy trade-off for video frame interpolation compared to existing methods. }

\begin{table*}[h]
\centering
\begin{subtable}[t]{.21\textwidth}
\resizebox{\textwidth}{!}{
\begin{tabular}{lcc}
\hline
Model & PSNR & SSIM \\
\midrule
R2D-18-2I & 33.98 &	0.966 \\
R2D-18-4I &	34.97 &	0.967 \\
R3D-18-4I &	\textbf{36.3} &	\textbf{0.975} \\
\bottomrule
\end{tabular}}
\captionsetup{width=.9\textwidth}
\subcaption{Effect of encoder arch.}
\label{tab:backbone}
\end{subtable}
\hfill
\begin{subtable}[t]{.23\textwidth}
\resizebox{\textwidth}{!}{
\begin{tabular}{l|cc}
\hline
Model & PSNR & SSIM \\
\midrule
No fusion &	35.1 &	0.9713 \\
fusion - add &	35.7 &	0.9737\\
fusion - concat &	\textbf{36.3} &	\textbf{0.975}\\
\bottomrule
\end{tabular}}
\captionsetup{width=.9\textwidth}
\subcaption{Type of feature fusion}
\label{tab:concat}
\end{subtable}
\hfill
\begin{subtable}[t]{.23\textwidth}
\resizebox{\textwidth}{!}{
\begin{tabular}{l|cc}
\hline
Model & PSNR & SSIM \\
\midrule
w/o stride   &	\textbf{36.3}  &	\textbf{0.975} \\
w/ 2x stride &	35.4  &	0.961 \\
w/ 4x stride &	35.21 &	0.96 \\
\bottomrule
\end{tabular}}
\captionsetup{width=.9\textwidth}
\subcaption{Effect of temporal striding}
\label{tab:stride}
\end{subtable}
\hfill
\begin{subtable}[t]{.23\textwidth}
\resizebox{\textwidth}{!}{
\begin{tabular}{l|cc}
\hline
Model & PSNR & SSIM \\
\midrule
L1 Loss      &	\textbf{36.3}    &	\textbf{0.975} \\
L2 Loss       &	35.3    &	0.965 \\
Huber Loss    &	35.3    &	0.964 \\
L1+VGG Loss      &	35.91   &	0.962 \\
\bottomrule
\end{tabular}}
\captionsetup{width=.9\textwidth}
\subcaption{Effect of loss function}
\label{tab:lossFn}
\end{subtable}
\captionsetup{width=\textwidth}
\vspace{-8pt}
\caption{{\bf Ablation results} for \Ours{} architecture on \subref{tab:backbone} different backbones, \subref{tab:concat} fusion methods, \subref{tab:stride} temporal striding, and \subref{tab:lossFn} loss functions.}
\label{tab:ablations}
\vspace{-8pt}
\end{table*}

{\bf Robustness to Task Difficulty.}
We validate the robustness in performance of our method using the SNU-Film dataset~\cite{choi2020channel} consisting of videos with varying difficulty for interpolation depending on the temporal gap between the input frames.
The four settings we use are \textit{easy} (120-240 FPS), e.g. predicting 240 FPS video from 120 FPS input, \textit{medium} (60-120 FPS), \textit{hard} (30-60 FPS) and \textit{extreme} (15-30 FPS). In~\figref{fig:SNUTest}, we compare the performance of our method with prior works including CAIN~\cite{choi2020channel} and AdaCoF~\cite{lee2020adacof}, and report better performance than all the methods consistently across all the difficulty settings. Specifically, we see a gain of $1.28 dB$ and $1.62 dB$ compared to the next best approach \cite{choi2020channel} in the \textit{hard} and \textit{medium} settings respectively, which are considered challenging for video frame interpolation because of large motions and longer time gaps between the frames indicating robustness to video frame rates.
%
%
%
%


\begin{figure}[t]
    \begin{center}
    \begin{subfigure}[b]{0.09\textwidth}
        \centering
        \includegraphics[width=\textwidth]{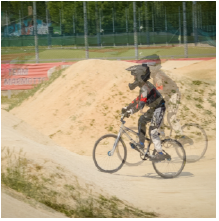}
    \end{subfigure}
    \hfill
    \begin{subfigure}[b]{0.09\textwidth}
        \centering
        \includegraphics[width=\textwidth]{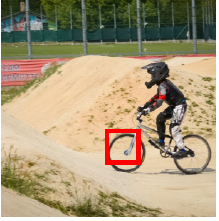}
    \end{subfigure}
    \hfill
    \begin{subfigure}[b]{0.09\textwidth}
        \centering
        \includegraphics[width=\textwidth]{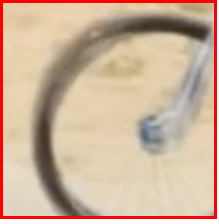}
    \end{subfigure}
    \hfill
    \begin{subfigure}[b]{0.09\textwidth}
        \centering
        \includegraphics[width=\textwidth]{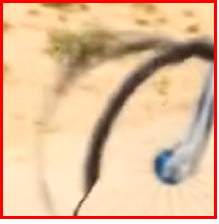}
    \end{subfigure}
    \hfill
    \begin{subfigure}[b]{0.09\textwidth}
        \centering
        \includegraphics[width=\textwidth]{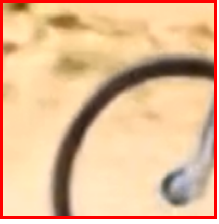}
    \end{subfigure}
    
    
    \begin{subfigure}[b]{0.09\textwidth}
        \centering
        \includegraphics[width=\textwidth]{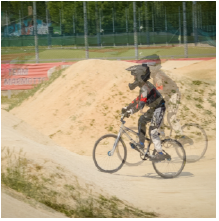}
    \end{subfigure}
    \hfill
    \begin{subfigure}[b]{0.09\textwidth}
        \centering
        \includegraphics[width=\textwidth]{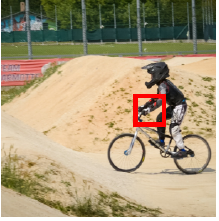}
    \end{subfigure}
    \hfill
    \begin{subfigure}[b]{0.09\textwidth}
        \centering
        \includegraphics[width=\textwidth]{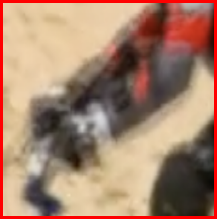}
    \end{subfigure}
    \hfill
    \begin{subfigure}[b]{0.09\textwidth}
        \centering
        \includegraphics[width=\textwidth]{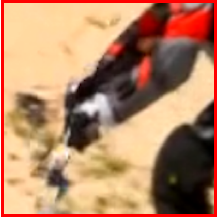}
    \end{subfigure}
    \hfill
    \begin{subfigure}[b]{0.09\textwidth}
        \centering
        \includegraphics[width=\textwidth]{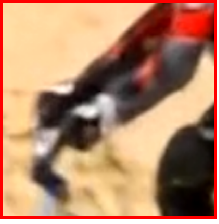}
    \end{subfigure}
    
    
    \begin{subfigure}[b]{0.09\textwidth}
        \centering
        \includegraphics[width=\textwidth]{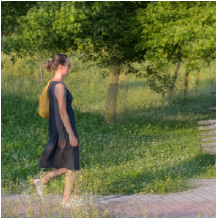}
        \captionsetup{width=\textwidth}
        \subcaption{Overlay}
    \end{subfigure}
    \hfill
    \begin{subfigure}[b]{0.09\textwidth}
        \centering
        \includegraphics[width=\textwidth]{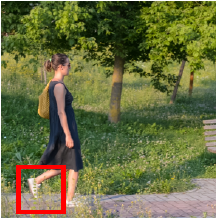}
        \captionsetup{width=\textwidth}
        \subcaption{GT}
    \end{subfigure}
    \hfill
    \begin{subfigure}[b]{0.09\textwidth}
        \centering
        \includegraphics[width=\textwidth]{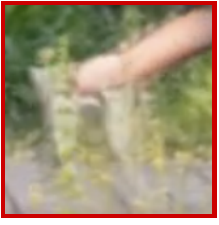}
        \subcaption{SSM~\cite{jiang2018super}}
    \end{subfigure}
    \hfill
    \begin{subfigure}[b]{0.09\textwidth}
        \centering
        \includegraphics[width=\textwidth]{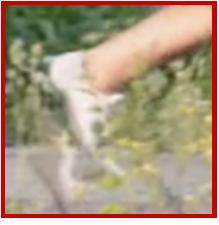}
        \captionsetup{width=\textwidth}
        \subcaption{QVI~\cite{xu2019quadratic}}
    \end{subfigure}
    \hfill
    \begin{subfigure}[b]{0.09\textwidth}
        \centering
        \includegraphics[width=\textwidth]{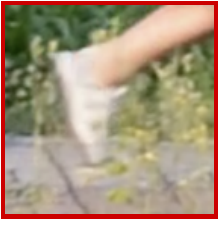}
        \captionsetup{width=\textwidth}
        \subcaption{\Ours{}}
    \end{subfigure}
    \end{center}
    \caption{{\bf Qualitative comparison with state-of-the-art methods}. We qualitatively compare FLAVR with Super-SloMo (SSM), QVI on a few video sequences on DAVIS. More qualitative results and generated videos are provided along with the \red{supplementary} material.}
    \label{fig:qualt_figs}
    \vspace{-12pt}
\end{figure}
\begin{table*}[!t]
    \begin{subtable}[b]{.35\textwidth} \centering
          
          \resizebox{0.98\textwidth}{!}{
        \begin{tabular}{l c c c cc}
            \toprule
            Method            &  pretrained on & Arch.        & UCF101        &&      HMDB51 \\
            \midrule
            \midrule
            Random Init.                                & -         & R3D-18                & 50.02 &&      19.00  \\
            Supervised & Kinetics-400                           &       R3D-18             &   87.70   &&      59.10 \\
            Contrastive~\cite{han2019video} & Kinetics-400              & R3D-18              &   68.20   &&      34.50  \\
            \midrule
            Video-GAN~\cite{vondrick2016generating} & UCF101    &     Custom          & 52.10 &&      -   \\
            LMD\cite{luo2017unsupervised}           & NTU RGB   &     Custom         & 53.00 &&      - \\
            DVF \cite{liu2017video} &  UCF101                   &   Custom            & 52.40 &&      - \\
            \Ours{} & Vimeo-90K                           &      R3D-18          & 63.10 &&      23.48 \\
            \bottomrule
        \end{tabular}}
        \caption{ {\bf Action recognition}.
        }
        \label{tab:actionUCF}
    \end{subtable}
    \hfill
   \begin{subtable}[b]{.32\textwidth} \centering
        \resizebox{0.98\textwidth}{!}{
        \begin{tabular}{l p{1.5cm} p{1.5cm} p{2cm} @{}}
            \toprule
            Dataset & FlowNet \cite{ilg2017flownet} & Random Init. & Finetune on \Ours{} \\
            \midrule
            MPI-Clean~\cite{Butler_ECCV_2012} & 2.02 & 4.41 & 2.92 \\
            MPI-Final~\cite{Butler_ECCV_2012} & 3.14 & 5.27 & 3.90 \\
            Kitti-12~\cite{Geiger2012CVPR} & 4.09 & 9.25  & 5.23 \\
            Kitti-15~\cite{Menze2015CVPR} & 10.06 & 17.22 & 13.68 \\
            \bottomrule
        \end{tabular}}
        \caption{{\bf Optical flow estimation}. }
        \label{tab:SSL_flow}
    \end{subtable}
    \hfill
       \begin{subtable}[b]{.28\textwidth} \centering
          \centering
          \resizebox{0.98\textwidth}{!}{
        \begin{tabular}{@{} l *{4}{c} @{}}
        \toprule
        \midrule
        & \multicolumn{3}{c}{\textbf{15 $\rightarrow$ 30 FPS}} \\
         \cline{2-4}
        &   $\mathcal{J}\&\mathcal{F}_{\text{m}}$ & $\mathcal{J}_{\text{m}}$ & $\mathcal{F}_{\text{m}}$ \\
        CRW & 65.5 & 62.8 & 68.2 \\
        FLAVR+CRW & \textbf{66.6} & \textbf{63.9} & \textbf{69.4} \\
        & \multicolumn{3}{c}{\textbf{8 $\rightarrow$ 30 FPS}} \\
         \cline{2-4}
        CRW & 61.9 & 59.3 & 64.5 \\
        FLAVR+CRW & \textbf{62.8} & \textbf{60.5} & \textbf{65.1} \\
        \bottomrule
        \end{tabular}}
        \captionsetup{width=\textwidth}
          \caption{{\bf Video object segmentation mask propagation}.
          \label{tab:tracking}
        }
        \end{subtable}
        \vspace{-8pt}
        \caption{{\bf \Ours{} for various downstream applications}. 
        \subref{tab:actionUCF} FLAVR as a self-supervised pretext task for action recognition on UCF101 and HMDB51. 
        \subref{tab:SSL_flow} for optical flow prediction on MPI (Sintel~\cite{Butler_ECCV_2012}) and Kitti~\cite{Menze2015CVPR} datasets.
        \subref{tab:tracking} for video object segmentation mask propagation for low fps DAVIS videos. $\mathcal{J}_m$ measures the region similarity as mean IoU, while $\mathcal{F}_m$ is a boundary alignment metric. 
        }
        \vspace{-8pt}
 \end{table*}

\begin{figure*}[t]
    \begin{center}
    \begin{subfigure}[c]{0.45\textwidth}
        \centering
        \includegraphics[width=\textwidth]{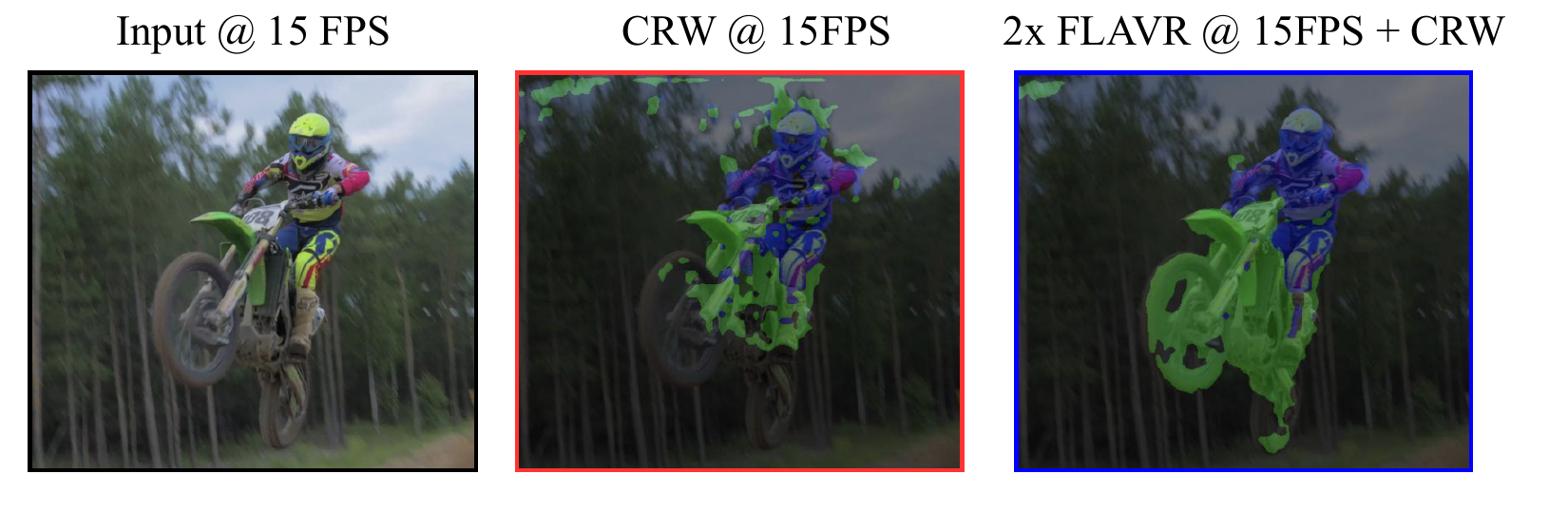}
    \end{subfigure}
    \hspace{12pt}
    \begin{subfigure}[c]{0.45\textwidth}
        \centering
        \includegraphics[width=\textwidth]{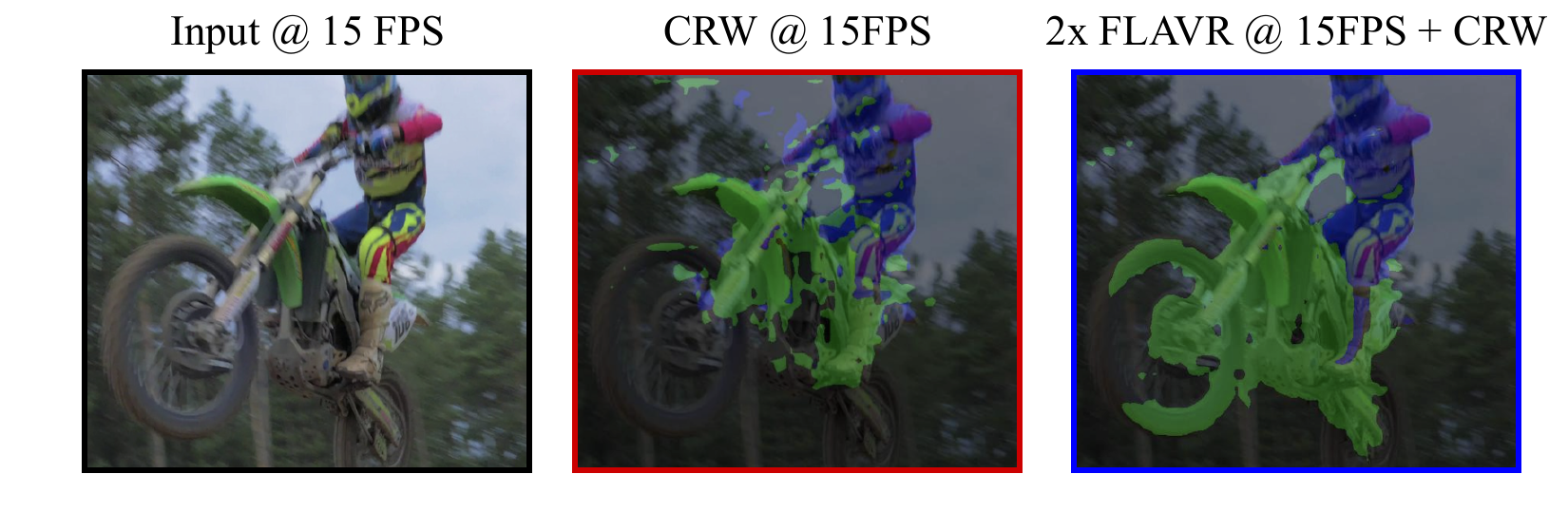}
    \end{subfigure}
    
    \end{center}
    \vspace{-16pt}
    \captionsetup{width=\textwidth, font=footnotesize}
    \caption{{\bf Video object segmentation mask propagation on DAVIS}. \Ours{} helps to improve video object tracking in low fps videos. \Ours{} is first used to up-sample video into higher frame rate, then a standard object segment propagation, e.g., CRW~\cite{jabri2020space}, is applied on interpolated videos. Refer \tabref{tab:tracking} for quantitative improvements.}
    \label{fig:downstream}
    \vspace{-16pt}
\end{figure*}


{\bf User Study.} 
We conduct a user survey on Amazon Mechanical Turk to analyze the performance of our method in comparison to \cite{xu2019quadratic} and \cite{jiang2018super} on the 90HD videos from Davis dataset for \eightx{} interpolation. 
We explain the details of the survey in \emph{supplementary}, and summarize the results in \figref{fig:userStudy}.
Firstly, when the comparison is between our method against Super-SloMo, users overwhelmingly preferred our videos as the generated videos looked more realistic with minimum artefacts around edges and motion boundaries owing to accurate interpolation. In comparison with QVI, users choose \Ours{} in 35\% of videos compared to QVI, which was chosen in 40\% of the videos; and for 20\% of videos the differences came out to be negligible. These results further support our hypothesis that in the interest of real world deployment, optical flow and warping based frame interpolation methods can be substituted with our learning based approach that offers faster inference (\figref{fig:TimeVsK}) with minimal loss in performance.

{\bf Ablations.} \new{  We provide detailed ablation into various design choices of the architecture, network and loss functions on the Vimeo-90K dataset in \tabref{tab:ablations}, and enlist the salient observations here. Firstly, we find that compared to an encoder with 2D Resnet-18 which takes a channel-wise concatenation of 4 images, \Ours{} gives a $1.3$dB gain on PSNR (\tabref{tab:backbone}) validating our choice of spatio temporal network. Also, we find that using no striding in the temporal dimension ($36.3$dB) performs better than using stride of 2 ($35.4$dB) or 4 ($35.21$dB), supporting the hypothesis that temporal striding hurts in capturing sharp pixel level detail (\tabref{tab:stride}). Likewise, we observe that adding VGG-based perception loss \cite{johnson2016perceptual} to the L1 losses during training is inferior in terms of PSNR (\tabref{tab:lossFn}). We include additional results on the effects of channel gating along with supporting qualitative results with the \red{supplementary material}.
}

\section{How useful is \Ours{} in enabling downstream applications?}
\label{sec:SSL}

{\bf Action Recognition} We evaluate the semantic properties of the internal representations learned by \Ours{} by reusing its trained encoder on a downstream action recognition task. We remove the decoder and attach a classification layer to the network. The whole network is then finetuned end to end on UCF101 and HMDB51 datasets. In order to isolate the benefits arising from pretraining the encoder on video interpolation task, we train a 3D resnet (R3D) baseline completely from scratch and observe from \tabref{tab:actionUCF} that \Ours{}, which is pretrained on Vimeo-90K dataset on frame interpolation task clearly outperforms random initialization baseline by $13.08\%$ on UCF-101 and $4.48\%$ on HMDB-51. \Ours{} also significantly outperforms prior self-supervised methods on video which use low level pretext tasks like Video-GAN~\cite{vondrick2016generating} and flow descriptors~\cite{luo2017unsupervised} indicating that frame interpolation can learn useful motion representations. 
Finally, \Ours{} also achieves better accuracy than pretraining using DVF~\cite{liu2017video} which underlines that our particular method for video frame interpolation invariably benefits downstream action recognition more than voxel flow.

{\bf Optical Flow Estimation} It is known that successful frame interpolation intrinsically depends on reliable optical flow estimation~\cite{wulff2018temporal}. We investigate this hypothesis by finetuning our trained network for optical flow estimation on MPI Sintel~\cite{Butler_ECCV_2012} and Kitti~\cite{Geiger2012CVPR, Menze2015CVPR} datasets, and report the corresponding EPE (end point error) in \tabref{tab:SSL_flow}. Finetuning using \Ours{} achieves much lower EPE compared with random initialization using the same backbone architecture, proving that our model learns useful flow features. We note that we do not aim to outperform more complex, flow-dedicated architectures\cite{ilg2017flownet, liu2019selflow} but aim to understand if we can learn useful flow features using a simple architecture like ours by pre-training on frame interpolation.

{\bf \Ours{} improves VOS at low fps} \red{ So far we evaluated \Ours{}'s representation quality for downstream task but how good is its raw output in improving downstream applications? To study this, we consider the task of video object segmentation label propagation where the task is to propagate object masks throughout the video by extracting visual correspondences~\cite{jabri2020space, wang2019learning, xu2021rethinking, li2019joint}. Most of current approaches which perform label propagation assume access to 30FPS videos during training and testing (for example, from DAVIS), but the ability to find correspondences, and hence the accuracy of label propagation, falls considerably if the inputs are from low fps videos. We show that in such cases, \Ours{} can be used to improve the accuracy of video object segmentation (VOS). To this end, we subsample the test videos from DAVIS dataset by \twox{} (30FPS $\rightarrow$ 15FPS) and \fourx{} (30FPS $\rightarrow$ 8FPS) factors, and then apply the label propagation algorithm proposed in CRW~\cite{jabri2020space}. Additionally, we also apply \Ours{} for frame interpolation with $k=2,4$ to recover the original 30FPS videos in each case respectively, and apply the CRW algorithm again on the upsampled videos. From \tabref{tab:tracking} and \figref{fig:downstream}, we observe that \Ours{} can be effectively used as an intermediate step to improve the results of label propagation on low fps videos. More details regarding the experiment are present in the \em{supplementary}.}

{\bf Motion Magnification} Motion magnification~\cite{wu2012eulerian, oh2018learning} is a complementary problem to frame interpolation, in which the task is to magnify the subtle motions from the input video. Using \Ours{} as pre-training, we fine-tune a motion magnification network for a fixed magnification factor of $10$. 
\Ours{} achieves an SSIM of $0.801$ on the synthetic CoCo-Synth dataset~\cite{oh2018learning} with a simple architecture.
More details and qualitative results are presented in {\em the supplementary}.

\vspace{-5pt}
\section{Discussion}
\label{sec:conclusion}

We present \Ours{} for flow-free and end-to-end video frame interpolation. \Ours{} uses 3D convolutions to model the spatio-temporal relations between the input frames improving interpolation accuracy under challenging motion profiles across various input frame rates. In extensive experiments and analysis presented across the main paper and the supplementary, we show that \Ours{} offers best trade-off in terms of inference speed vs. interpolation accuracy compared to many existing approaches. We show that the representations learned by \Ours{} are useful for various downstream tasks such as action recognition, optical flow estimation, video object segmentation mask propagation and motion magnification. We also invite the reviewers to look at more qualitative results and generated videos which are provided along with the \textbf{supplementary material}.

\Ours{} still requires retraining for each interpolation factor $k$ although for most of the practical applications, it is well known beforehand what would be the desired interpolation factor. Being a data-driven end-to-end approach, \Ours{} shares with other deep learning based approaches the limited generalization capability to data outside the training distribution. Nevertheless, we expect \Ours{} to stimulate new directions for frame interpolation with ample opportunity for simpler and efficient methods to address these limitations.


\section{ Ablations}

In \tabref{tab:ablations}, we present a detailed ablation study of the proposed architecture design in terms of the skip connections, strides and loss functions. In addition to the brief insight provided in the main text, we explain each of them in detail next. We conduct all the ablation studies on the Vimeo-90K dataset.

{\bf Backbone Architecture} In this work, we propose using 3D convolutions that model space-time relations for improved frame interpolation. To verify this hypothesis, we train a video interpolation network using 2D convolutions instead, and present the results in \tabref{tab:backbone}. While training 2D Resnet, we concatenate RGB channels of the input before feeding into the network. We observe that the \textit{R2D-18-2I} baseline, which uses a 2D ResNet-18 encoder decoder with $2$ input frames ($C=1$) performs worse than \textit{2D-R18-4I} baseline, which uses 4 input frames ($C=2$) justifying the need for a larger input context. Next, our proposed architecture \textit{3D-R18-4I} which uses 3D convolutions along with 4 inputs, clearly outperforms both these baselines by $1.3$ and $2.3$dB, respectively. This indicates the importance of temporal modeling for the task. 

\begin{figure}
    \begin{center}
    \includegraphics[width=0.35\textwidth]{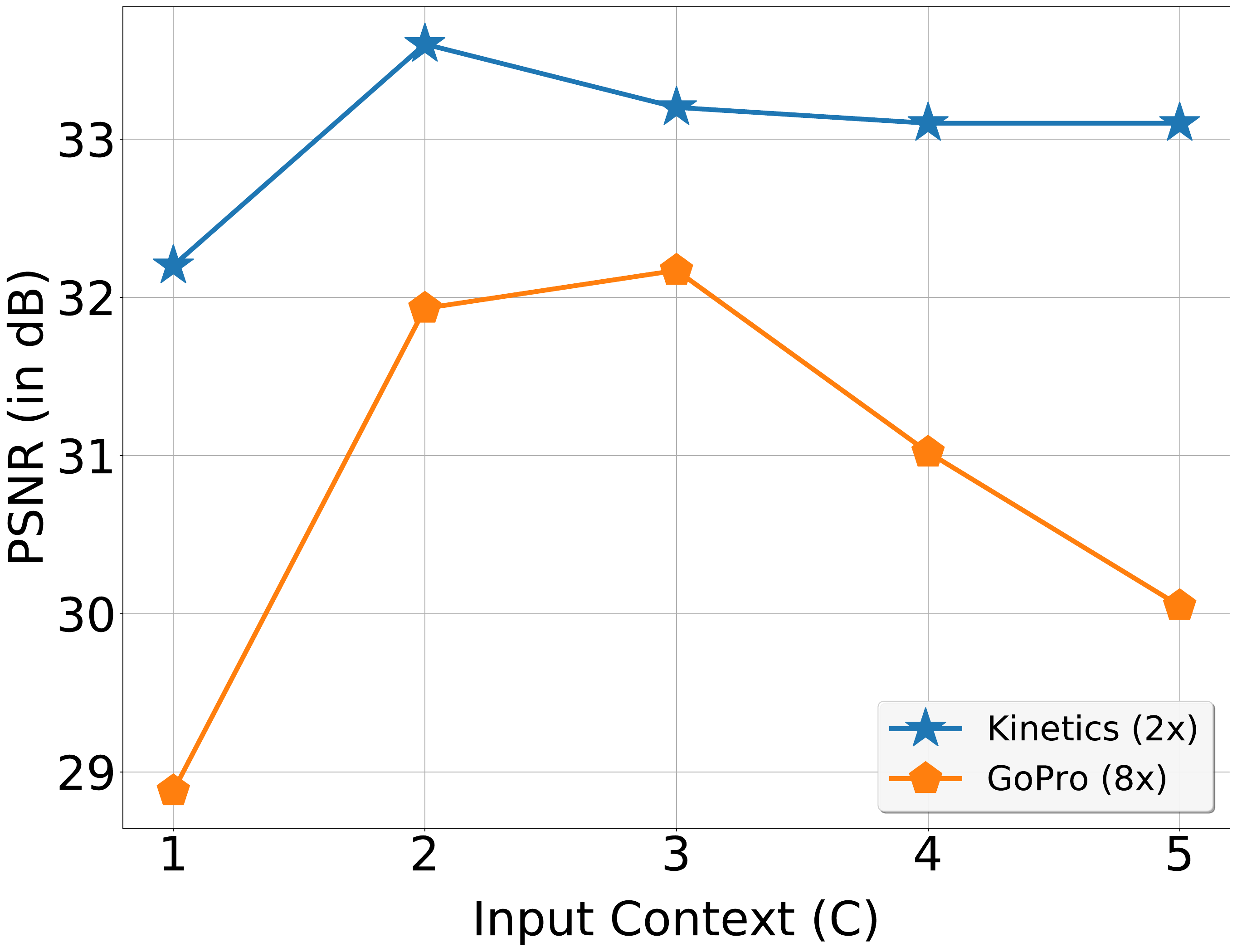}
    \end{center}
    \caption{{\bf Effect of Input Context} Comparison of the effect of input context, $C$, for video frame interpolation. For \twox{} interpolation, we observed that a value of $C=2$ which corresponds to using 4 input frames, 2 each from the past and future, gives best results. Beyond $C=3$, we observe no further improvements. For \eightx{} interpolation, a value of $C=3$ gave the best accuracy. }
    \label{fig:effectOfC}
\end{figure}

%
\begin{table*}[h]
\centering
\begin{subtable}[t]{.21\textwidth}
\resizebox{\textwidth}{!}{
\begin{tabular}{lcc}
\hline
Model & PSNR & SSIM \\
\midrule
R2D-18-2I & 33.98 &	0.966 \\
R2D-18-4I &	34.97 &	0.967 \\
R3D-18-4I &	\textbf{36.3} &	\textbf{0.975} \\
\bottomrule
\end{tabular}}
\captionsetup{width=.9\textwidth}
\subcaption{Effect of encoder arch.}
\label{tab:backbone}
\end{subtable}
\hfill
\begin{subtable}[t]{.23\textwidth}
\resizebox{\textwidth}{!}{
\begin{tabular}{l|cc}
\hline
Model & PSNR & SSIM \\
\midrule
No fusion &	35.1 &	0.9713 \\
fusion - add &	35.7 &	0.9737\\
fusion - concat &	\textbf{36.3} &	\textbf{0.975}\\
\bottomrule
\end{tabular}}
\captionsetup{width=.9\textwidth}
\subcaption{Type of feature fusion}
\label{tab:concat}
\end{subtable}
\hfill
\begin{subtable}[t]{.23\textwidth}
\resizebox{\textwidth}{!}{
\begin{tabular}{l|cc}
\hline
Model & PSNR & SSIM \\
\midrule
w/o stride   &	\textbf{36.3}  &	\textbf{0.975} \\
w/ 2x stride &	35.4  &	0.961 \\
w/ 4x stride &	35.21 &	0.96 \\
\bottomrule
\end{tabular}}
\captionsetup{width=.9\textwidth}
\subcaption{Effect of temporal striding}
\label{tab:stride}
\end{subtable}
\hfill
\begin{subtable}[t]{.23\textwidth}
\resizebox{\textwidth}{!}{
\begin{tabular}{l|cc}
\hline
Model & PSNR & SSIM \\
\midrule
L1 Loss      &	\textbf{36.3}    &	\textbf{0.975} \\
L2 Loss       &	35.3    &	0.965 \\
Huber Loss    &	35.3    &	0.964 \\
L1+VGG Loss      &	35.91   &	0.962 \\
\bottomrule
\end{tabular}}
\captionsetup{width=.9\textwidth}
\subcaption{Effect of loss function}
\label{tab:lossFn}
\end{subtable}
\captionsetup{width=\textwidth}
\vspace{-8pt}
\caption{{\bf Ablation results} for \Ours{} architecture on \subref{tab:backbone} different backbones, \subref{tab:concat} fusion methods, \subref{tab:stride} temporal striding, and \subref{tab:lossFn} loss functions.}
\label{tab:ablations}
\vspace{-8pt}
\end{table*}
In \figref{fig:effectOfC}, we present a more detailed ablation about the effect of input context ($C$) on the performance of interpolation. From \figref{fig:effectOfC}, we observe that for both \twox{} and \eightx{} interpolations, using two input frames ($C=1$), one each from past and present is sub-optimal, as it fails to accurately reason about complex motion profiles and occlusions. Furthermore, for \twox{} interpolation, we found that a value of $C=2$ gave the best result, and beyond that the performance saturates. This is because the outer frame generally contain less useful information for interpolation and in some cases might contain significant scene shifts which hurts the interpolation accuracy. In the case of \eightx{} interpolation, the time gap between the frames is tinier, so we find that a value of $C=3$ performs the best, while any larger value of $C$ hurts the accuracy.

{\bf Choice of Fusion} \tabref{tab:concat} compares and reports the different choices for the skip connection (in 
Figure 2 \textcolor{black}{of the main submission}) used for combining features across encoder and corresponding decoder. \textit{No fusion } corresponds to having no skip connection between the layers of the encoder and decoder. While \textit{fusion - add } corresponds to adding the features from the encoder to the decoder, \textit{fusion - concat} refers to concatenating the corresponding feature maps along the channels. We find that using some kind of feature transfer across encoder and decoder is essential, than having \textit{No fusion } (PSNR of $36.11$ vs. $35.1$), as the complementary information learnt in the low level and high level features needs to be aggregated for accurate interpolation. We settle on using \textit{fusion - concat} in our final model as it gives better performance than \textit{fusion - add}. 

{\bf Temporal Striding} Striding or pooling in CNNs are known to remove lot of fine level details in images, which are essential for generative tasks like frame interpolation. We verify this with experiments using \twox{}(1/\twox{}) and \fourx{}(1/\fourx{}) temporal striding in the encoder(decoder), and observe from \tabref{tab:stride} that the performance decreases from $36.3$ to $35.2$ with larger temporal striding. We conclude that temporal striding hurts, and use a temporal stride of 1 in all the 3D convolution layers.

\begin{figure*}
     \centering
     
     \begin{subfigure}[t]{0.28\textwidth}
         \centering
         \includegraphics[width=\textwidth]{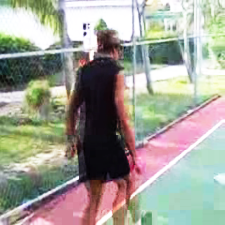}
         \subcaption{Overlayed inputs}
         \label{fig:overlay}
     \end{subfigure}
     \begin{subfigure}[t]{0.28\textwidth}
         \centering
         \includegraphics[width=\textwidth]{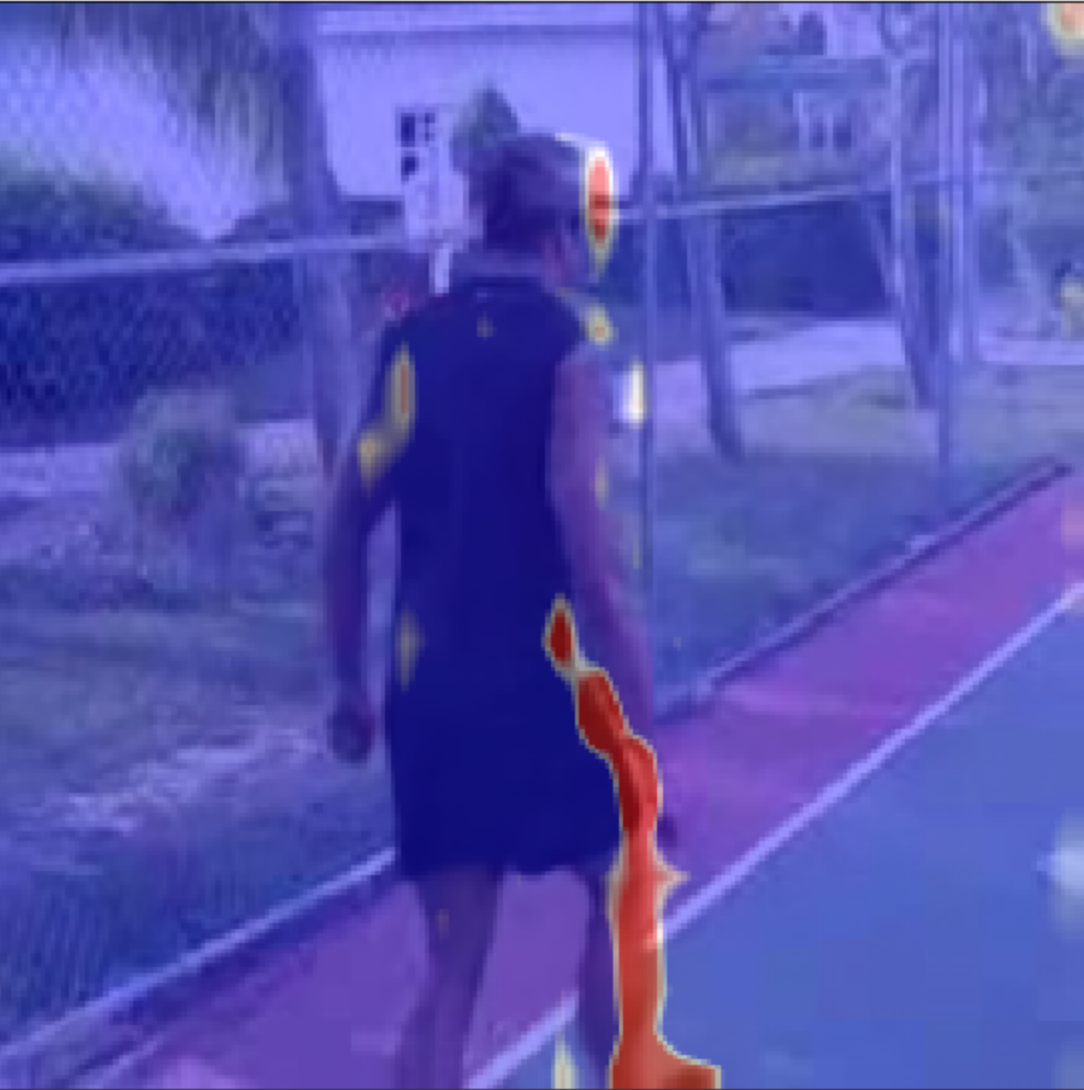}
         \subcaption{Activation w/ gating}
         \label{fig:withAttn}
     \end{subfigure}
     \begin{subfigure}[t]{0.28\textwidth}
         \centering
         \includegraphics[width=\textwidth]{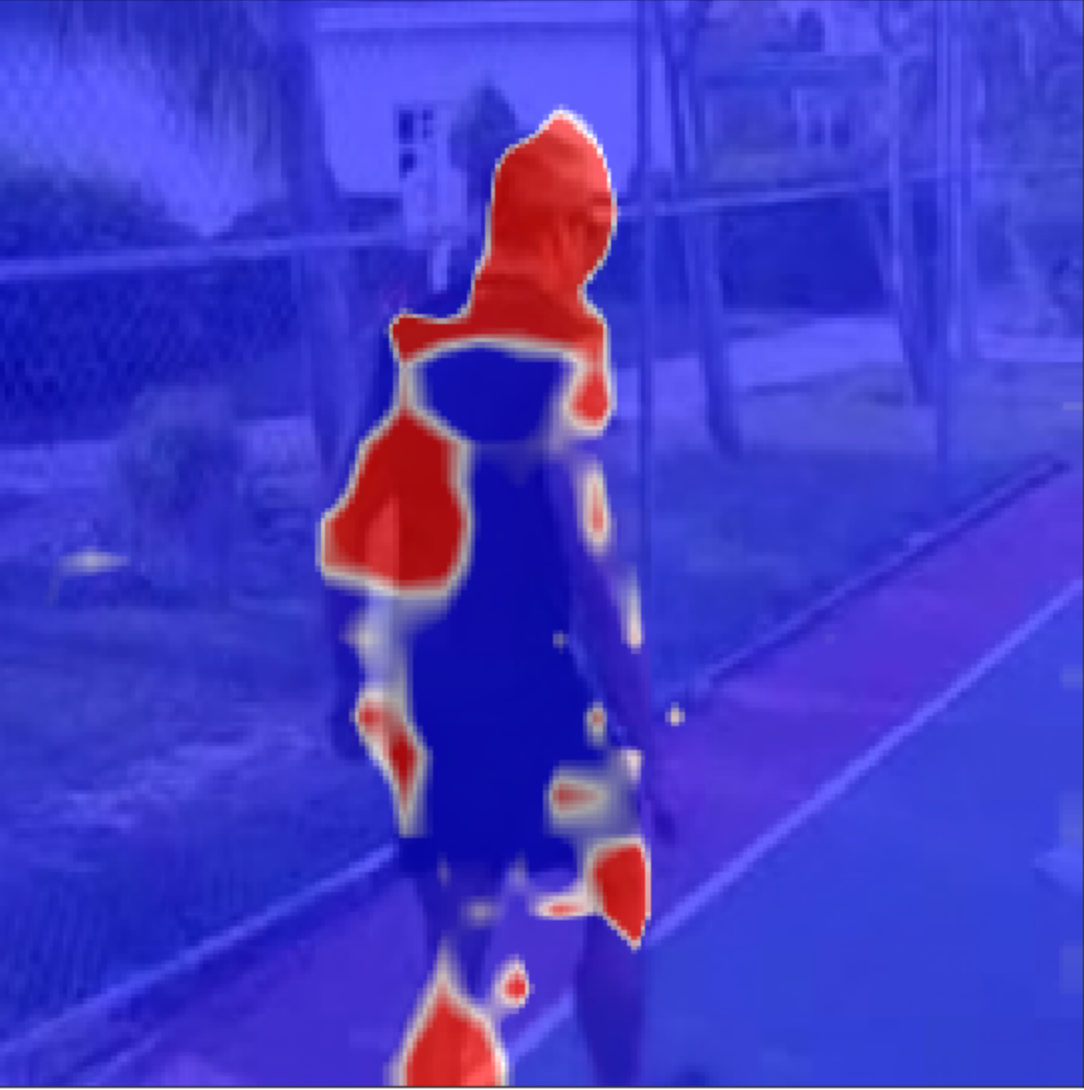}
         \captionsetup{width=\textwidth}
         \subcaption{Activation w/o gating}
         \label{fig:withoutAttn}
     \end{subfigure}
    \caption{{\bf Visualization of attention weighted feature maps} (a) The overlayed input frames. (b) The feature map of the channel with the highest attention weight in the network with feature gating. (c)  The same feature map without using the gating module. We observe higher activation (red) in (b) along the motion boundaries. Best viewed in color.
    }
    \label{fig:attnMaps}
\end{figure*}

{\bf Channel Gating} We visualize the role of channel gating module in the network in \figref{fig:attnMaps}. We show the overlapped input frames in \figref{fig:overlay} to highlight the parts which have motion. In \figref{fig:withAttn} and \figref{fig:withoutAttn}, we plot the feature maps corresponding to the channel dimension with the largest activation while using and without using the feature gating respectively. We observe that the network trained with spatio-temporal gating (\figref{fig:withAttn}) learns to focus on parts of input with visible motions (high activations in red), thus resulting in confident predictions of the interpolated motion estimates compared to \figref{fig:withoutAttn}. In fact, training without spatio-temporal gating results in a drop in PSNR value from $36.3$ to $36.1$, further validating the utility of having the gating module.

{\bf Loss Function} Many previous works~\cite{niklaus2017video} have studied the effect of using purely pixel loss vs. perception based losses \cite{johnson2016perceptual}. Using only L1 or L2 loss would improve on the PSNR metric, but would cause blur in predictions. On the other hand, adding VGG based perception loss would result in sharper images visually. We observe from \tabref{tab:lossFn} that we did not improve upon the PSNR or SSIM metric by using any additional loss functions like VGG loss or Huber loss, apart from just L1 loss which also resulted in visually sharper images in our case.  

\section{Experiment Settings for downstream applications}

\subsection{Low-fps video object segmentation details}

To examine the effectiveness of using the outputs of \Ours{}, we choose the task of object segmentation in videos using mask propagation. 

{\bf Motivation} Achieving good label (or mask) propagation requires estimating perfect pixel level correspondences between frames of a video, using similarity between the respective feature maps. However, estimating such correspondences might be challenging if the frame sequences are extracted from low-fps videos. We want to validate if using \Ours{} can improve low-fps video object segmentation.

{\bf Setup and baseline}. DAVIS is the standard benchmark popularly used for video object segmentation which include videos at 30FPS. To adopt to low-fps setup, we purposely downsample DAVIS videos into lower frame rates, e.g., 15FPS or 8 FPS, and evaluate different object segmentation approaches on these low-fps videos. We choose CRW \cite{jabri2020space}, the current state-of-the-art method for video object segmentation, as a baseline which is applied directly on downsampled low-fps videos. We then compare this baseline with using the same method, i.e. CRW, on interpolated videos generated by \Ours{} by 2x or 4x interpolation from 15FPS or 8FPS videos. Results are shown in the main submission showing that \Ours{} helps to improve low-fps video object tracking. The label propagation mechanism is the same as used in \cite{jabri2020space}.

\subsection{Motion magnification}
\label{appendix:motion_magnification}

Motion Magnification~\cite{oh2018learning, wu2012eulerian, Wadhwa13PhaseBased} deals with magnifying subtle yet important motions from videos, which are often imperceptible by human eyes. From~\cite{oh2018learning}, we define motion magnification as follows. For an Image $I(\mathbf{x},t) = f(\mathbf{x} + \delta(\mathbf{x} , t))$, the goal of motion magnification is to generate an output image $\Tilde{I}(\mathbf{x} , t)$ such that

\begin{equation}
    \Tilde{I}(\mathbf{x} , t) = f(\mathbf{x} +(1+\alpha) \delta(\mathbf{x} , t))
\end{equation}

\noindent for a magnification factor $\alpha$. For frame interpolation, $\alpha < 1$, since we are interested in what happens between two frames while for motion magnification, $\alpha > 1$, since we look to extrapolate existing motions beyond visible regime. While prior works~\cite{oh2018learning, wu2012eulerian, Wadhwa13PhaseBased} used custom architectures along with various post processing filters for this task, we offer a complementary perspective and look into how much a simple architecture like FLAVR pretrained on frame interpolation helps motion magnification. For this purpose, we use the synthetic dataset \textit{CoCo-Synth}~\cite{oh2018learning} to perform the training. We train the network for a fixed magnification factor of 10 ($\alpha=10$). On this dataset, when compared to no pretraining at all, pretraining on \Ours{} improved the SSIM values on a held-out validation set from $0.732$ to $0.801$. We provide sample videos after magnification and compare it with phase based approach~\cite{wu2012eulerian} in our supplementary video. We emphasize that we do not apply any post processing such as temporal or spatial filters for removing noise on the outputs. The videos are generated directly as an output of the FLAVR architecture pretrained on frame interpolation, and finetuned for motion magnification.

\subsection{Experiment setting for action recognition}
\label{appendix:action_recognition}

For downstream experiments on action recognition, we use the train and validation split 1 of UCF101~\cite{UCF101} and HMDB51~\cite{kuehne2011hmdb}. We remove the decoder from the architecture and use the pretrained encoder along with a classifier (a global average pooling, a fully-connected layer, and a softmax) for training on downstream actions and add a temporal stride of 4. For UCF101, we use an input size of $32 \times 3 \times 224 \times 224$ and for HMDB51 we use an input size of $16 \times 3 \times 224 \times 224$ with a batch size of 16. The networks are fine-tuned using SGD with batch norm with a learning rate of $0.02$ for 40 epochs. During inference, we sample 10 consecutive overlapping clips of length 32 from the test video and average predictions over all the clips.

\subsection{Experiment setting for optical flow estimation}
\label{appendix:flow_estimation}

One crucial point to consider in downstream training on optical flow is that the flow networks generally take only two input frames which is considered too short for 3D CNN. Nevertheless, to examine the effectiveness of features learnt using frame interpolation for optical flow, we use the same encoder and decoder, and initialize the last prediction layer to output two channels instead (corresponding to x and y values of flow at each pixel). Since the interpolation network was trained to take 4-frame inputs, we apply copy padding to the inputs, e.g. repeating each input frame 2 times. We use an EPE (end point error) loss and train our network for 200 epochs. We report numbers using 5-fold cross validation over the MPI-Sintel clean and fina as well as Kitti subsets.

\section{Qualitative Results}
\label{appendix:qual}


\begin{figure*}[t]
    \begin{center}
    \begin{subfigure}[b]{0.1\textwidth}
        \centering
        \includegraphics[width=\textwidth]{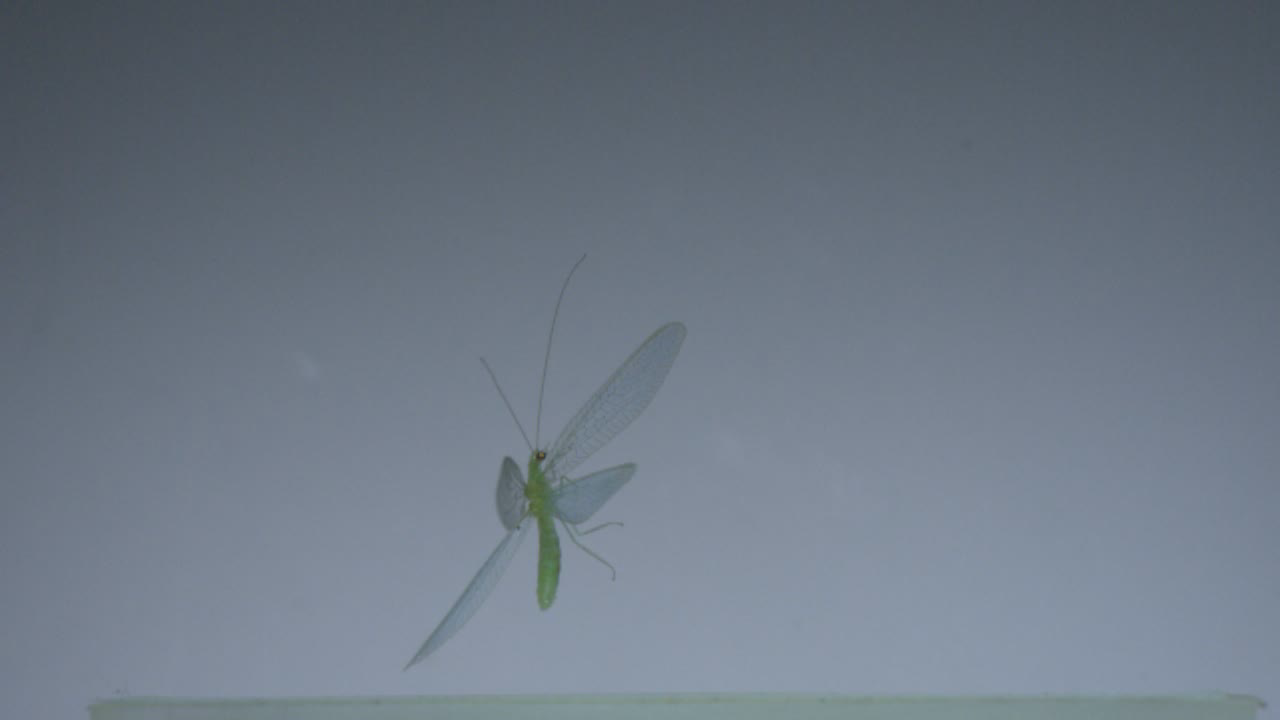}
    \end{subfigure}
    \hfill
    \begin{subfigure}[b]{0.1\textwidth}
        \centering
        \includegraphics[width=\textwidth]{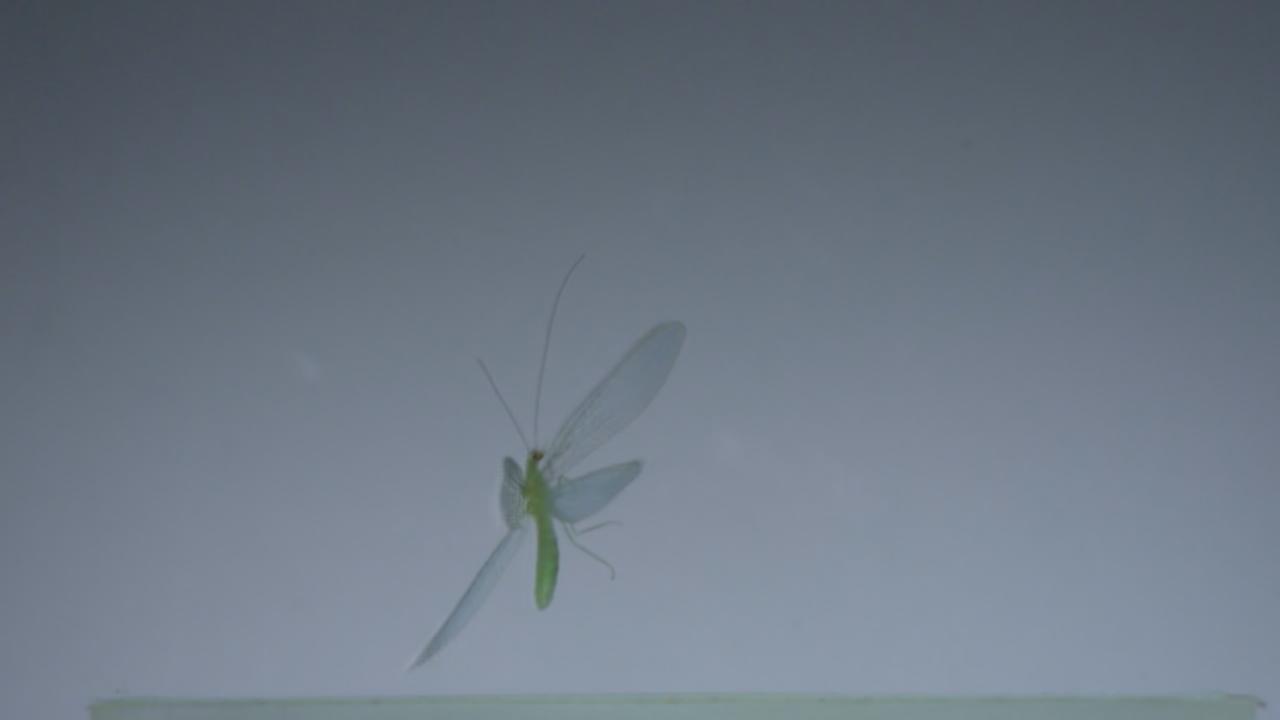}
    \end{subfigure}
    \hfill
    \begin{subfigure}[b]{0.1\textwidth}
        \centering
        \includegraphics[width=\textwidth]{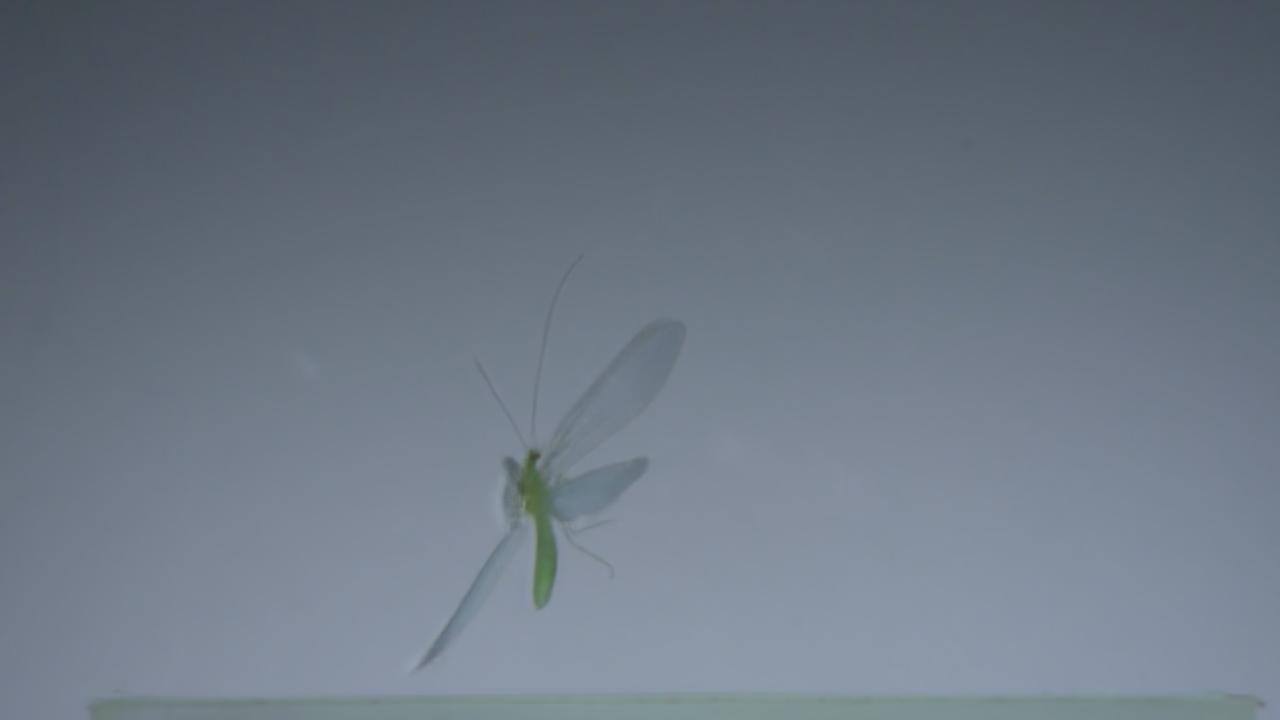}
    \end{subfigure}
    \hfill
    \begin{subfigure}[b]{0.1\textwidth}
        \centering
        \includegraphics[width=\textwidth]{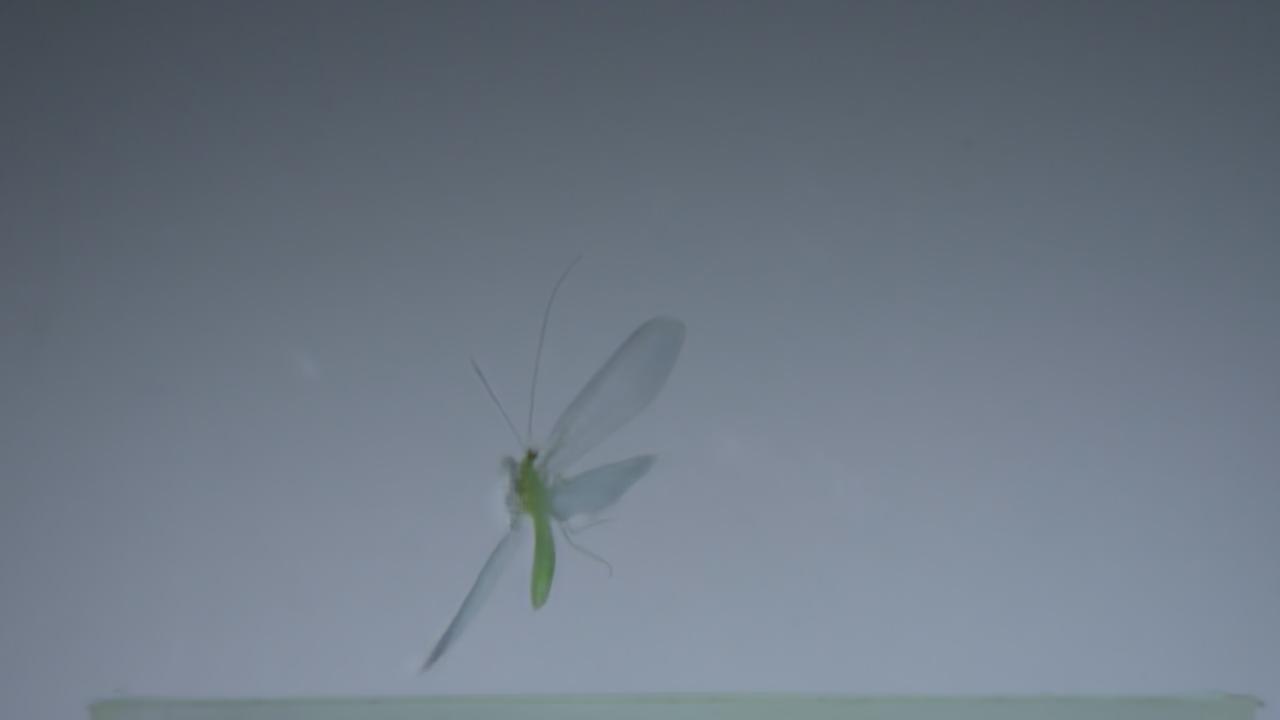}
    \end{subfigure}
    \hfill
    \begin{subfigure}[b]{0.1\textwidth}
        \centering
        \includegraphics[width=\textwidth]{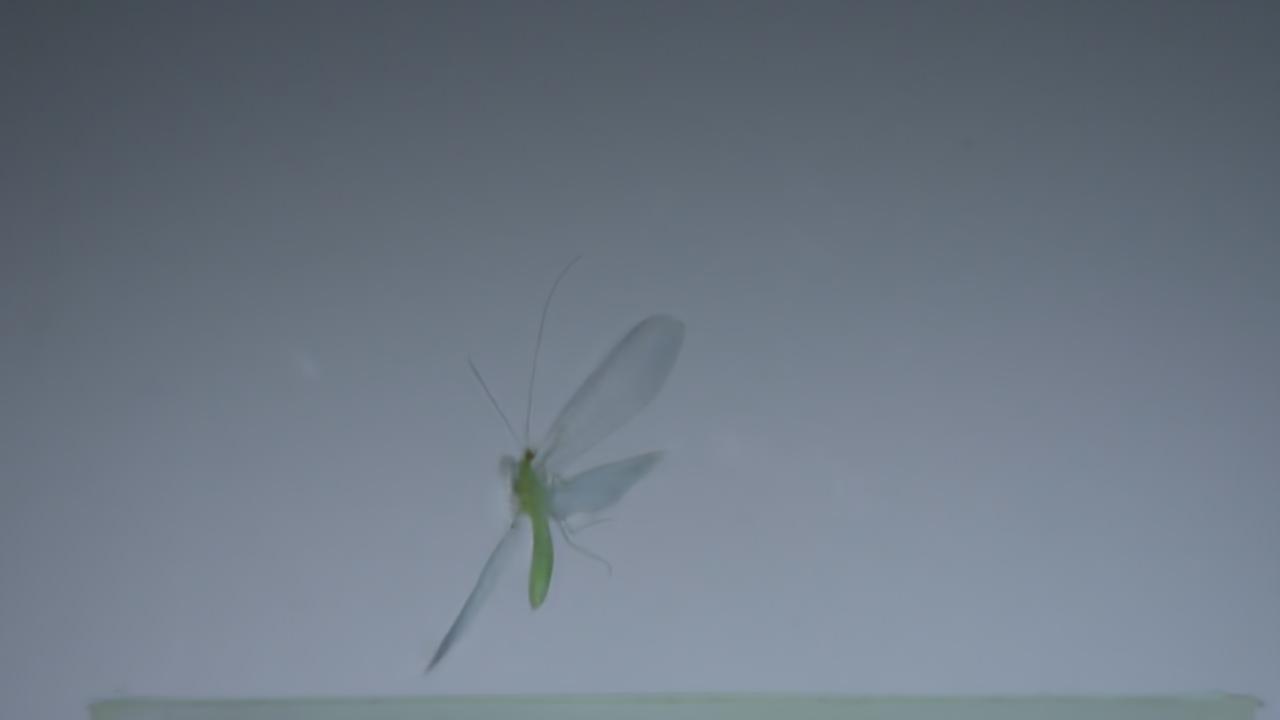}
    \end{subfigure}
    \hfill
    \begin{subfigure}[b]{0.1\textwidth}
        \centering
        \includegraphics[width=\textwidth]{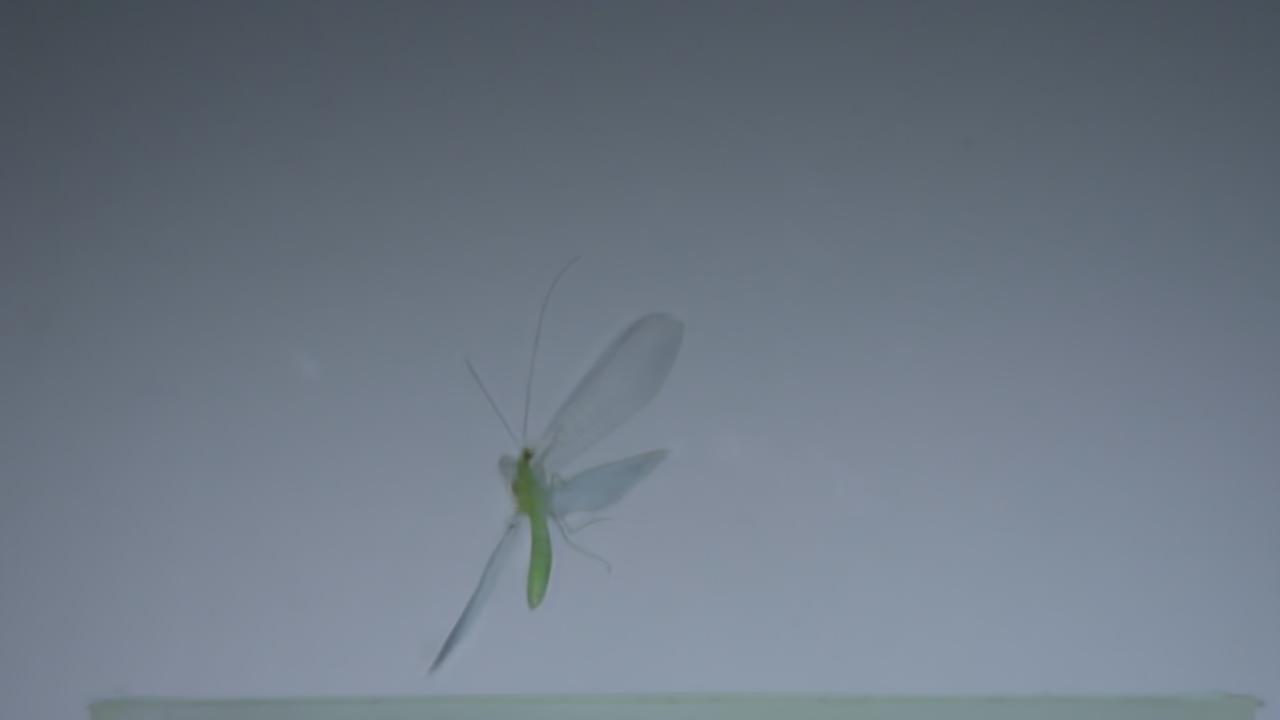}
    \end{subfigure}
    \hfill
    \begin{subfigure}[b]{0.1\textwidth}
        \centering
        \includegraphics[width=\textwidth]{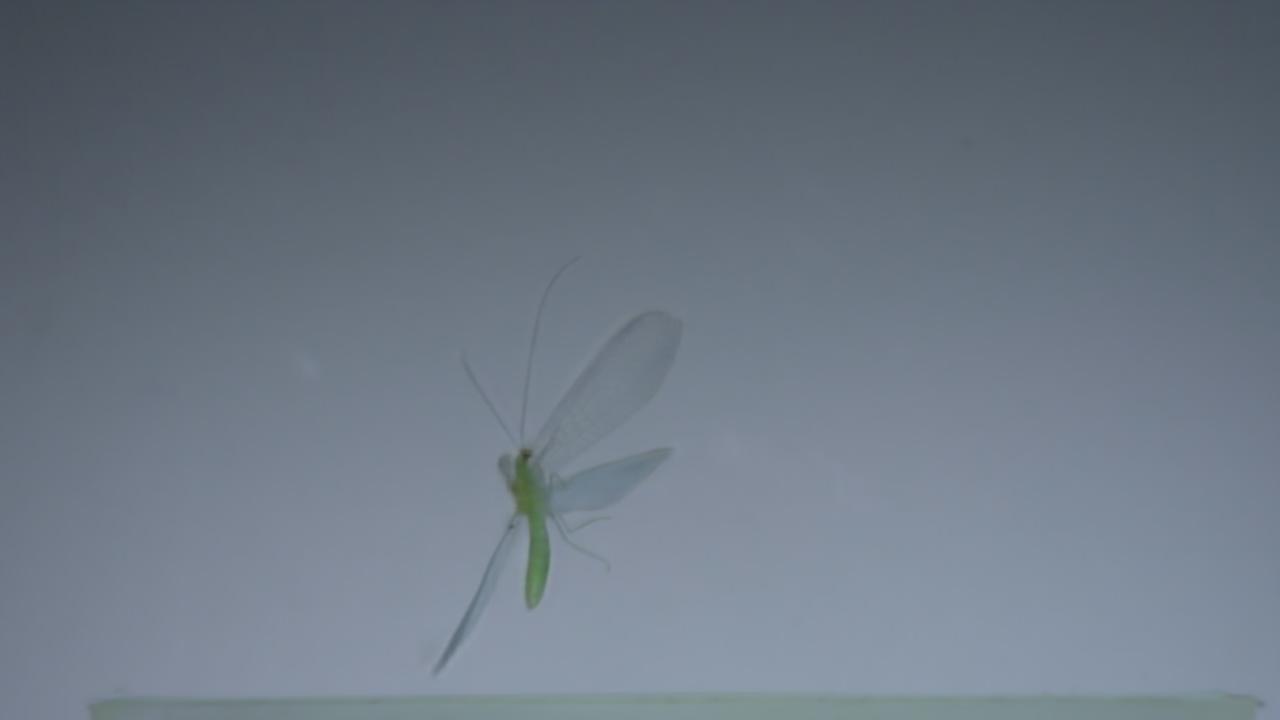}
    \end{subfigure}
    \hfill
    \begin{subfigure}[b]{0.1\textwidth}
        \centering
        \includegraphics[width=\textwidth]{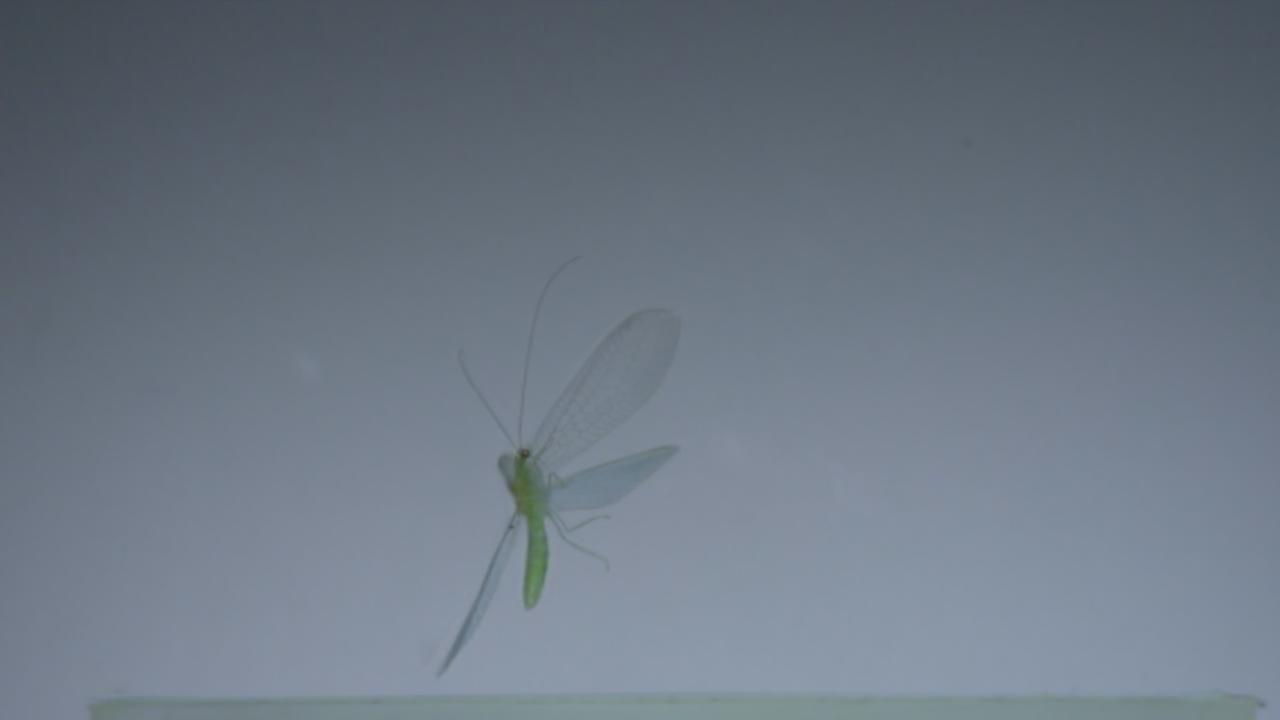}
    \end{subfigure}
    \hfill
    \begin{subfigure}[b]{0.1\textwidth}
        \centering
        \includegraphics[width=\textwidth]{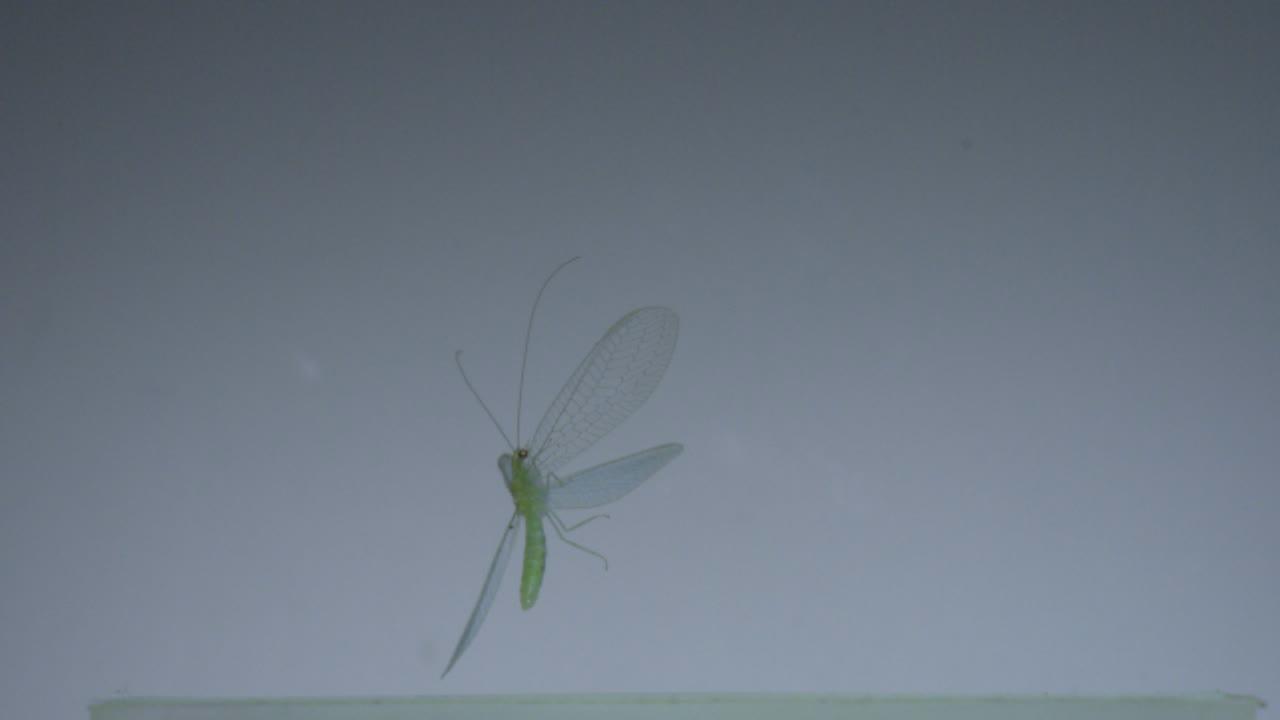}
    \end{subfigure}
    
    
    \begin{subfigure}[b]{0.1\textwidth}
        \centering
        \includegraphics[width=\textwidth]{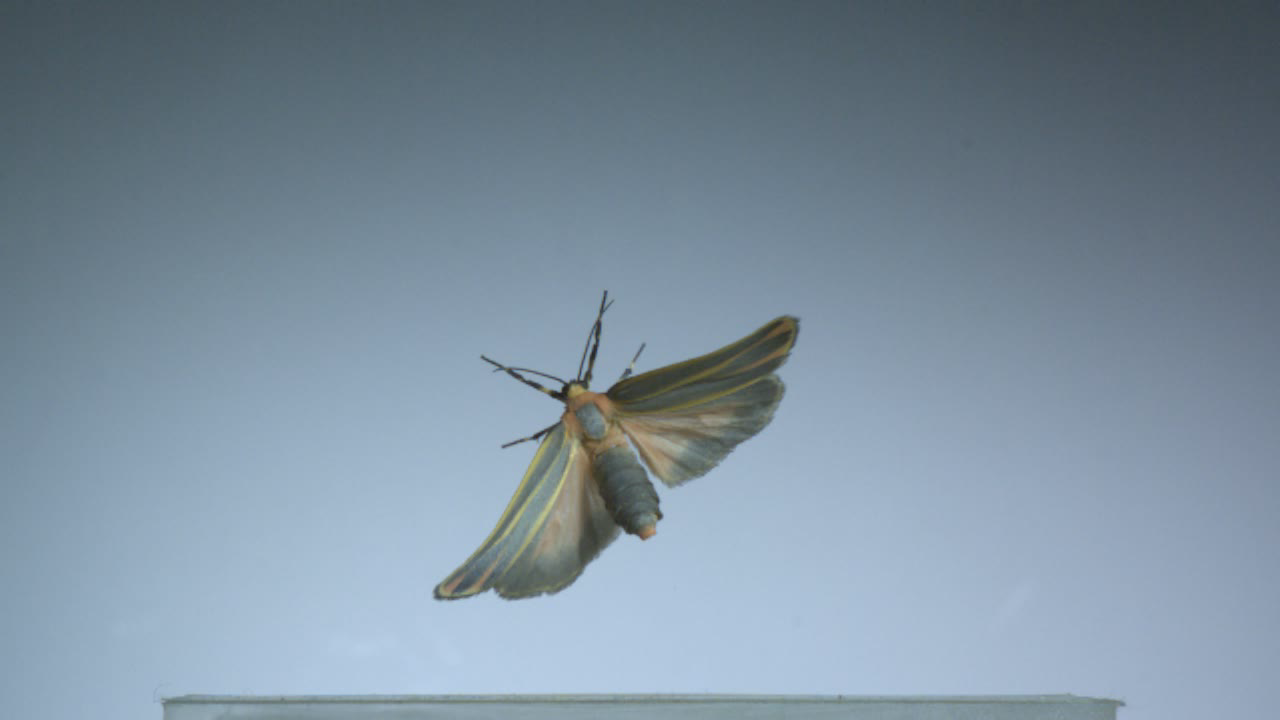}
    \end{subfigure}
    \hfill
    \begin{subfigure}[b]{0.1\textwidth}
        \centering
        \includegraphics[width=\textwidth]{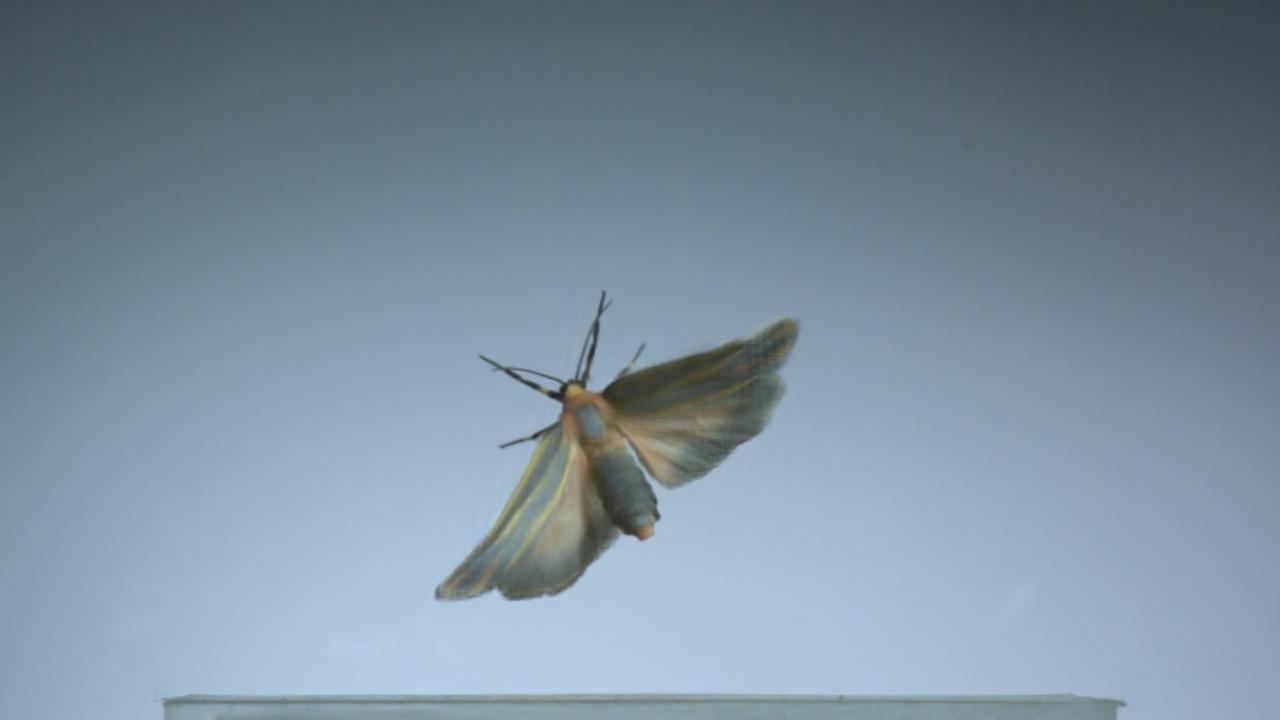}
    \end{subfigure}
    \hfill
    \begin{subfigure}[b]{0.1\textwidth}
        \centering
        \includegraphics[width=\textwidth]{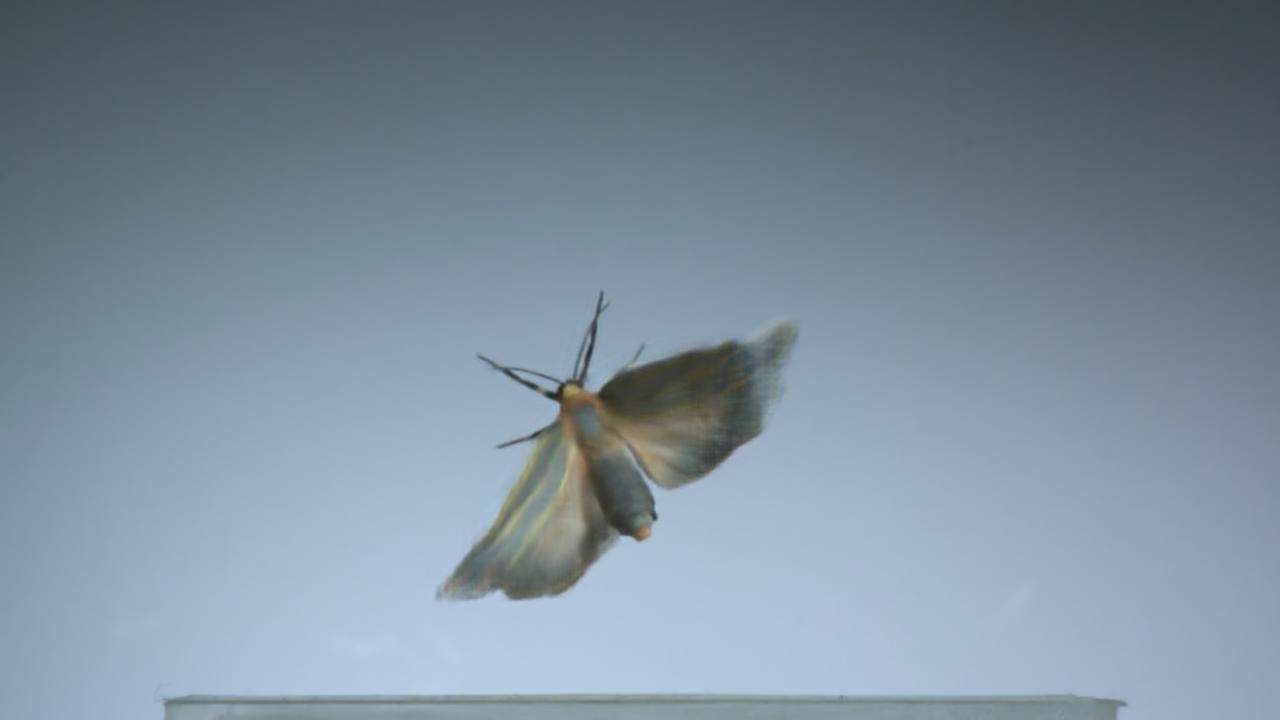}
    \end{subfigure}
    \hfill
    \begin{subfigure}[b]{0.1\textwidth}
        \centering
        \includegraphics[width=\textwidth]{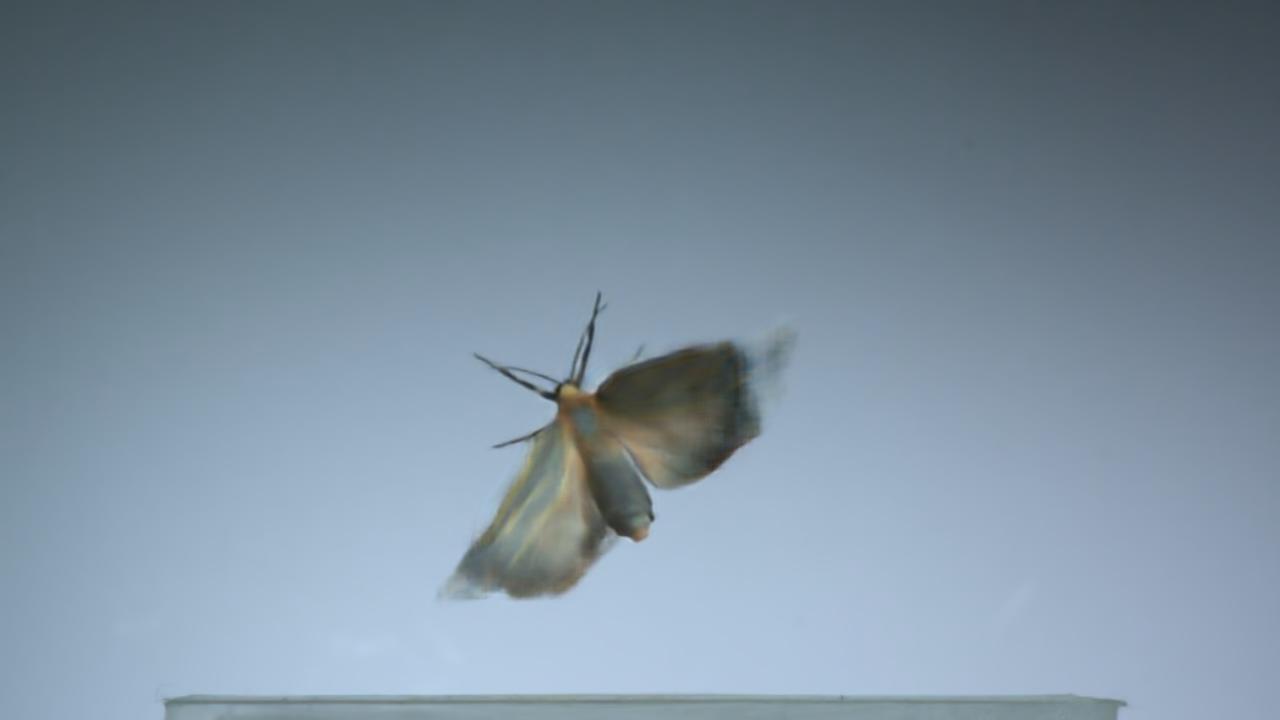}
    \end{subfigure}
    \hfill
    \begin{subfigure}[b]{0.1\textwidth}
        \centering
        \includegraphics[width=\textwidth]{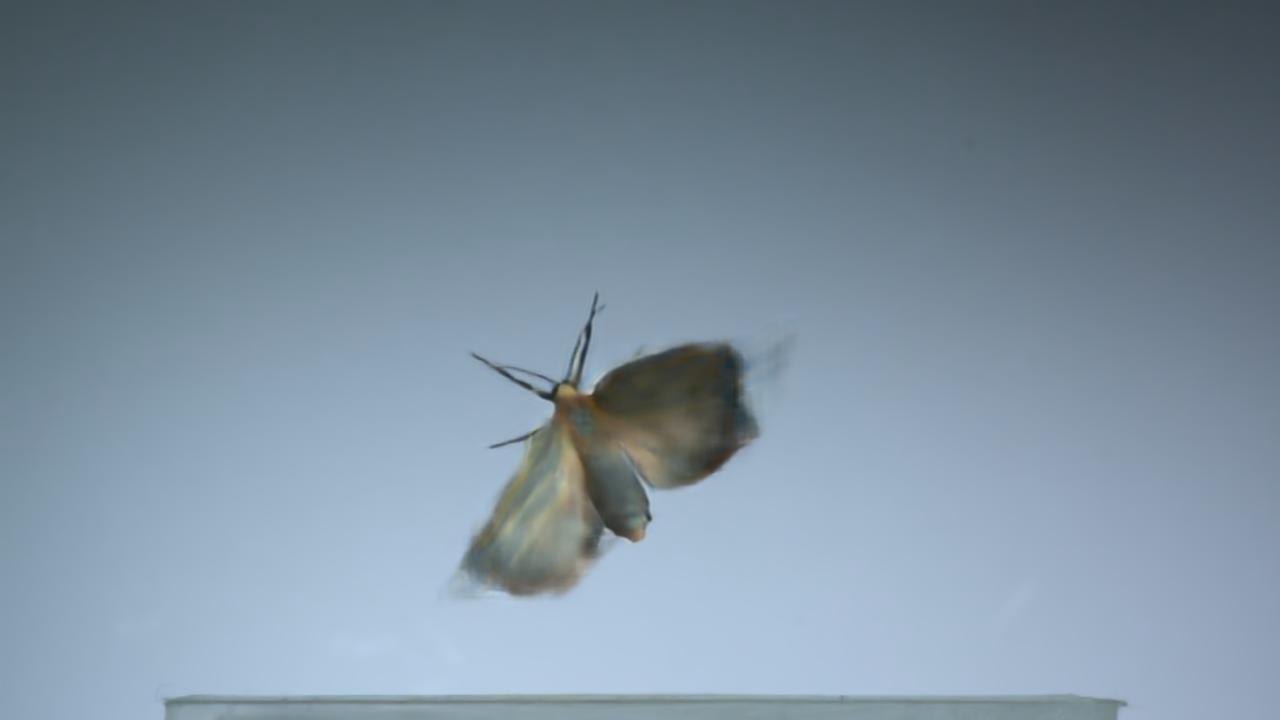}
    \end{subfigure}
    \hfill
    \begin{subfigure}[b]{0.1\textwidth}
        \centering
        \includegraphics[width=\textwidth]{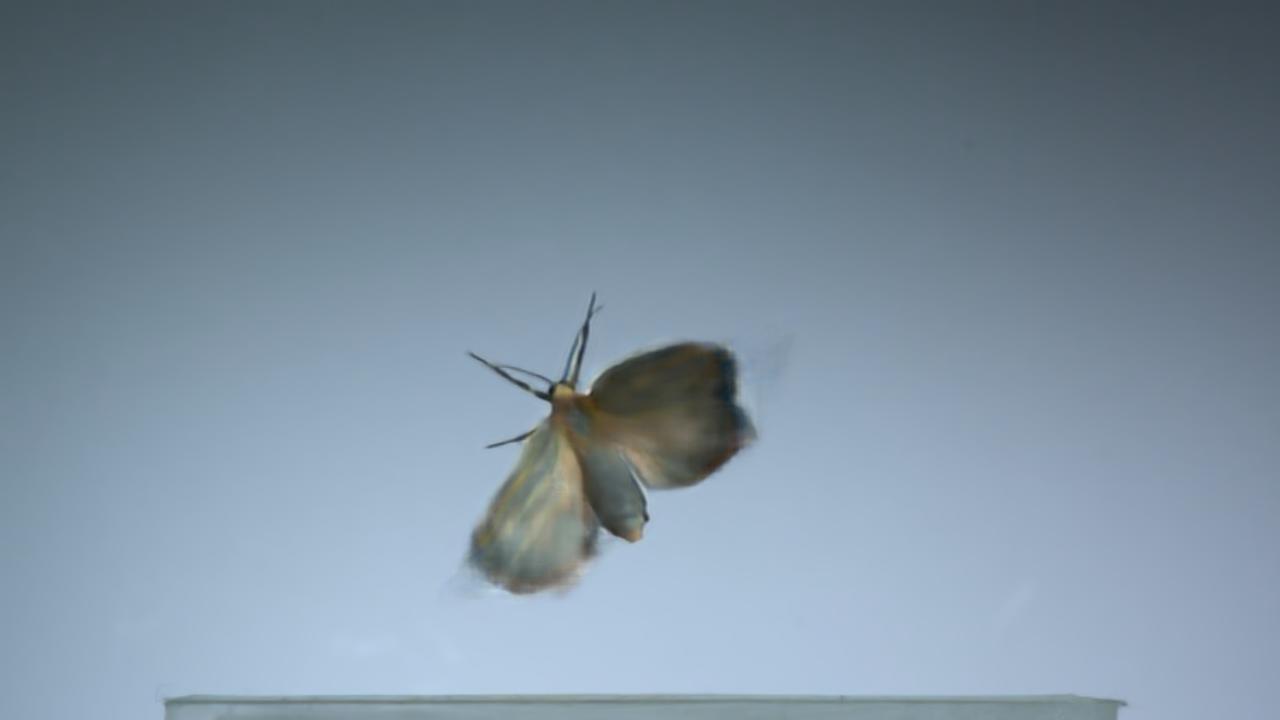}
    \end{subfigure}
    \hfill
    \begin{subfigure}[b]{0.1\textwidth}
        \centering
        \includegraphics[width=\textwidth]{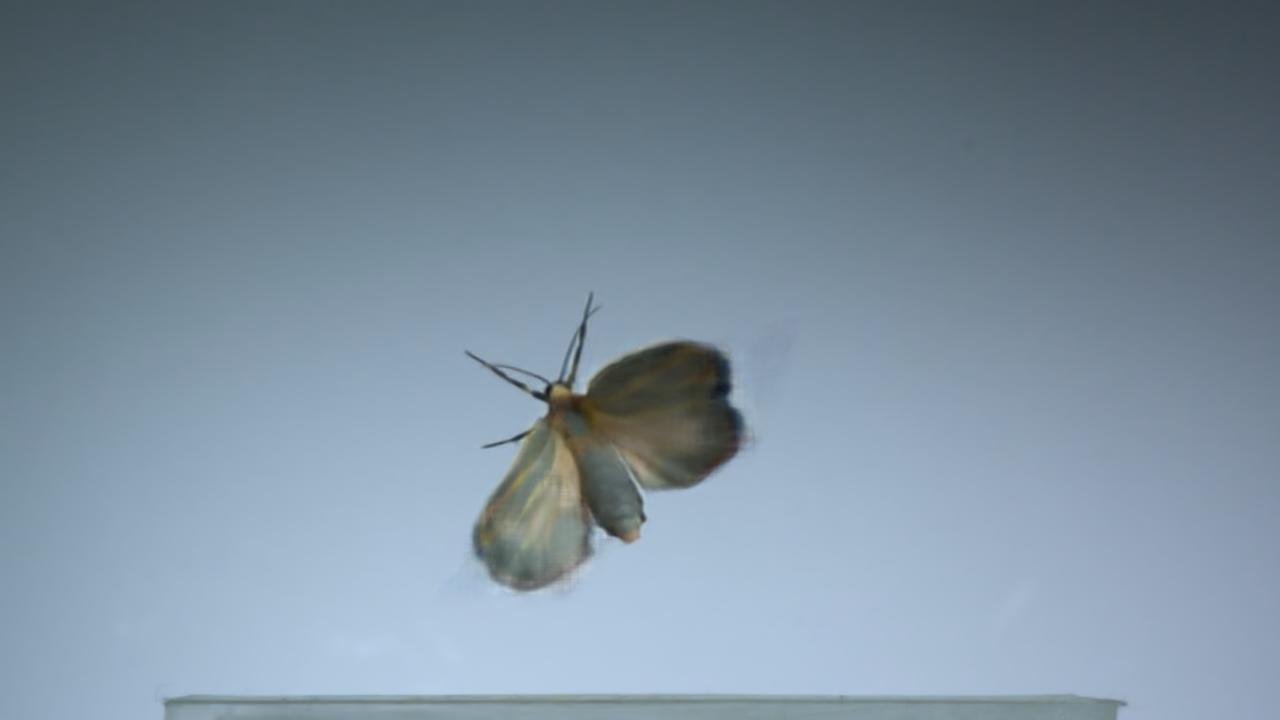}
    \end{subfigure}
    \hfill
    \begin{subfigure}[b]{0.1\textwidth}
        \centering
        \includegraphics[width=\textwidth]{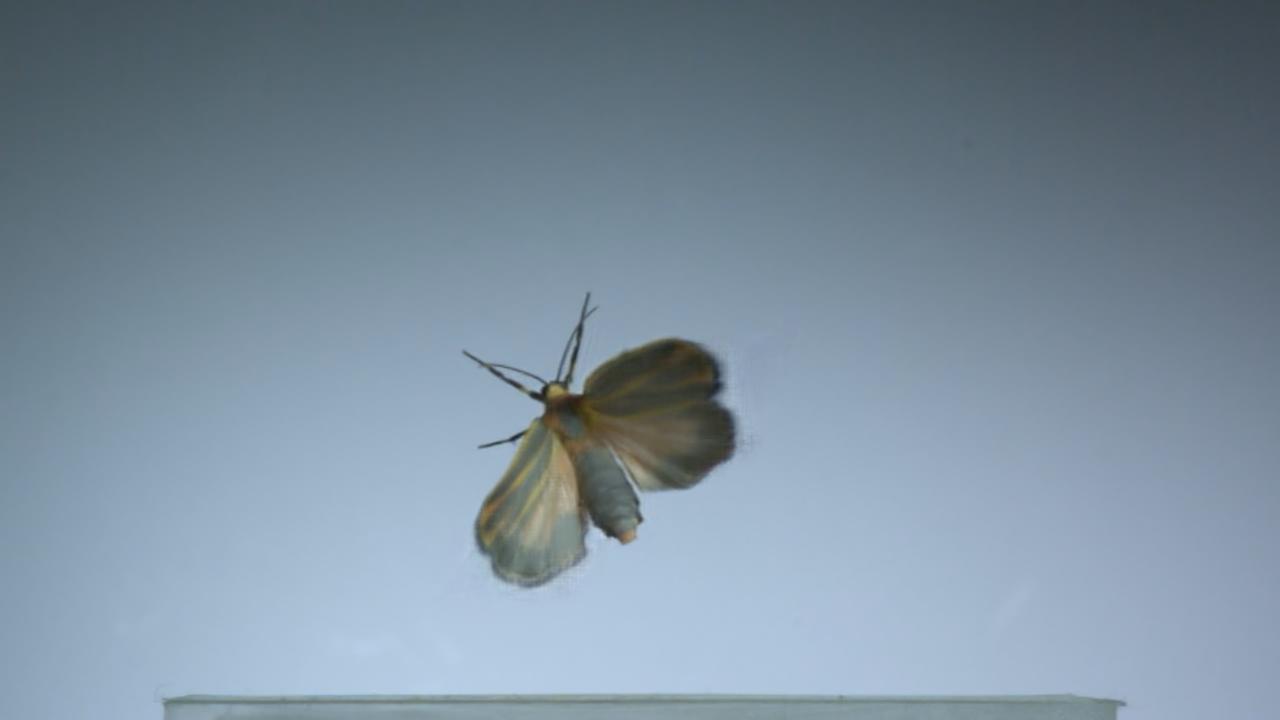}
    \end{subfigure}
    \hfill
    \begin{subfigure}[b]{0.1\textwidth}
        \centering
        \includegraphics[width=\textwidth]{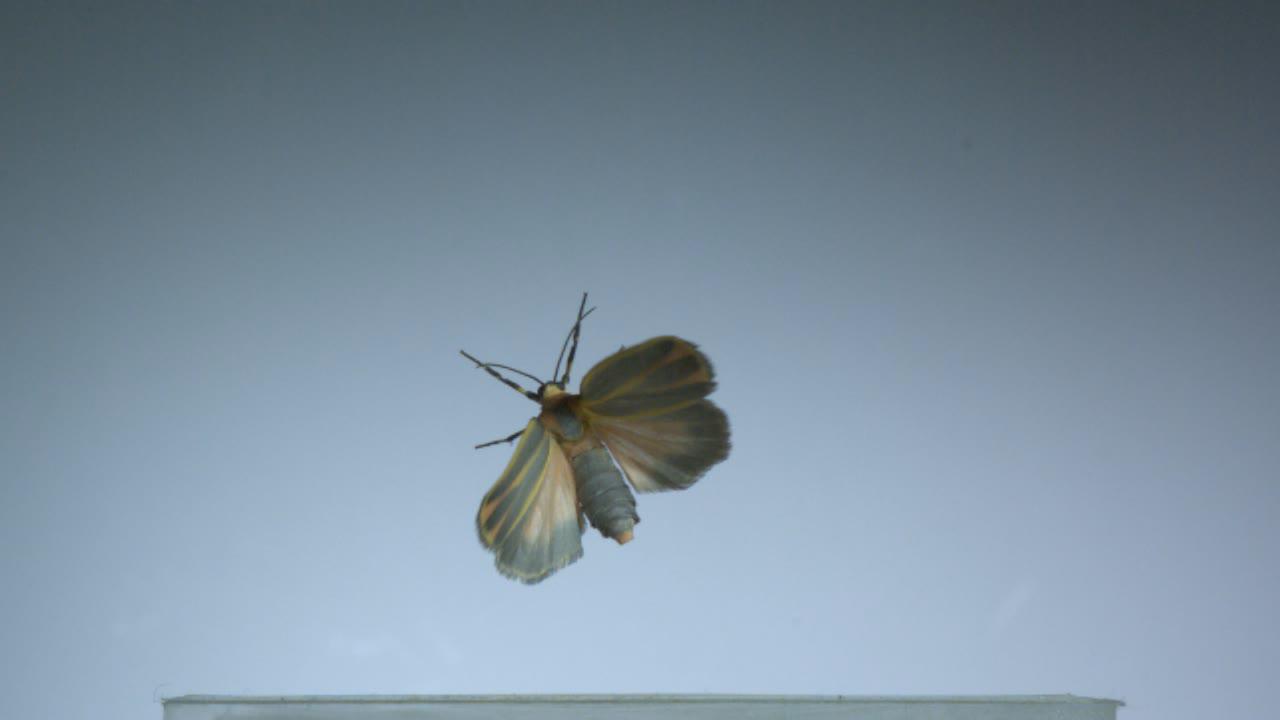}
    \end{subfigure}
    
    
    \begin{subfigure}[b]{0.1\textwidth}
        \centering
        \includegraphics[width=\textwidth]{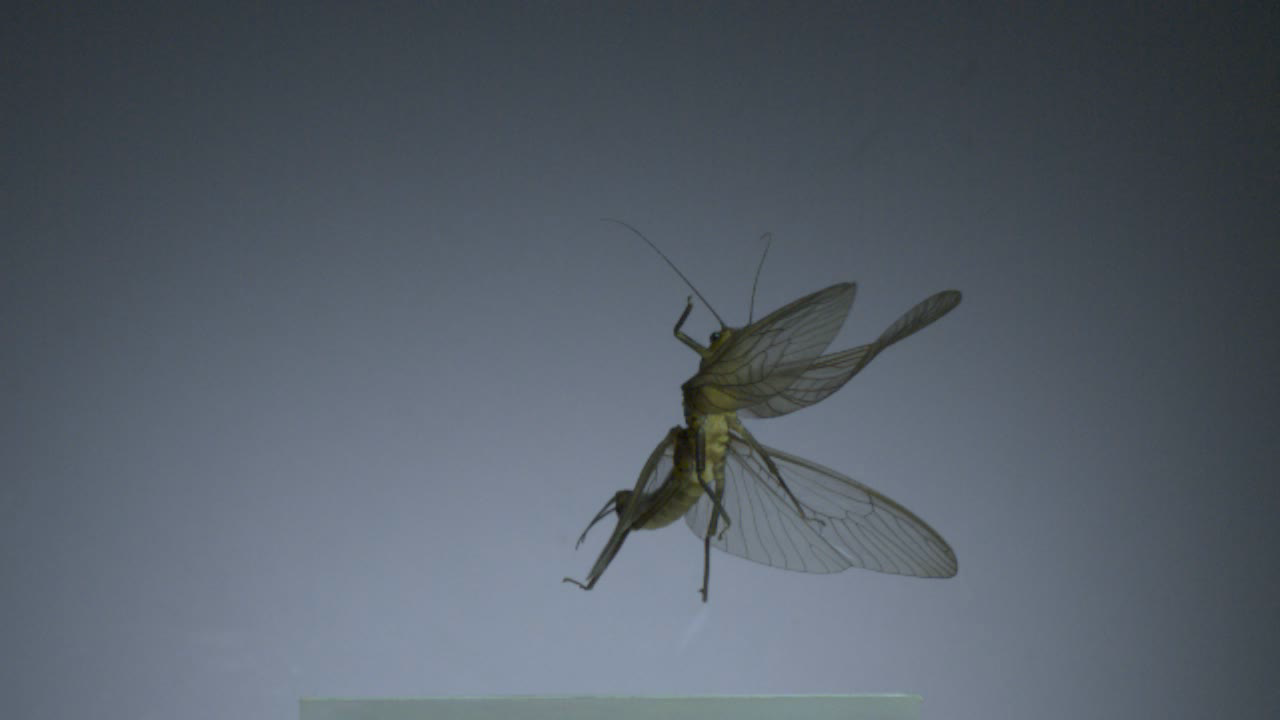}
    \end{subfigure}
    \hfill
    \begin{subfigure}[b]{0.1\textwidth}
        \centering
        \includegraphics[width=\textwidth]{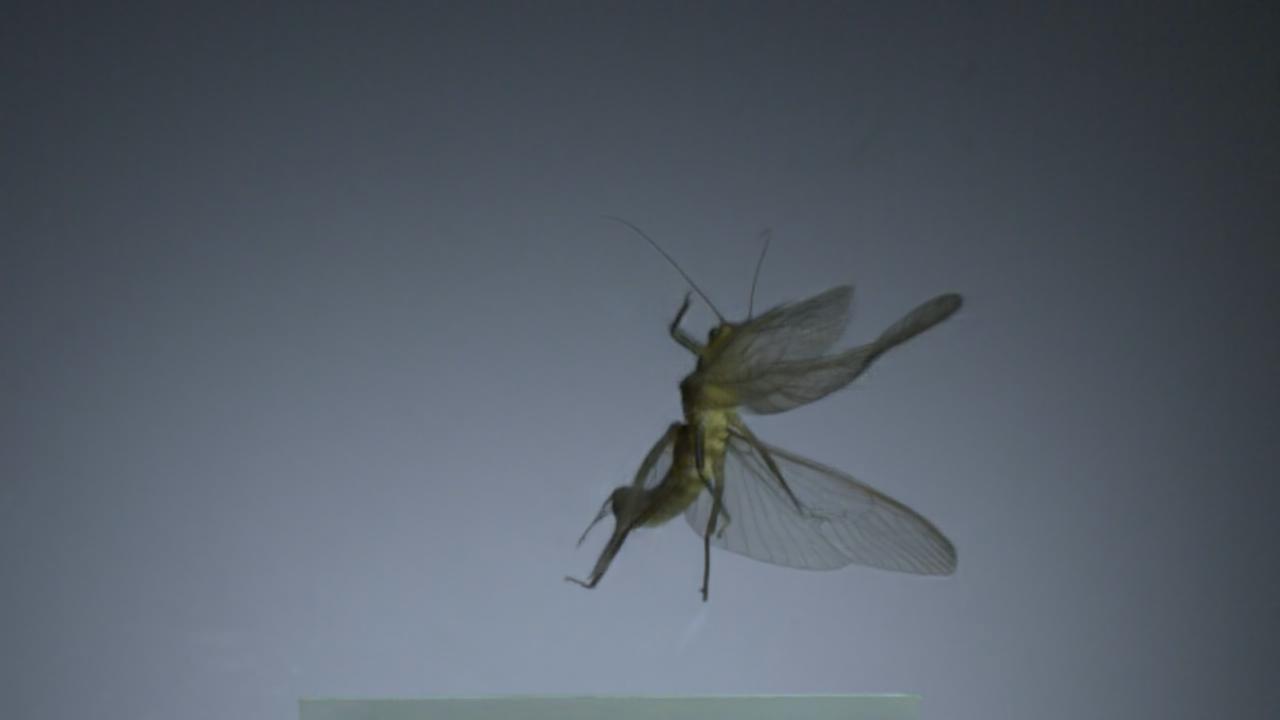}
    \end{subfigure}
    \hfill
    \begin{subfigure}[b]{0.1\textwidth}
        \centering
        \includegraphics[width=\textwidth]{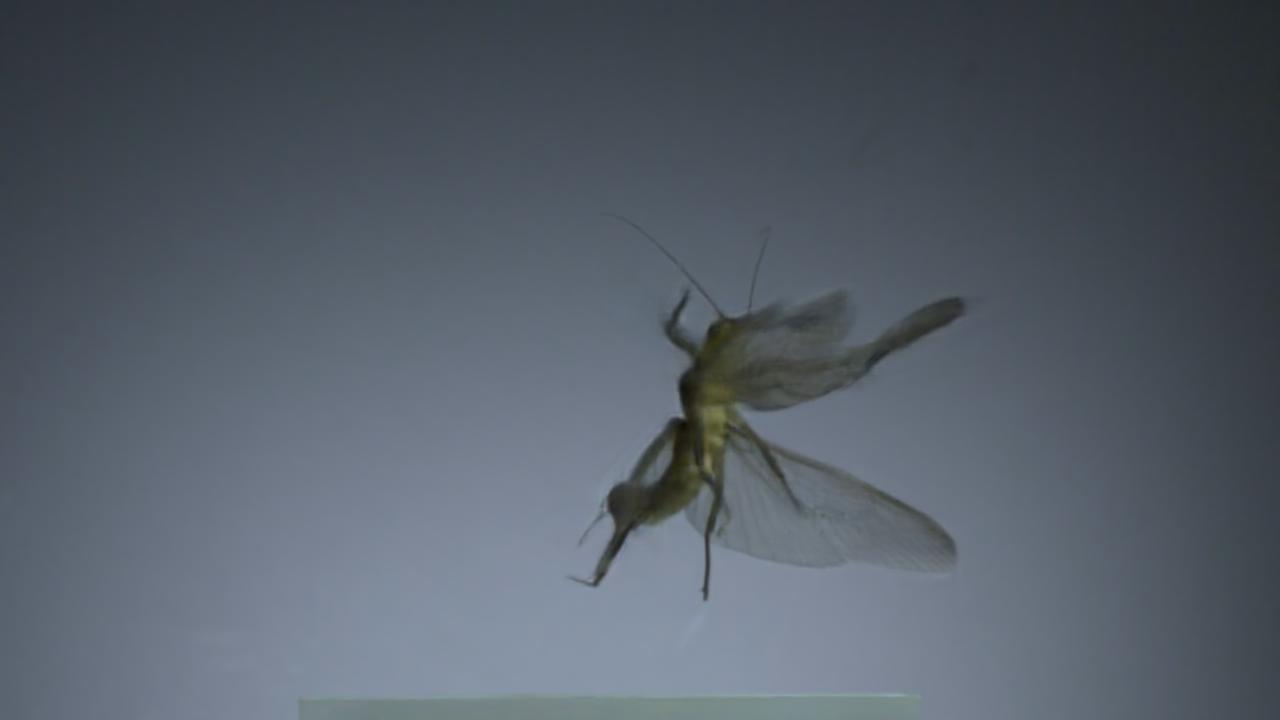}
    \end{subfigure}
    \hfill
    \begin{subfigure}[b]{0.1\textwidth}
        \centering
        \includegraphics[width=\textwidth]{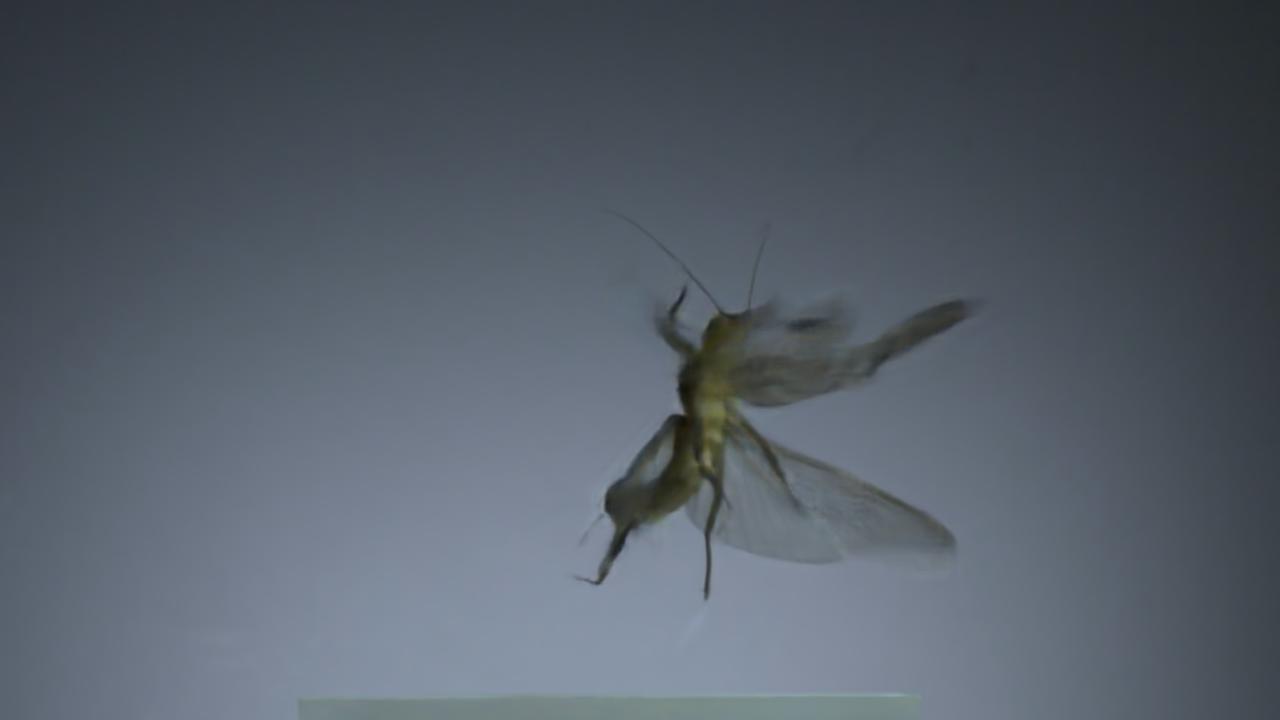}
    \end{subfigure}
    \hfill
    \begin{subfigure}[b]{0.1\textwidth}
        \centering
        \includegraphics[width=\textwidth]{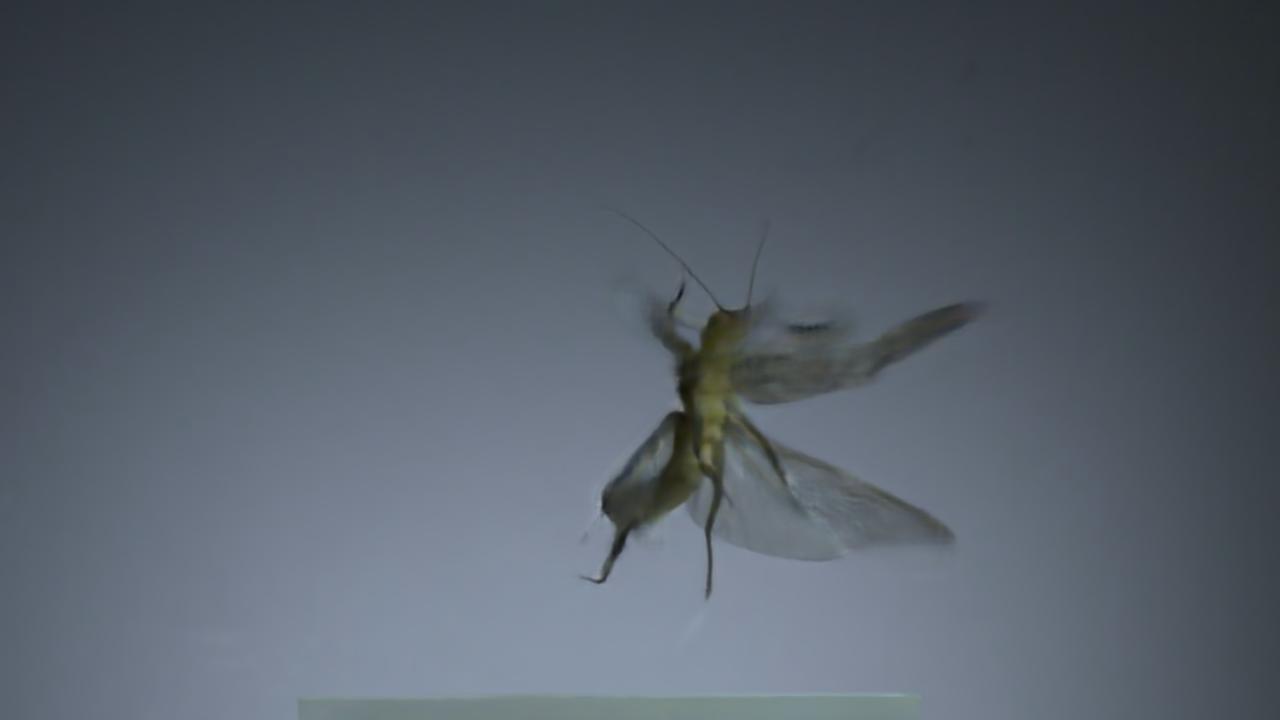}
    \end{subfigure}
    \hfill
    \begin{subfigure}[b]{0.1\textwidth}
        \centering
        \includegraphics[width=\textwidth]{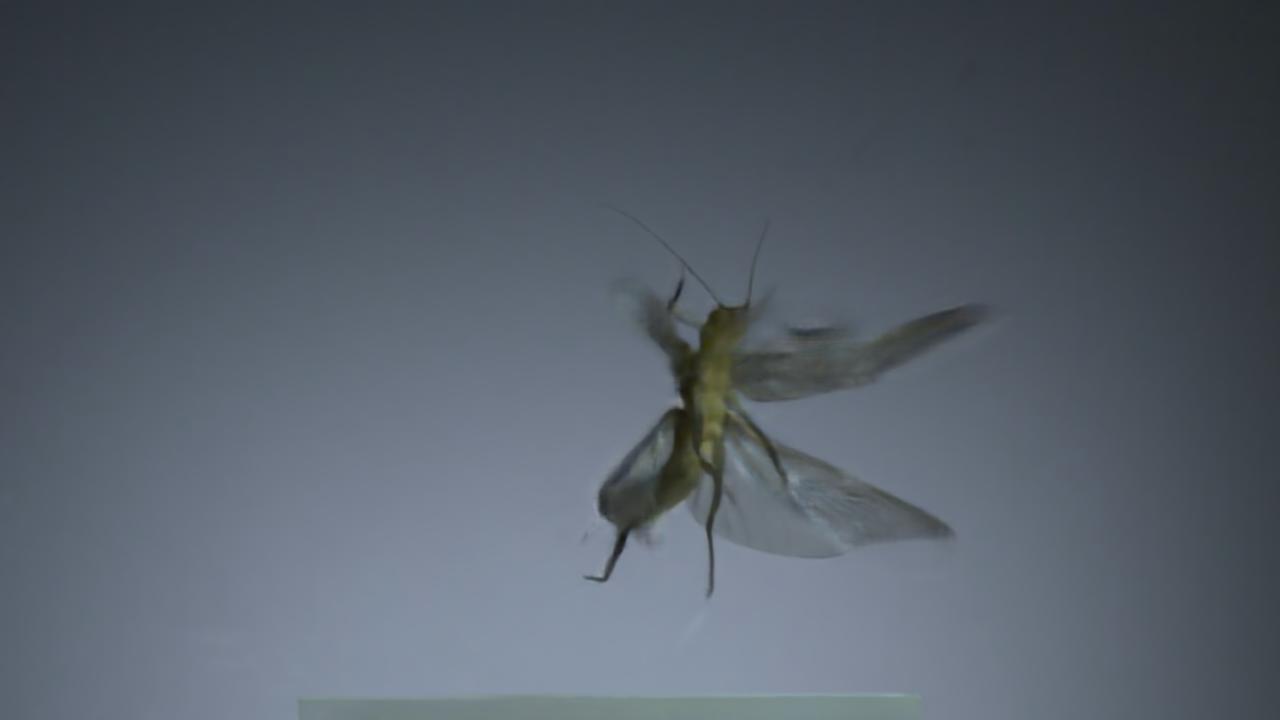}
    \end{subfigure}
    \hfill
    \begin{subfigure}[b]{0.1\textwidth}
        \centering
        \includegraphics[width=\textwidth]{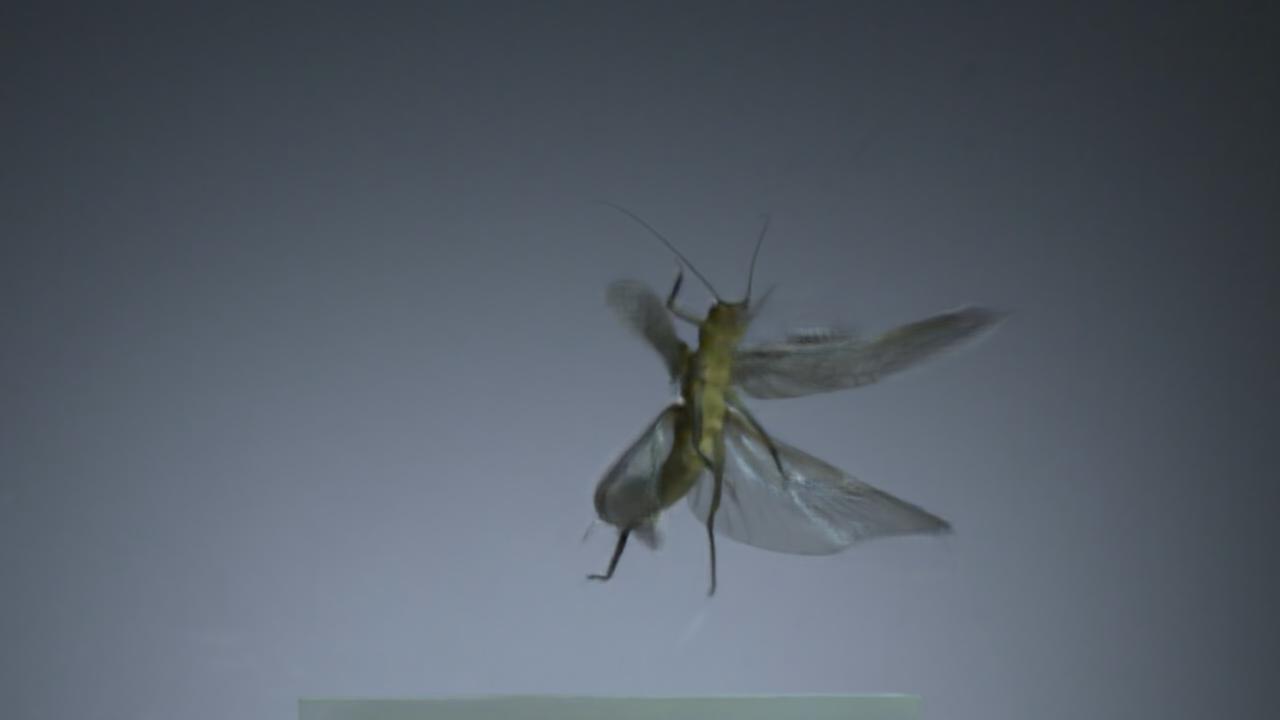}
    \end{subfigure}
    \hfill
    \begin{subfigure}[b]{0.1\textwidth}
        \centering
        \includegraphics[width=\textwidth]{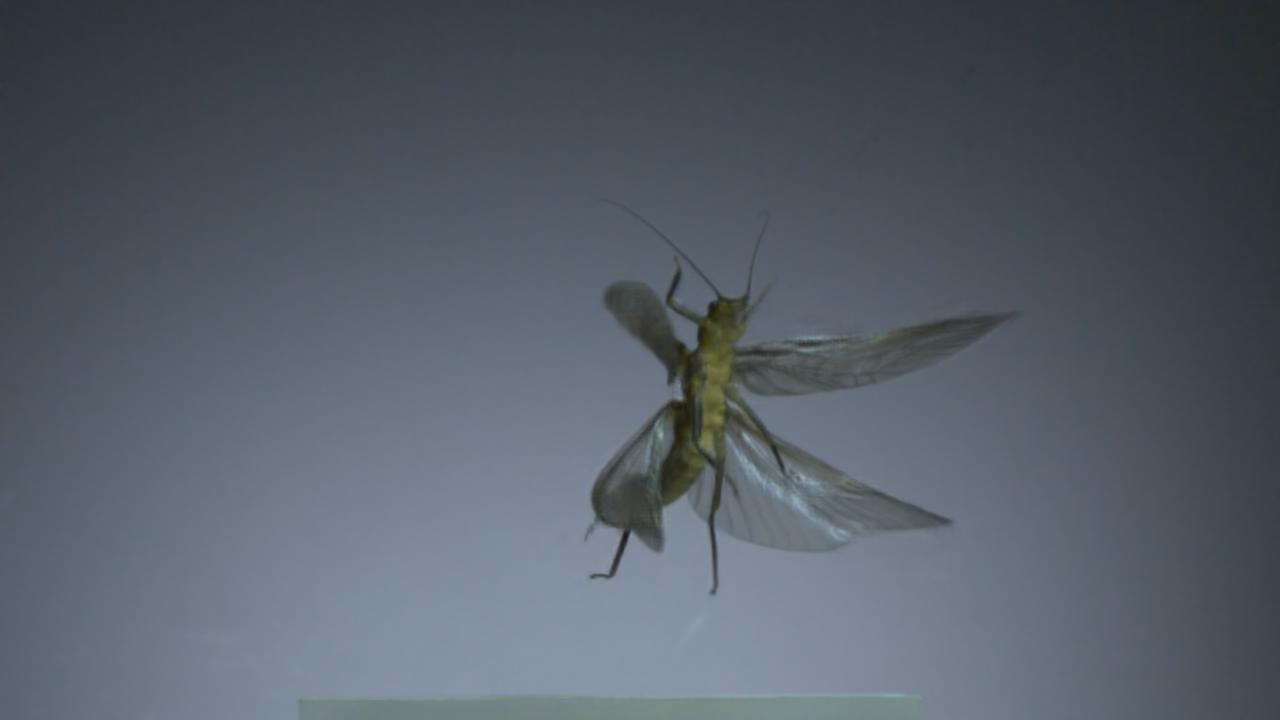}
    \end{subfigure}
    \hfill
    \begin{subfigure}[b]{0.1\textwidth}
        \centering
        \includegraphics[width=\textwidth]{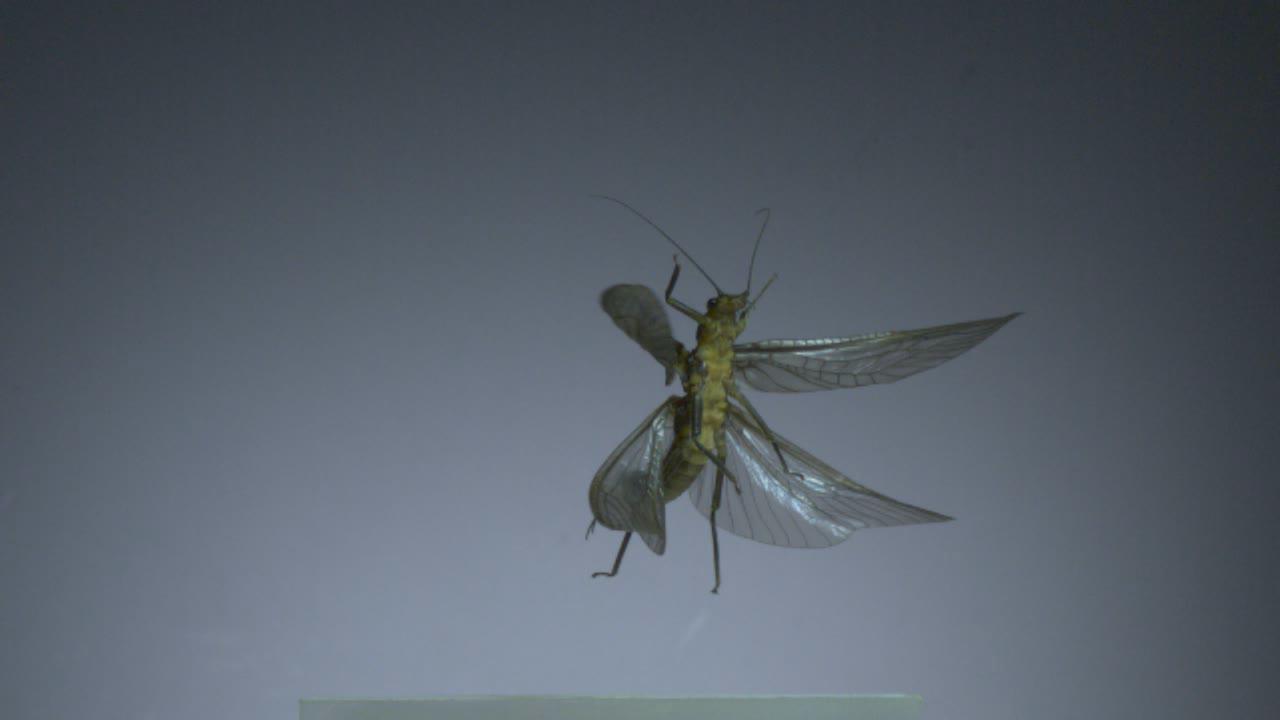}
    \end{subfigure}
    
    
    \begin{subfigure}[b]{0.1\textwidth}
        \centering
        \includegraphics[width=\textwidth]{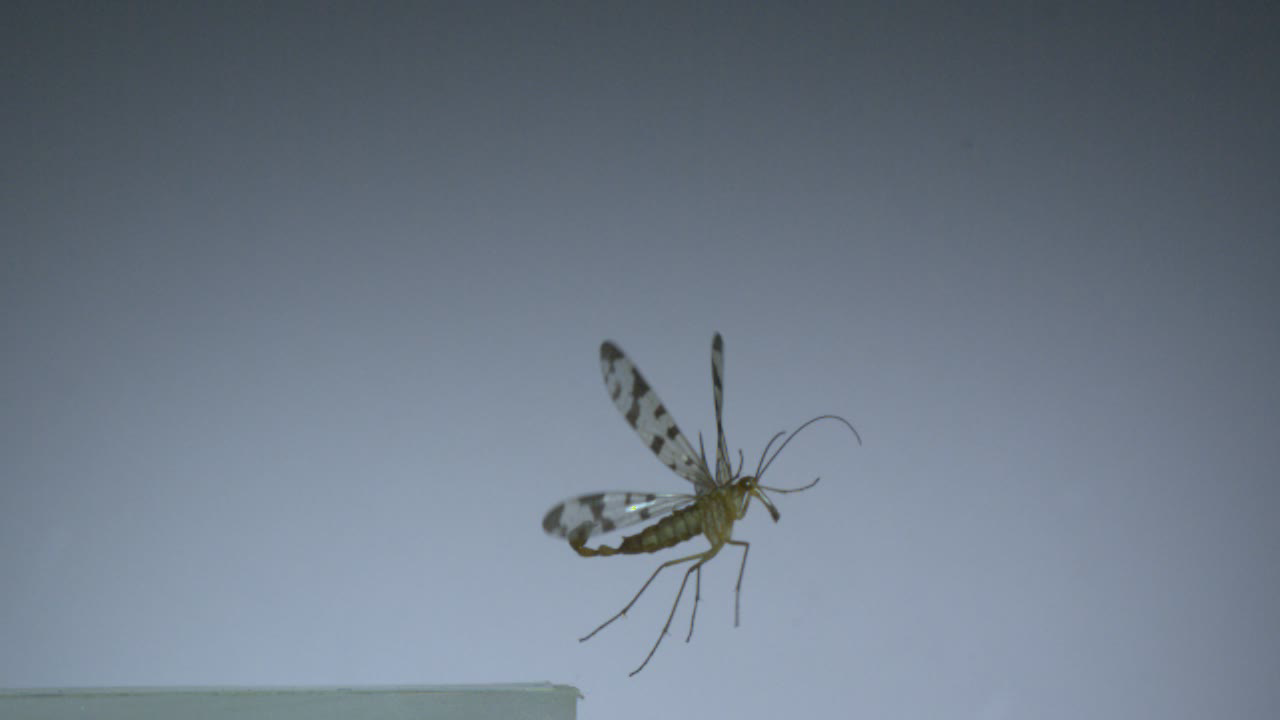}
        \captionsetup{width=\textwidth}
        \caption{t=0}
    \end{subfigure}
    \hfill
    \begin{subfigure}[b]{0.1\textwidth}
        \centering
        \includegraphics[width=\textwidth]{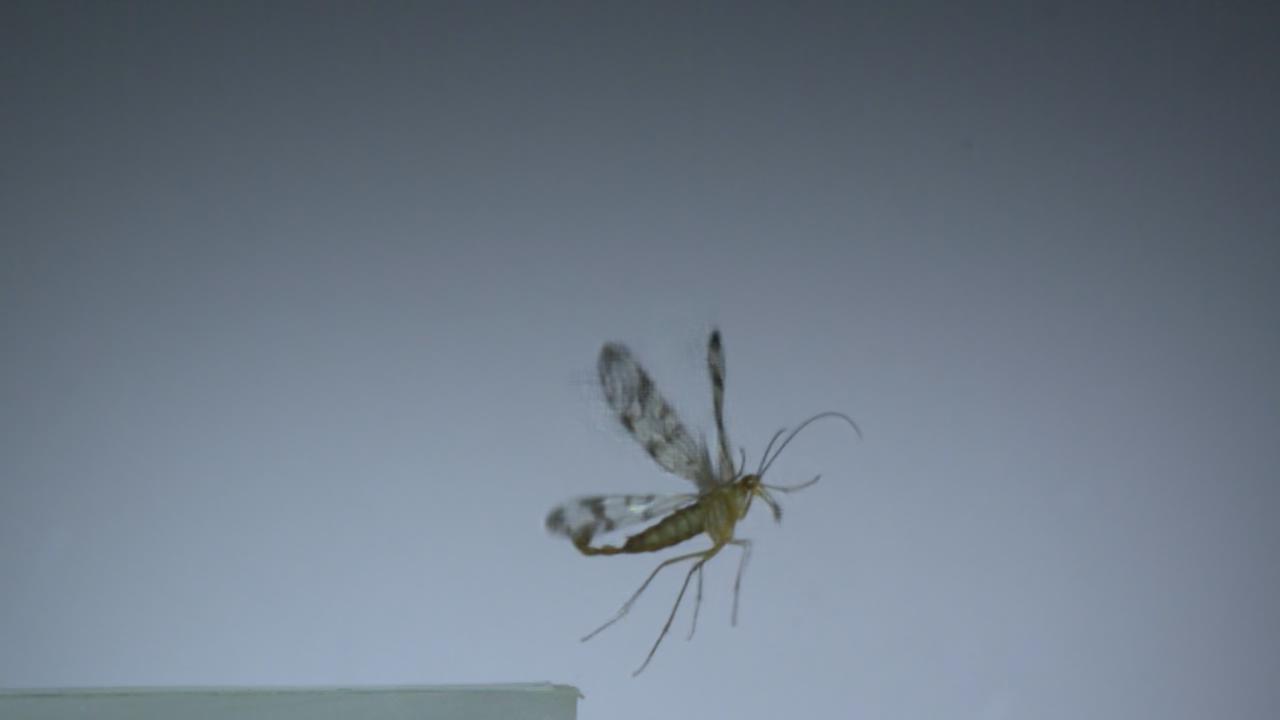}
        \captionsetup{width=\textwidth}
        \caption{t=0.125}
    \end{subfigure}
    \hfill
    \begin{subfigure}[b]{0.1\textwidth}
        \centering
        \includegraphics[width=\textwidth]{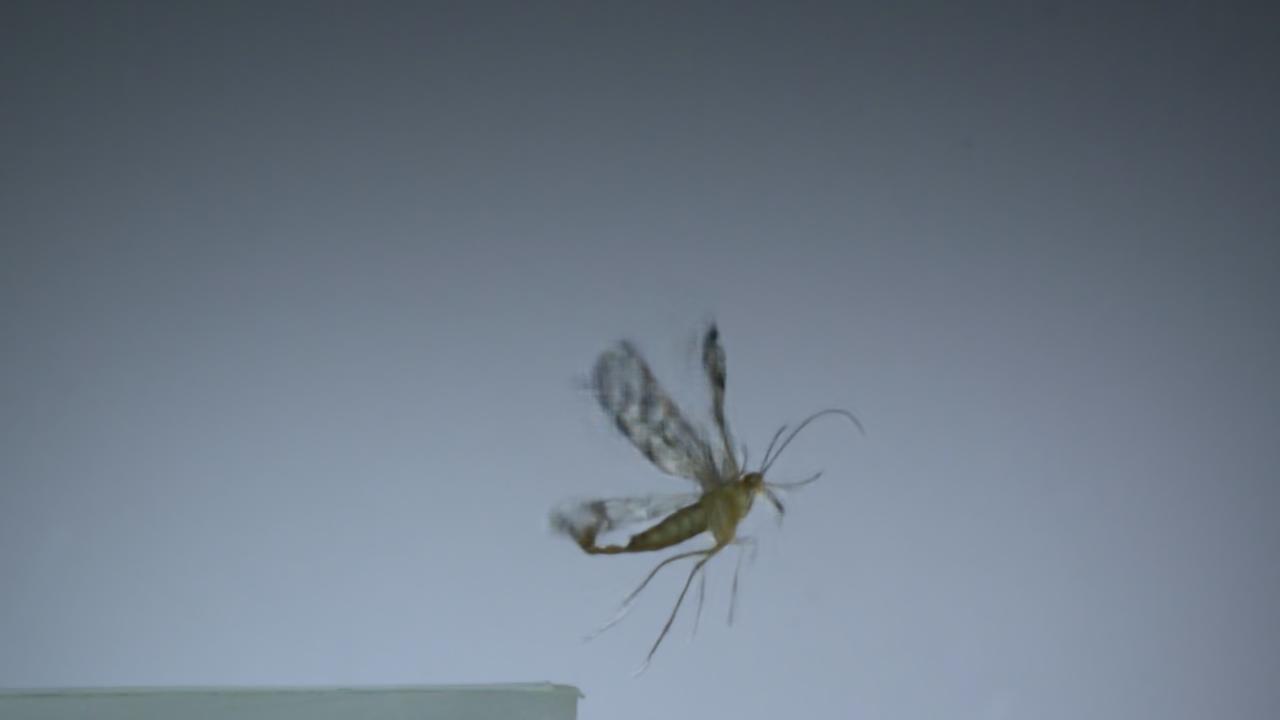}
        \captionsetup{width=\textwidth}
        \caption{t=0.25}
    \end{subfigure}
    \hfill
    \begin{subfigure}[b]{0.1\textwidth}
        \centering
        \includegraphics[width=\textwidth]{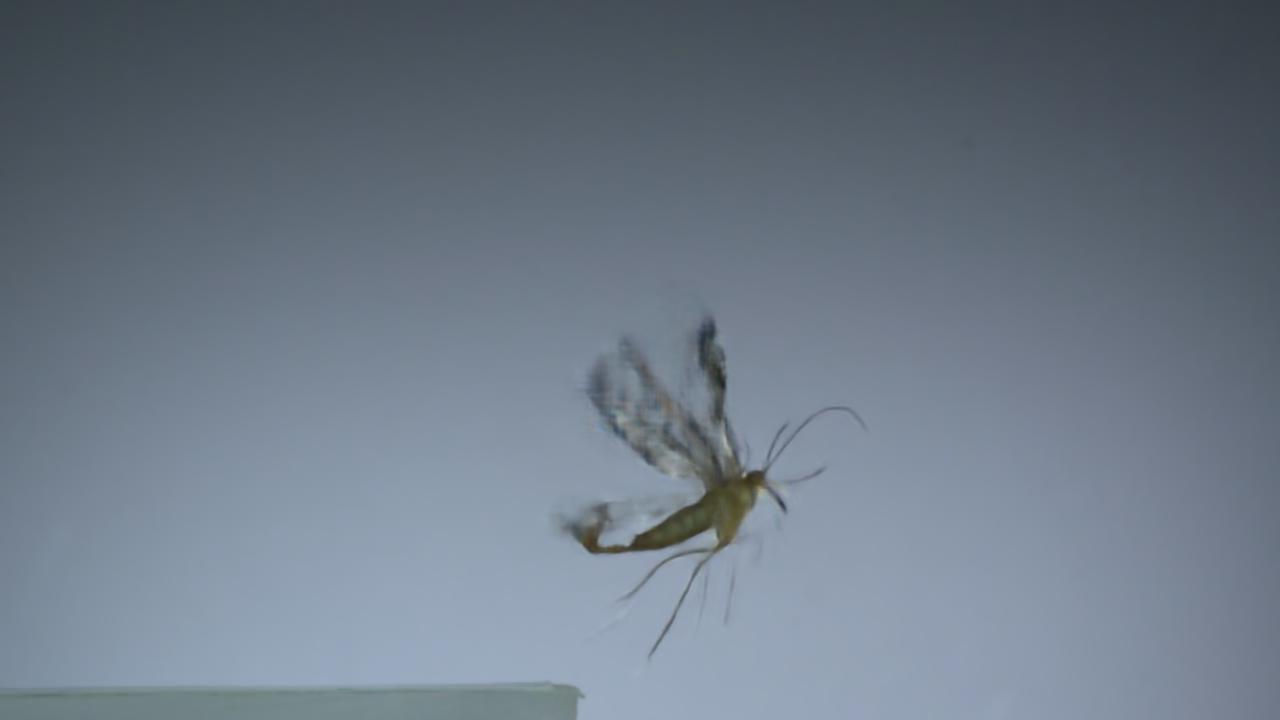}
        \captionsetup{width=\textwidth}
        \caption{t=0.375}
    \end{subfigure}
    \hfill
    \begin{subfigure}[b]{0.1\textwidth}
        \centering
        \includegraphics[width=\textwidth]{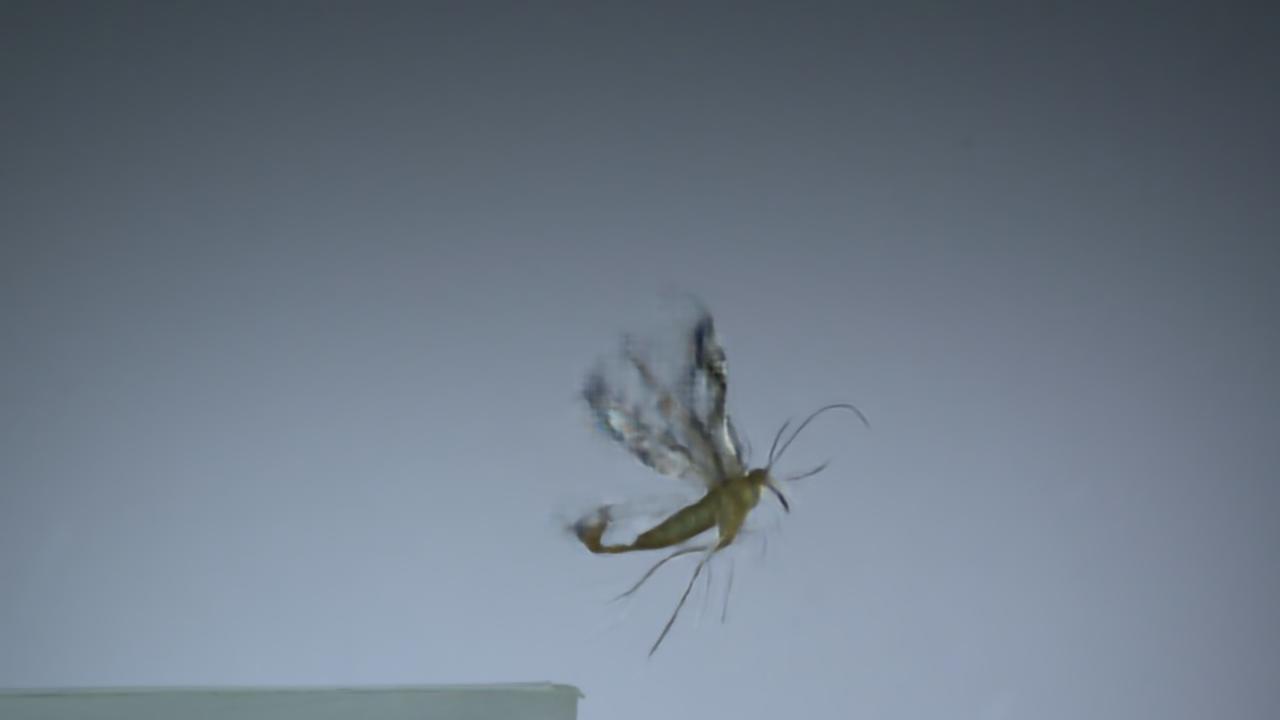}
        \captionsetup{width=\textwidth}
        \caption{t=0.5}
    \end{subfigure}
    \hfill
    \begin{subfigure}[b]{0.1\textwidth}
        \centering
        \includegraphics[width=\textwidth]{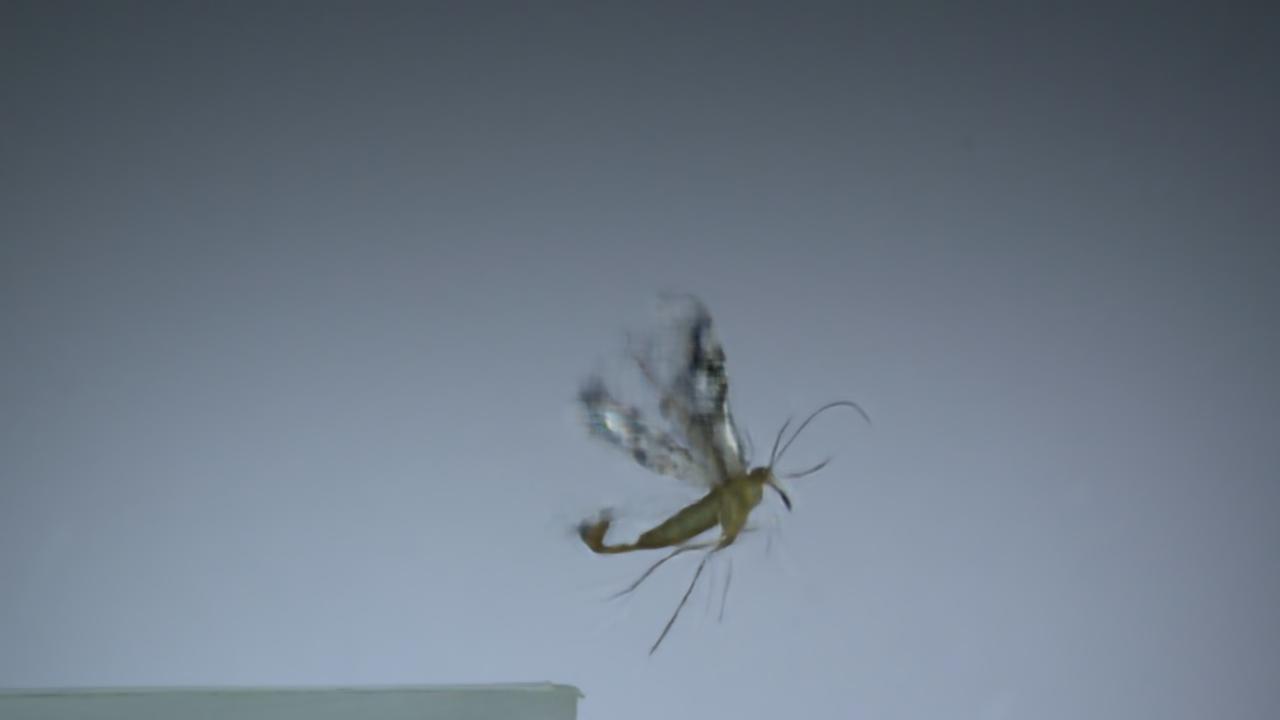}
        \captionsetup{width=\textwidth}
        \caption{t=0.625}
    \end{subfigure}
    \hfill
    \begin{subfigure}[b]{0.1\textwidth}
        \centering
        \includegraphics[width=\textwidth]{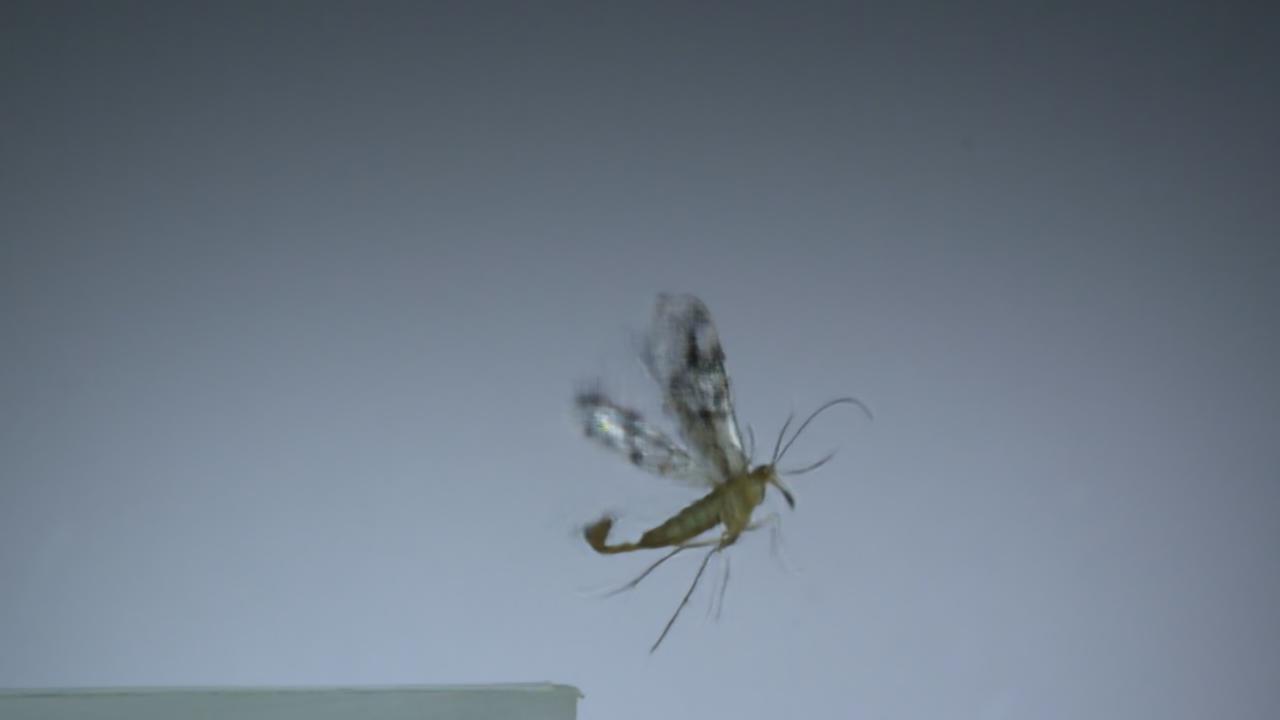}
        \captionsetup{width=\textwidth}
        \caption{t=0.75}
    \end{subfigure}
    \hfill
    \begin{subfigure}[b]{0.1\textwidth}
        \centering
        \includegraphics[width=\textwidth]{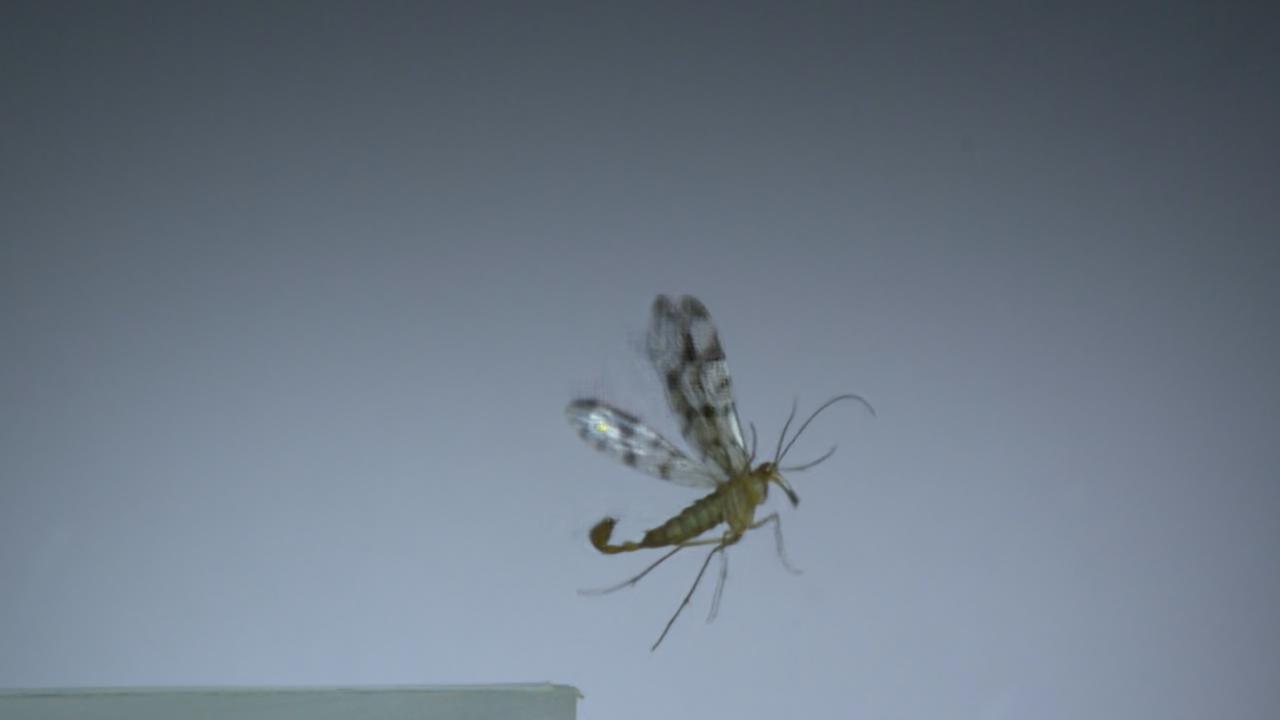}
        \captionsetup{width=\textwidth}
        \caption{t=0.875}
    \end{subfigure}
    \hfill
    \begin{subfigure}[b]{0.1\textwidth}
        \centering
        \includegraphics[width=\textwidth]{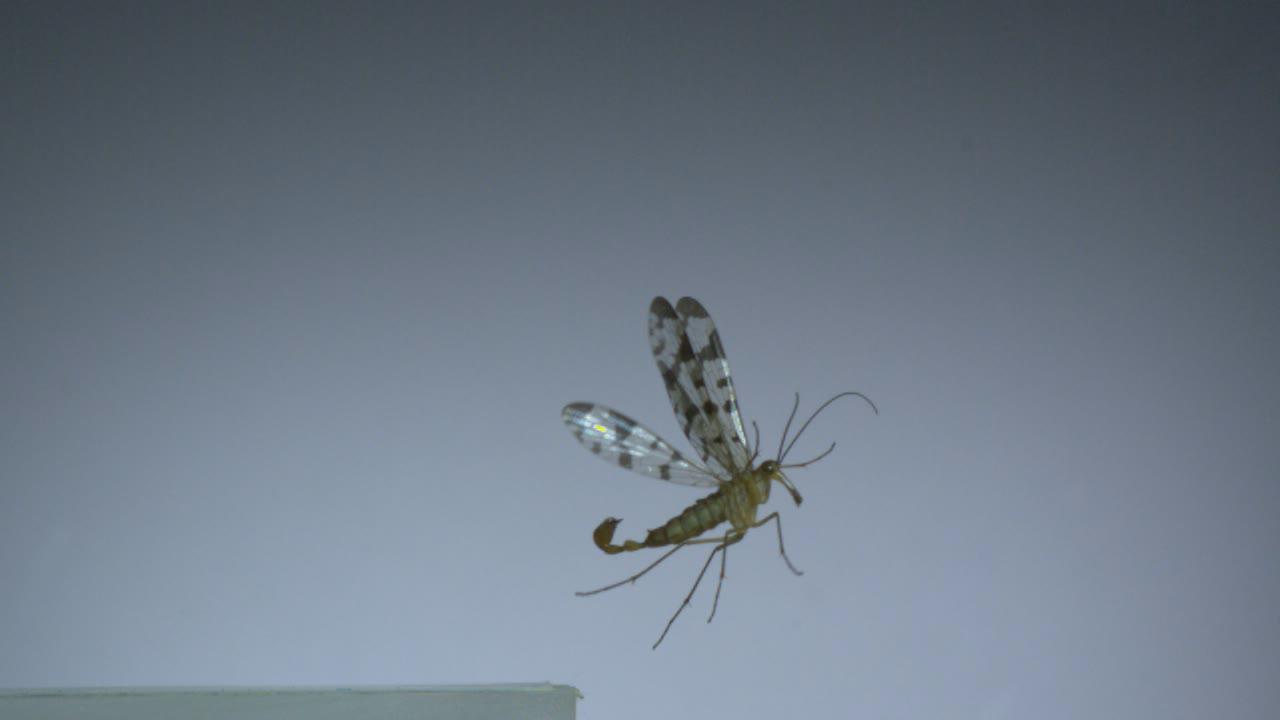}
        \captionsetup{width=\textwidth}
        \caption{t=1}
    \end{subfigure}
    

    \end{center}
    \caption{{Qualitative Results for \eightx{} video frame interpolation on Insect Motion Videos. Frame at $t=0$ and $t=1$ are given as inputs to the network to predict the remaining 7 intermediate frames. Original Videos acquired from \texttt{AntLab} Youtube Channel. }}
    \label{fig:insect_figs}
    \vspace{-6pt}
\end{figure*}
We show additional qualitative results by applying frame interpolation technique on insect motion videos in \figref{fig:insect_figs}. We believe that this application is of immense use for closer inspection of biological properties from videos. We obtain videos from AntLab Youtube channel\footnote{\url{https://www.youtube.com/user/adrianalansmith}} that have insect takeoff and flying captured at very high FPS. We down-sample the frame rate to 15FPS and apply our interpolation network to recover videos of higher frame rate. We apply our \eightx{} model once to obtain videos of 120FPS. The images are shown in \figref{fig:insect_figs}. Complete videos are available in our supplementary video.


{\bf Middlebury Dataset.} 
%

\begin{figure*}[t]
    \begin{center}
    
    \begin{subfigure}[b]{0.18\textwidth}
        \centering
        \includegraphics[width=\textwidth]{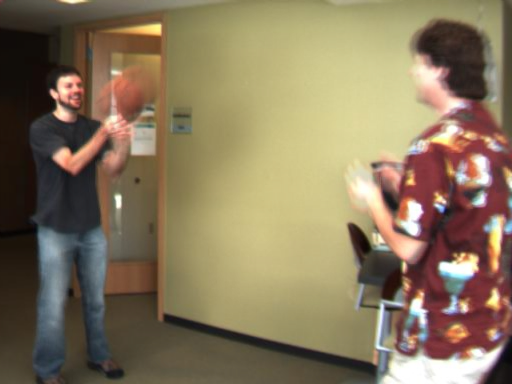}
    \end{subfigure}
    \hfill
    \begin{subfigure}[b]{0.18\textwidth}
        \centering
        \includegraphics[width=\textwidth]{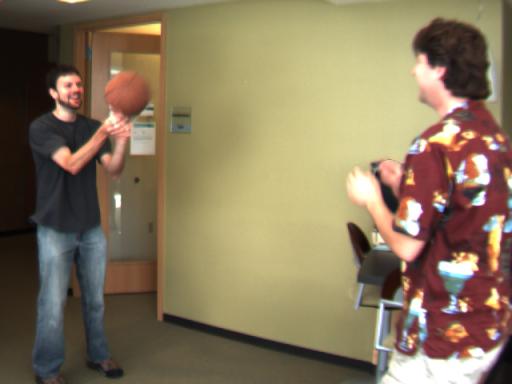}
    \end{subfigure}
    \hfill
    \begin{subfigure}[b]{0.18\textwidth}
        \centering
        \includegraphics[width=\textwidth]{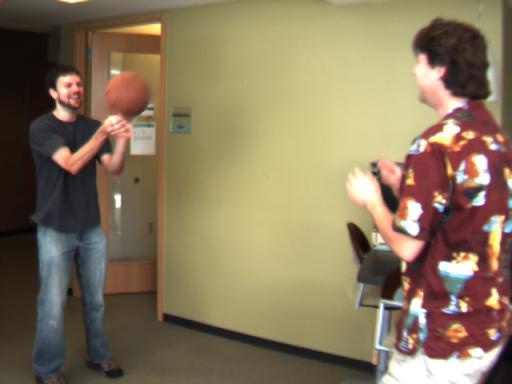}
    \end{subfigure}
    \hfill
    \begin{subfigure}[b]{0.18\textwidth}
        \centering
        \includegraphics[width=\textwidth]{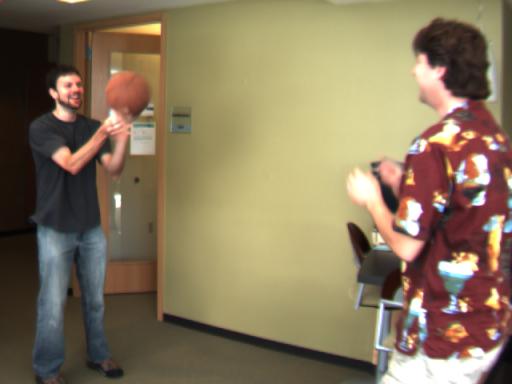}
    \end{subfigure}
    \hfill
    \begin{subfigure}[b]{0.18\textwidth}
        \centering
        \includegraphics[width=\textwidth]{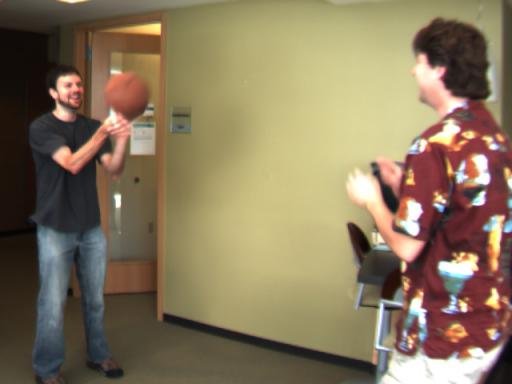}
    \end{subfigure}
    
     \begin{subfigure}[b]{0.18\textwidth}
        \centering
        \includegraphics[width=\textwidth]{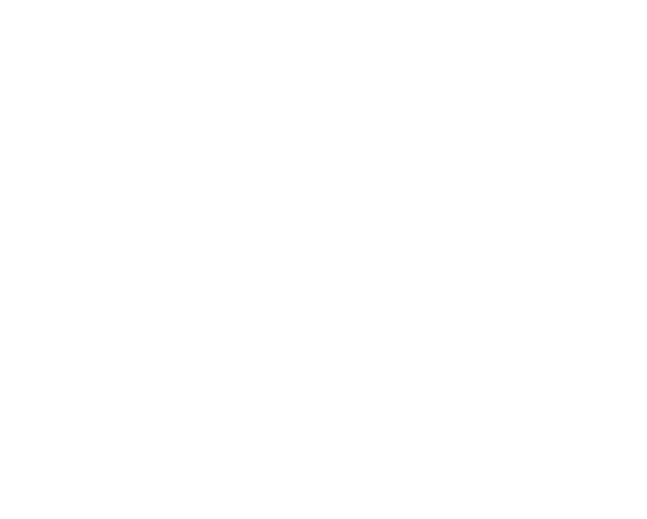}
    \end{subfigure}
    \hfill
    \begin{subfigure}[b]{0.18\textwidth}
        \centering
        \includegraphics[width=\textwidth]{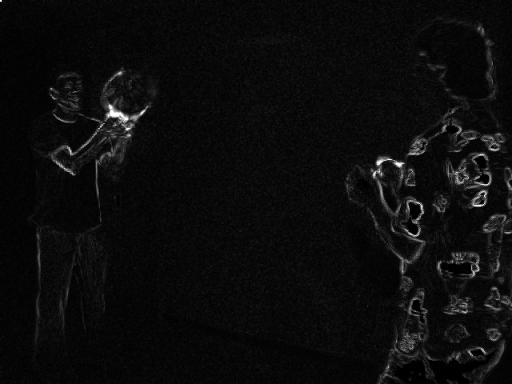}
        \caption*{SSMO~\cite{jiang2018super}; IE=5.37}
    \end{subfigure}
    \hfill
    \begin{subfigure}[b]{0.18\textwidth}
        \centering
        \includegraphics[width=\textwidth]{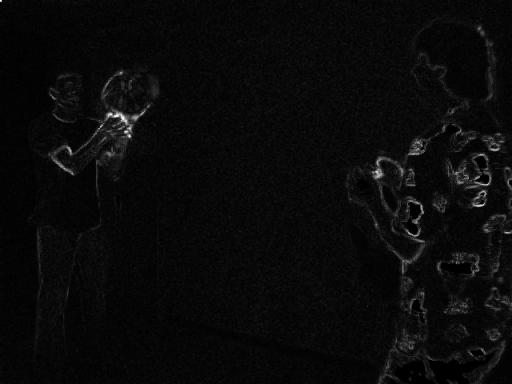}
        \caption*{BMBC~\cite{park2020bmbc}; IE=4.08}
    \end{subfigure}
    \hfill
    \begin{subfigure}[b]{0.18\textwidth}
        \centering
        \includegraphics[width=\textwidth]{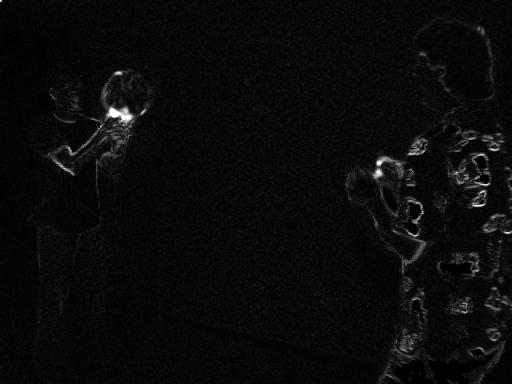}
        \caption*{EDSC~\cite{cheng2020multiple}; IE=4.89}
    \end{subfigure}
    \hfill
    \begin{subfigure}[b]{0.18\textwidth}
        \centering
        \includegraphics[width=\textwidth]{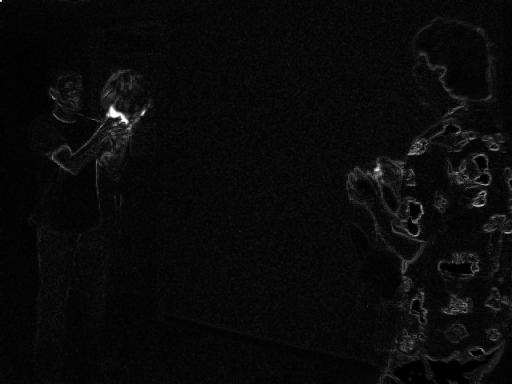}
        \caption*{FLAVR, IE=4.20}
    \end{subfigure}
    
    
    \vspace{1em}
    \begin{subfigure}[b]{0.18\textwidth}
        \centering
        \includegraphics[width=\textwidth]{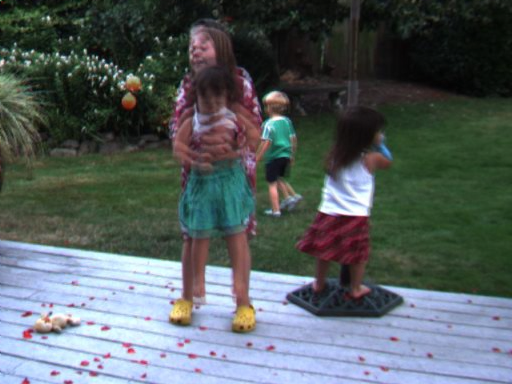}
    \end{subfigure}
    \hfill
    \begin{subfigure}[b]{0.18\textwidth}
        \centering
        \includegraphics[width=\textwidth]{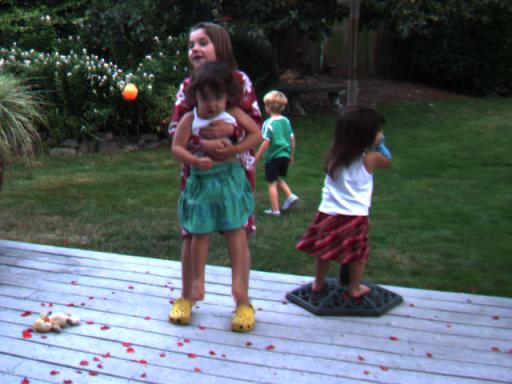}
    \end{subfigure}
    \hfill
    \begin{subfigure}[b]{0.18\textwidth}
        \centering
        \includegraphics[width=\textwidth]{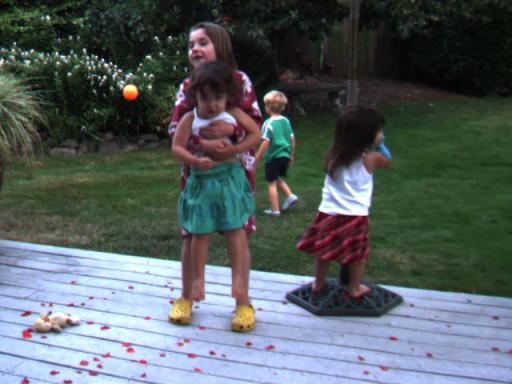}
    \end{subfigure}
    \hfill
    \begin{subfigure}[b]{0.18\textwidth}
        \centering
        \includegraphics[width=\textwidth]{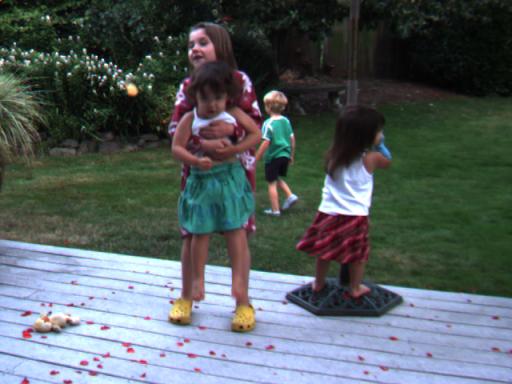}
    \end{subfigure}
    \hfill
    \begin{subfigure}[b]{0.18\textwidth}
        \centering
        \includegraphics[width=\textwidth]{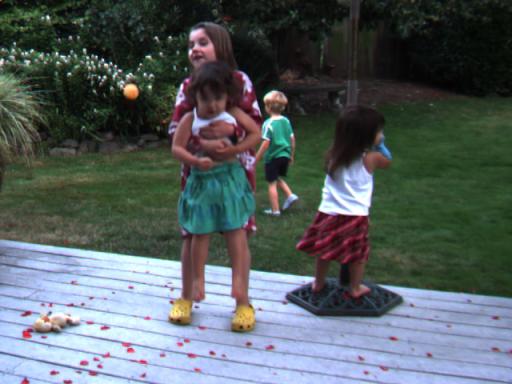}
    \end{subfigure}
    
     \begin{subfigure}[b]{0.18\textwidth}
        \centering
        \includegraphics[width=\textwidth]{scripts/Middleburry/blank.png}
    \end{subfigure}
    \hfill
    \begin{subfigure}[b]{0.18\textwidth}
        \centering
        \includegraphics[width=\textwidth]{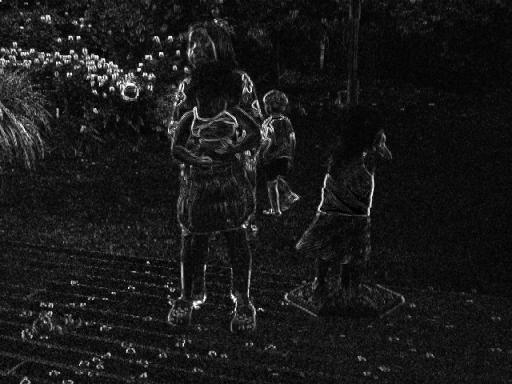}
        \caption*{SSMO~\cite{jiang2018super}; IE=9.56}
    \end{subfigure}
    \hfill
    \begin{subfigure}[b]{0.18\textwidth}
        \centering
        \includegraphics[width=\textwidth]{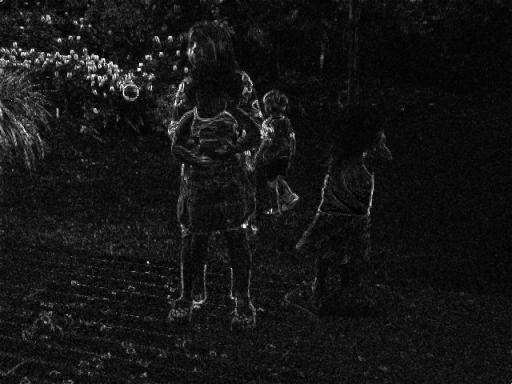}
        \caption*{BMBC~\cite{park2020bmbc}; IE=7.79}
    \end{subfigure}
    \hfill
    \begin{subfigure}[b]{0.18\textwidth}
        \centering
        \includegraphics[width=\textwidth]{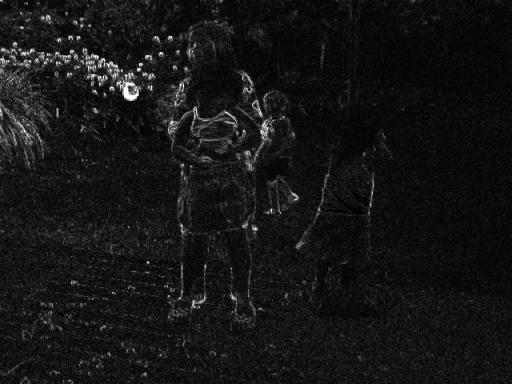}
        \caption*{EDSC~\cite{cheng2020multiple}; IE=8.05}
    \end{subfigure}
    \hfill
    \begin{subfigure}[b]{0.18\textwidth}
        \centering
        \includegraphics[width=\textwidth]{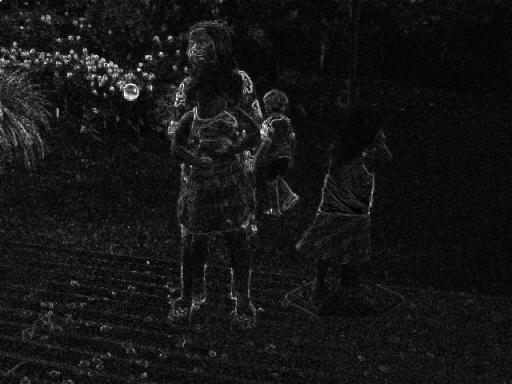}
        \caption*{\textbf{FLAVR, IE=7.35}}
    \end{subfigure}
    
    \vspace{1em}
    \begin{subfigure}[b]{0.18\textwidth}
        \centering
        \includegraphics[width=\textwidth]{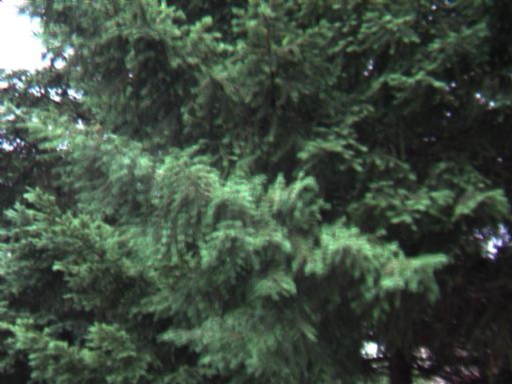}
    \end{subfigure}
    \hfill
    \begin{subfigure}[b]{0.18\textwidth}
        \centering
        \includegraphics[width=\textwidth]{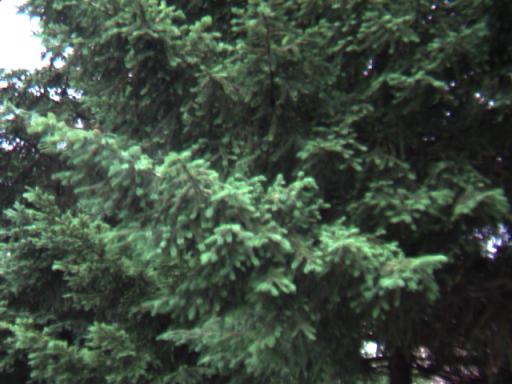}
    \end{subfigure}
    \hfill
    \begin{subfigure}[b]{0.18\textwidth}
        \centering
        \includegraphics[width=\textwidth]{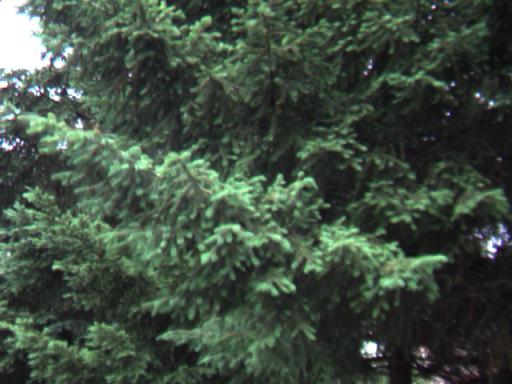}
    \end{subfigure}
    \hfill
    \begin{subfigure}[b]{0.18\textwidth}
        \centering
        \includegraphics[width=\textwidth]{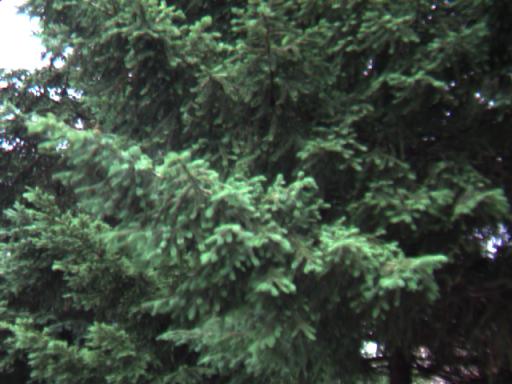}
    \end{subfigure}
    \hfill
    \begin{subfigure}[b]{0.18\textwidth}
        \centering
        \includegraphics[width=\textwidth]{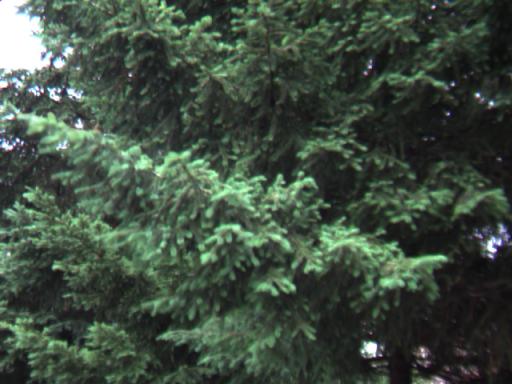}
    \end{subfigure}
    
     \begin{subfigure}[b]{0.18\textwidth}
        \centering
        \includegraphics[width=\textwidth]{scripts/Middleburry/blank.png}
    \end{subfigure}
    \hfill
    \begin{subfigure}[b]{0.18\textwidth}
        \centering
        \includegraphics[width=\textwidth]{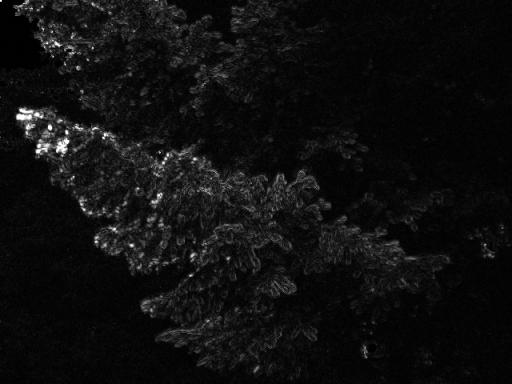}
        \caption*{SSMO~\cite{jiang2018super}; IE=6.73}
    \end{subfigure}
    \hfill
    \begin{subfigure}[b]{0.18\textwidth}
        \centering
        \includegraphics[width=\textwidth]{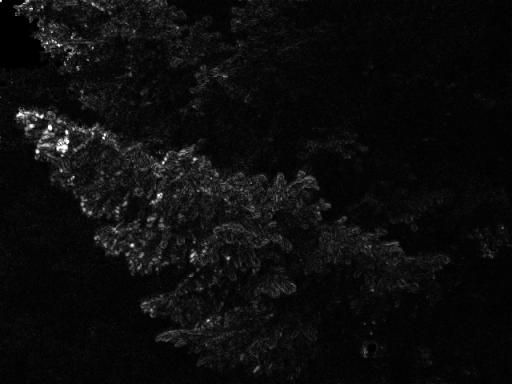}
        \caption*{BMBC~\cite{park2020bmbc}; IE=5.55}
    \end{subfigure}
    \hfill
    \begin{subfigure}[b]{0.18\textwidth}
        \centering
        \includegraphics[width=\textwidth]{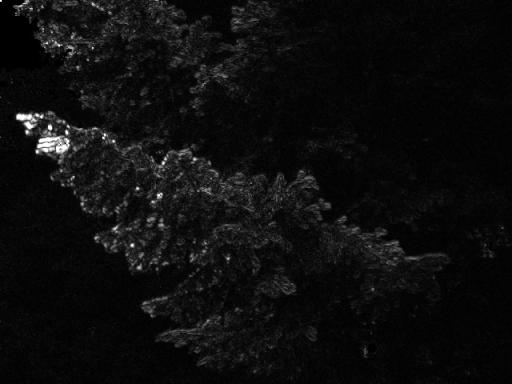}
        \caption*{EDSC~\cite{cheng2020multiple}; IE=6.42}
    \end{subfigure}
    \hfill
    \begin{subfigure}[b]{0.18\textwidth}
        \centering
        \includegraphics[width=\textwidth]{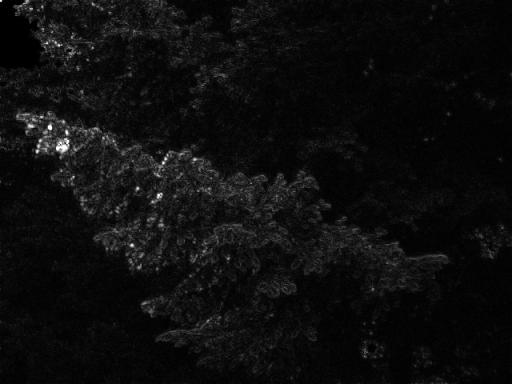}
        \caption*{\textbf{FLAVR, IE=5.53}}
    \end{subfigure}
    
    
    \end{center}
    \caption{Interpolation results for \twox{} interpolation on Middleburry test set. The leftmost images in each row repesent the overlayed inputs. The first row in each set represents the interpolated frame, while the second row shows the error maps with respect to the ground truth. IE shows the interpolation error of the method. The interpolation errors for all the baselines are reported on the \href{https://vision.middlebury.edu/flow/eval/results/results-i1.php}{official leaderboard}.}
    \label{fig:middleburry_figs}
    \vspace{-6pt}
\end{figure*}
\red{We evaluate FLAVR on the publicly available test images from Middleburry dataset~\cite{scharstein2014high} on the task of single frame interpolation. However, Middleburry has test samples with only two input frames while FLAVR requires 4-frame inputs. In those examples, we simply duplicate them into 4 frames and evaluate with FLAVR. For two frame sequences like \textit{teddy}, duplicating inputs is obviously sub-optimal. On sequences where multi-frame inputs are available, FLAVR outperforms most prior interpolation works like SuperSloMo~\cite{jiang2018super}, BMBC~\cite{park2020bmbc} and EDSC~\cite{cheng2020multiple}. Qualitative results for some such sequences are presented in \figref{fig:middleburry_figs}. 
The complete results are available on the \href{https://vision.middlebury.edu/flow/eval/results/results-i1.php}{public leaderboard}}.

\section{User study}
\label{appendix:user_study}

We carry the user study on the Amazon Mechanical Turk (AMT) platform. We select two representative works that belong to two broad families that perform linear (SuperSloMo~\cite{jiang2018super}) and quadratic (QVI~\cite{xu2019quadratic}) warping for multi-frame interpolation. Then, we compare each video generated by \Ours{} with videos generated using each of SuperSloMo and QVI separately. For this purpose we use \emph{all} 90 HD videos from the DAVIS dataset, generate \eightx{} interpolated videos and place the two interpolated videos one beside the other and randomly shuffle the order of videos. We then show each pair of videos to 6 AMT workers and ask them to choose which video, right or left, looked more realistic. The method preferred by more users is chosen as a winner for that particular video. In case of tie, that is if each method is chosen by 3 users, we place the video under ``no preference'' category. Workers are paid in accordance with minimum wages rules. With this setting, we find that users overwhelmingly chose our videos in preference against SuperSloMo~\cite{jiang2018super}. More details are provided in subsection 5.1 of the main paper.



\section{Training details}
\label{appendix:training_details}

We train the \twox{} interpolation network on Vimeo-90K dataset and \fourx{} and \eightx{} interpolation networks on the GoPro dataset and use the official train and validation splits with the sampling strategy explained in subsection 3 of the paper. We use a crop size of $256 \stimes 256$ and $512 \stimes 512$ for Vimeo-90K and GoPro datasets, respectively. We employ random frame order reversal and random horizontal flipping as augmentation strategies on both the datasets. We use use an initial learning rate of $2\times10^{-4}$ and divide the learning rate by $2$ whenever the training plateaus. We train the \twox{} interpolation network for 200 epochs, while \fourx{} and \eightx{} interpolation network were trained for 120 epochs. We use a mini-batch size of $64$ on Vimeo-90K dataset and $32$ on GoPro dataset, and train our network on $8$ 2080Ti GPUs. We reduce the learning rate by half whenever the training plateaus which is cross-validated by the validation set. We apply mean normalization once for every mini-batch of input frames separately rather than using global mean normalization or batch normalization inside the network to achieve training stability. We use 8 GPUs and a mini-batch of 32 to train each model, and training is completed in about 36 hours for \twox{} and 22 hours for \eightx{} interpolation networks. 

\section{Benchmarking inference time}
\label{appendix:inference_time}

The inference time benchmarking was performed using an NVIDIA-2080Ti GPU with 12GB memory. The calculated time only includes forward pass excluding the data pre-processing time and CPU/GPU transfer. The results were obtained by averaging over 100 samples from Adobe-240FPS dataset using 512$\times$512 crop size. For multi-frame interpolation, the time required is calculated as the aggregate time required for interpolating all the frames. Non-blocking CUDA operations as well as GPU warm start time were accounted for during inference time computation.

\section{Statement on potential negative impact}
\label{sec:negImpact}

Frame interpolation aims to generate non-existent frames between existing frames of a video. While achieving state-of-the-art performance using simple architectures through \Ours{} is a plus, any kind of generative models can be misused to forge or tamper a video which may have a negative impact on applications where outputs of \Ours{} have a bearing on reliability. Moreover, one of the applications of \Ours{} is to improve object tracking in videos, which might have a potential to be used in surveillance for nefarious purposes.

\bibliographystyle{ieee_fullname}
\bibliography{main}

\end{document}